%% file: main.tex
\documentclass[fleqn,10pt]{wlscirep}
\usepackage[utf8]{inputenc}
\usepackage[T1]{fontenc}
\usepackage{multirow}
\usepackage{placeins}

\title{Transformation of Biological Networks into Images via Semantic Cartography for Visual Interpretation and Scalable Deep Analysis}

\author[1]{Sakib Mostafa}
\author[1,*]{Lei Xing}
\author[1,*]{Md Tauhidul Islam}
\affil[1]{Department of Radiation Oncology, Stanford University, Stanford, CA, USA}

\affil[*]{Correspondence: tauhid@stanford.edu, lei@stanford.edu}

\begin{abstract}
Complex biological networks are fundamental to biomedical science, capturing interactions among molecules, cells, genes, and tissues. Deciphering these networks is critical for understanding health and disease, yet their scale and complexity represent a daunting challenge for current computational methods. Traditional biological network analysis methods, including deep learning approaches, while powerful, face inherent challenges such as limited scalability, oversmoothing long-range dependencies, difficulty in multimodal integration, expressivity bounds, and poor interpretability. We present Graph2Image, a framework that transforms large biological networks into sets of two-dimensional images by spatially arranging representative network nodes on a 2D grid. This transformation decouples the nodes as images, enabling the use of convolutional neural networks (CNNs) with global receptive fields and multi-scale pyramids, thus overcoming limitations of existing biological network analysis methods in scalability, memory efficiency, and long-range context capture. Graph2Image also facilitates seamless integration with other imaging and omics modalities and enhances interpretability through direct visualization of node-associated images. When applied to several large-scale biological network datasets, Graph2Image improved classification accuracy by up to 67.2\% over existing methods and provided interpretable visualizations that revealed biologically coherent patterns. It also allows analysis of very large biological networks (nodes > 1 billion) on a personal computer. Graph2Image thus provides a scalable, interpretable, and multimodal-ready approach for biological network analysis, offering new opportunities for disease diagnosis and the study of complex biological systems.
\end{abstract}

\begin{document}

\flushbottom
\maketitle
\thispagestyle{empty}

\section{Introduction}
Complex biological networks are inherent components of biomedical science, representing the vast web of interactions among molecules, cells, genes, and tissues. Fundamentally, a biological network/graph consists of a set of nodes and the connections between them, known as edges, which represent the node-node relationships. From protein-protein interaction networks that orchestrate cellular function~\cite{tucker2001, barabasi2004} and gene regulatory networks that guide cellular developmental processes~\cite{davidson2002}, to the immense architecture of neurons and their synaptic connections in the brain~\cite{sporns2005, bullmore2009}, these biological graph-based models are critical for representing the mechanisms of health and disease~\cite{ideker2008, kitano2002}. Their applications are vast, spanning metabolic pathways that govern cellular energy~\cite{jeong2000}, microbial interaction networks within the microbiome that influence host health~\cite{rounge2020}, and epidemiological networks that model the spread of infectious diseases~\cite{keeling2005}. Even entire ecosystems can be understood through the structure of their food webs~\cite{dunne2002}. The ability to analyze these diverse and complex biological networks is therefore a prerequisite for advancing our understanding of biological systems~\cite{hartwell1999, alon2007}, identifying therapeutic targets in drug discovery~\cite{yildirim2007, camacho2018next, gaudelet2021utilizing}, and developing new diagnostic strategies~\cite{zhang2017network, chhibber2021}.

The advancement of high-throughput technologies such as single-cell RNA-seq, spatial transcriptomics, proteomics, and metabolomics has led to an explosion in the scale of biomedical data, generating biological networks of unprecedented size and complexity \cite{hasin2017multi}. These biological graphs often contain hundreds of thousands to millions of nodes, accompanied by an even larger number of edges~\cite{szklarczyk2021string}. Analyzing data at this scale poses formidable challenges: the computational and memory demands of processing such massive biological graphs can exceed the capabilities of even state-of-the-art hardware~\cite{batarfi2015large}. As a result, these resource constraints create a critical bottleneck that impedes the translation of large-scale biological data into meaningful and actionable insights~\cite{zou2023survey}.

To address the challenge of analyzing these large-scale biological networks, many analytical techniques and, more recently, deep learning models have been proposed. The analytical methods primarily rely on handcrafted features derived from graph theory for representing the data. These methods include spectral approaches that use the graph Laplacian for community detection~\cite{shi2000normalized, von2007tutorial, ng2002spectral}, and kernel-based methods that measure similarity by counting predefined substructures like random walks, shortest paths, or graphlets~\cite{vishwanathan2010graph, shervashidze2011weisfeiler, shervashidze2009fast, borgwardt2005protein}. While useful for certain tasks, these techniques are often simple and may fail to capture the complex, non-linear patterns inherent in biological data.

To overcome the reliance on engineered features, deep learning methods, specifically, Graph Neural Networks (GNN) have emerged as the dominant paradigm, learning node representations directly from biological network topology and node features through a flexible message-passing framework~\cite{wu2021comprehensive, zhou2020graph, zhang2020deep}. Early GNN models such as Graph Convolutional Networks (GCNs) generalize convolutions to biological graph data by operating in the spectral domain~\cite{kipf2016semi, defferrard2016convolutional}. More flexible spatial methods that directly aggregate information from a node’s neighbors have also been proposed. Among them, GraphSAGE, which introduce neighbor sampling to achieve scalability for massive biological graphs~\cite{hamilton2017inductive, chen2017stochastic}, and Graph Attention Network (GAT), which learn to dynamically weigh the importance of different neighbors are the prominent ones~\cite{velivckovic2017graph, zhang2018gaan}. Further advanced models, such as Graph Isomorphism Network (GIN)~\cite{xu2018powerful, morris2019weisfeiler}, have been developed to improve the expressive power of GNNs. 

\begin{figure}[pt]
    \centering
    \includegraphics[width=\linewidth, page=1]{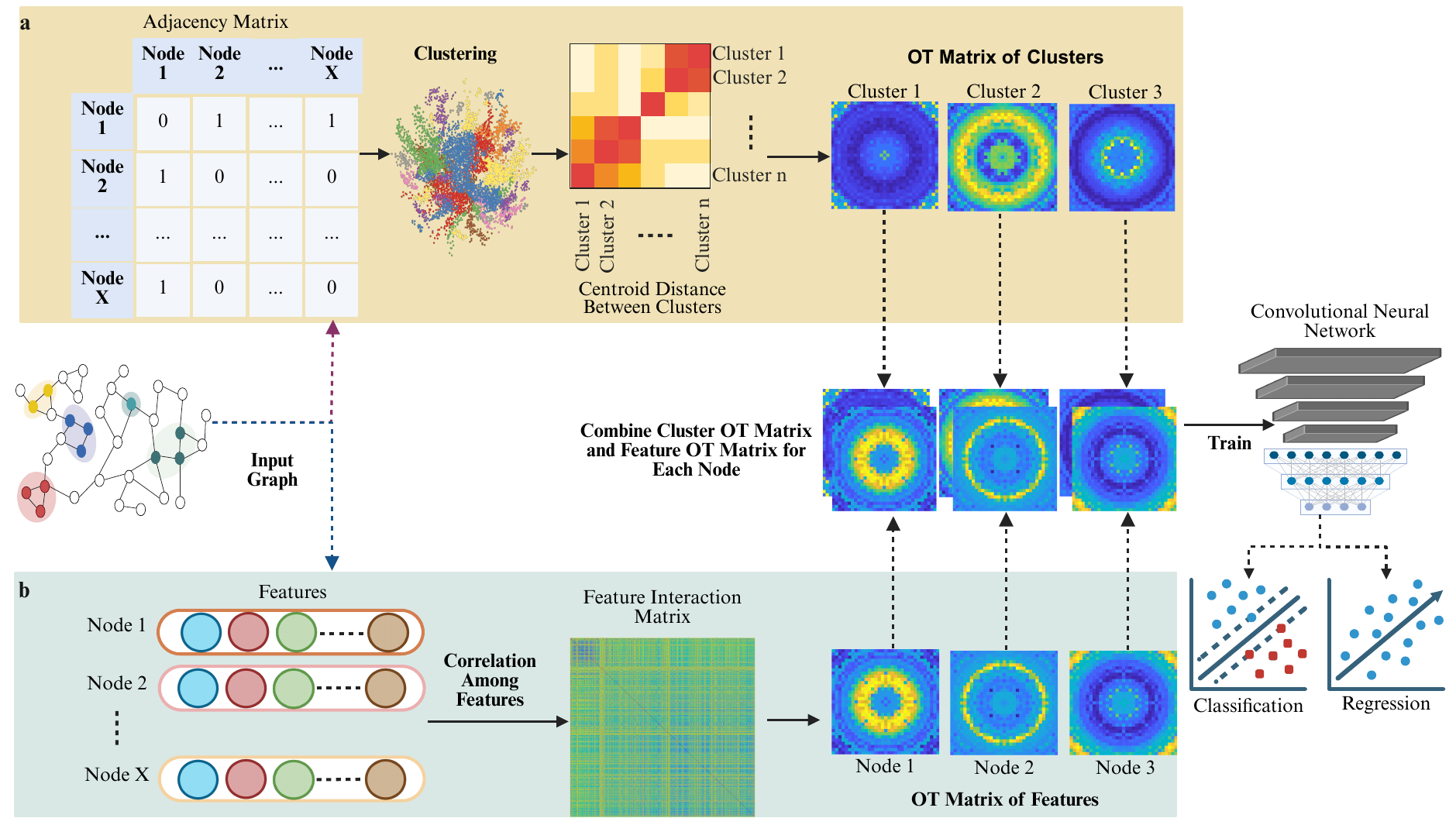}
    \caption{The Graph2Image framework for transforming attributed graphs into multi-channel images. The pipeline learns in two parallel paths. \textbf{a}, The structural path learns the graph's topology. It takes the Adjacency Matrix, uses \textbf{Clustering} to find communities, and maps the centroid distance between clusters on a 2D grid using the \textbf{Optimal Transport (OT)} algorithm to create a structural image. \textbf{b}, The feature path learns node feature relationships. It calculates a feature interaction matrix based on the \textbf{Correlation Among Features} and maps the features on a 2D grid using a different Optimal Transport algorithm to generate feature images. These two images are then combined into a final, multi-channel image for each node. This set of images is used to train a Convolutional Neural Network for downstream tasks like Classification and Regression.}
    \label{figure:graph2image}
\end{figure}

There are several fundamental constraints of the existing biological network analysis methods. Classical methods are often limited by rigid mathematical assumptions that may not capture the full complexity of biological data. On the other hand, modern deep learning (DL) based graph analysis methods also have their limitations~\cite{wu2020comprehensive, hamilton2017representation}. A primary limitation of the DL methods is scalability, as the recursive message-passing central to GNNs is computationally prohibitive for large biological graphs~\cite{hamilton2017inductive}. This recursive computation can also result in degraded analysis performance for two reasons: 1) oversmoothing, where the repeated averaging of neighbor information causes distinct nodes to become indistinguishable~\cite{li2018deeper}, and 2) oversquashing, where important long-range dependencies are lost as information is compressed through informational bottlenecks~\cite{alon2020bottleneck}. Beyond these performance issues, GNNs are also limited by theoretical expressivity bounds, as they are inherently unable to distinguish certain fundamental graph structures, such as different regular graphs or simple cycles of varying lengths~\cite{garg2020generalization, chen2020can}. Perhaps most critically for scientific application, their "black box" nature severely limits interpretability, making it difficult to trust their predictions or use them to uncover new biological mechanisms~\cite{chami2022machine, ying2019gnnexplainer, yuan2022explainability}.

Here we present Graph2Image, a biological network cartography framework that overcomes these limitations by transforming the biological network analysis problem into an image interpretation task. Our approach reframes a large biological network as a collection of two-dimensional semantically meaningful images, one for each node (Fig.~\ref{figure:graph2image}). First, we deconstruct the biological network by clustering it into functional communities. We then employ an optimal transport (OT) algorithm to generate a "map" of these communities, arranging them on a 2D grid such that their spatial proximity reflects their interaction strength in the original network. This process yields a set of images that encode the relational information of the graph in their spatial layout, decoupling the network's complexity and enabling a linearly scalable solution. This transformation allows for the application of a Convolutional Neural Network (CNN) for biological network analysis. By leveraging properties like global receptive fields, CNN is not bound by the same local aggregation constraints as GNNs, offering a solution that is both memory-efficient and adept at capturing long-range context. Furthermore, the image-based representation provides a natural framework for multimodal biological data integration and enhances interpretability through direct visualization.

We demonstrate the efficacy of Graph2Image on a wide range of large-scale, multi-omics cancer datasets, showing that our approach consistently outperforms established GNN-based methods in complex classification tasks. The features learned by our model are biologically meaningful and result in interpretable, actionable insights. We also show that Graph2Image is highly scalable, capable of analyzing massive graphs that are infeasible for traditional GNNs. By providing a scalable, interpretable, and multimodal-ready solution, Graph2Image offers a new paradigm for network analysis, creating new opportunities for disease diagnosis and the study of complex biological systems.

\section{Results}
\subsection{Graph2Image recovers tissue-of-expression programs in a pathway-derived interactome}
\begin{figure}[pth]
    \centering
    \includegraphics[width=0.95\linewidth, page=2]{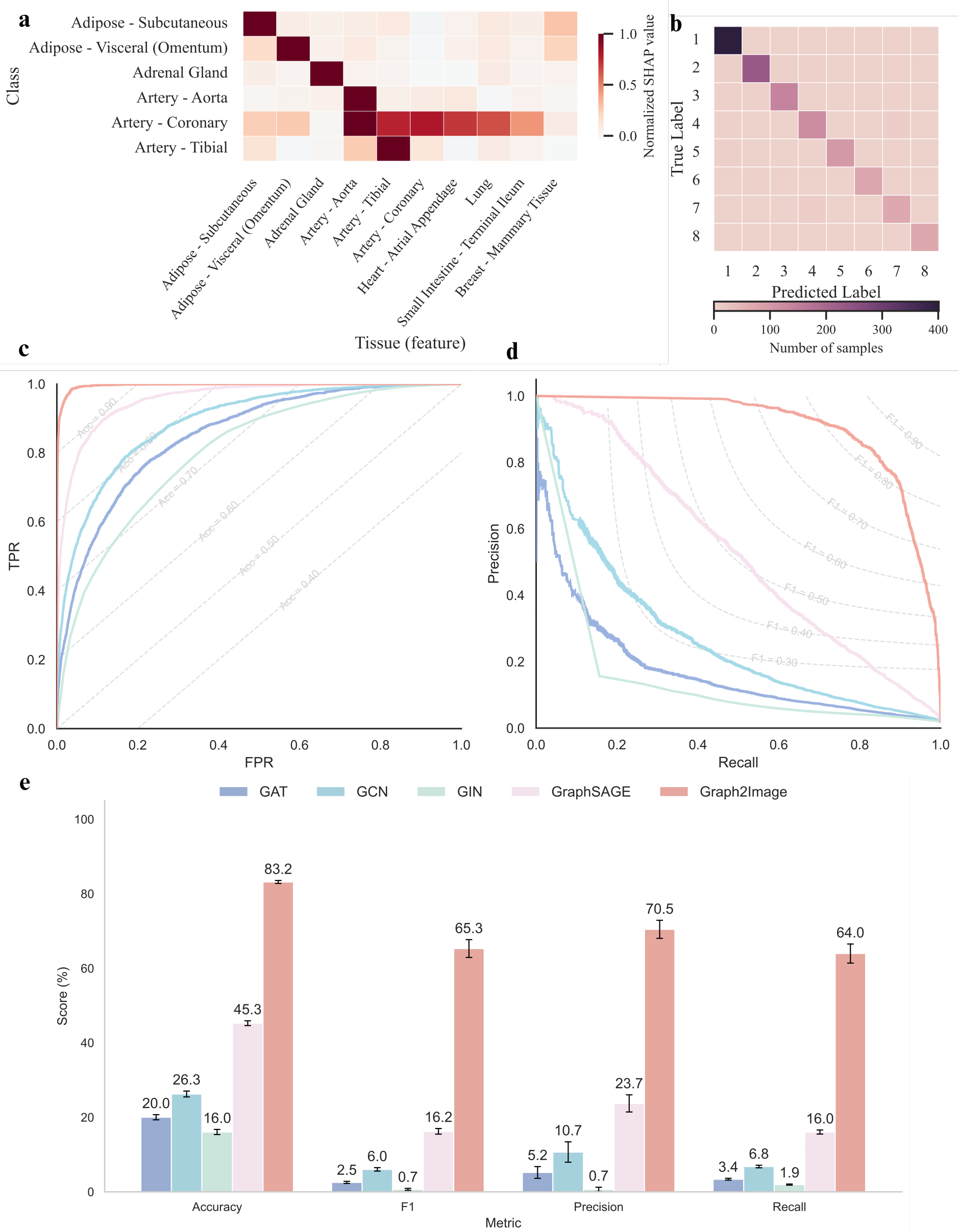}
    \caption{Performance and tissue-level interpretability on the PP-Pathways dataset.
    \textbf{a}, Class-averaged SHAP heatmap showing normalized contributions of selected GTEx tissues (features) to Graph2Image predictions for eight dominant-tissue classes (rows). 
    \textbf{b}, Confusion matrix of Graph2Image predictions across the eight tissue classes (counts per cell). 
    \textbf{c}, Macro-averaged ROC curves comparing Graph2Image with GNN baselines (GAT, GCN, GIN and GraphSAGE). 
    \textbf{d}, Macro-averaged precision–recall curves for the same methods. 
    \textbf{e}, Summary comparison of Accuracy, macro F1, Precision and Recall between Graph2Image and the GNN baselines. 
    See Supplementary Figs. ~\ref{supp_fig:ppathway_confusion}, \ref{supp_fig:ppathway_shap_heatmap}, \ref{supp_fig:ppathway_dendrogram}, and \ref{supp_fig:ppathway_class_dendogram} for the full 54-tissue confusion matrix, SHAP clustermap, tissue-level SHAP dendrograms, and class dendrogram.}
    \label{figure:PPathways_Dataset}
\end{figure}

To demonstrate the superior performance of the proposed Graph2Image method in comparison to existing approaches, we first analyzed pathway-derived protein–protein interaction networks from the PP-Pathways human interactome, which includes experimentally measured tissue labels \cite{pppathways_snap}. PP-Pathways aggregates physical protein–protein interactions from multiple curated pathway resources into a large, heterogeneous network of human proteins, providing a systems-level view of signaling and metabolic pathways. For each node (gene), we constructed features from GTEx v8 bulk RNA-seq by averaging the Transcripts Per Million (TPM) across samples from the same tissue\cite{gtex_science20,gtex_portal}. We then assigned a biologically interpretable label by taking the tissue in which that gene had the highest mean expression across the 54 GTEx tissues, yielding a 54-class tissue-of-expression prediction task (Table~\ref{supp_tab:pppathways_classes})\cite{gtex_science20,uhlen2015proteome}. We highlight eight well-sampled and physiologically distinct tissues, i.e., subcutaneous and visceral adipose tissue, adrenal gland, aorta, coronary and tibial arteries, small intestine (terminal ileum), and breast mammary tissue, in the SHAP heatmap and reduced confusion matrix in Fig.~\ref{figure:PPathways_Dataset}a,b, while the model is trained and evaluated on all 54 tissues (full confusion matrix in Supplementary Fig.~\ref{supp_fig:ppathway_confusion}; see also Supplementary Table~\ref{supp_tab:pppathways_classes} for class names, and Fig.~\ref{supp_fig:class_distribution} for class distribution)~\cite{gtex_portal}. These tissues collectively span key cardiometabolic and endocrine axes: adipose depots act as metabolically active endocrine organs that secrete adipokines and regulate systemic energy balance\cite{kershaw2004,ahima2000}, the adrenal gland produces corticosteroids and catecholamines that coordinate stress and metabolic responses\cite{dutt2023}, large arteries and coronary vessels represent central and peripheral components of the cardiovascular system\cite{chaudhry2022,ccf_heartflow}, the terminal ileum is a major site of nutrient and bile acid absorption\cite{basile2023,collins2024}, and the mammary gland contains secretory epithelium specialized for milk production during lactation\cite{kobayashi2023,shennan2000}.

As seen in Fig.~\ref{figure:PPathways_Dataset}c–e, Graph2Image achieved substantially higher classification performance on this 54-way tissue-of-expression task than all GNN baselines. Across accuracy, macro F1-score, precision, and recall, Graph2Image consistently outperformed GAT, GCN, GIN, and GraphSAGE by large margins. For example, Graph2Image reached an accuracy of 83.2\% and a macro F1-score of 65.3\% when distinguishing all 54 tissues, whereas the best-performing GNN baseline (GraphSAGE) reached 45.3\% accuracy and 16.2\% macro F1. The precision–recall and ROC curves in Fig.~\ref{figure:PPathways_Dataset}c,d further highlight this gap: the Graph2Image curves dominate those of all GNNs across almost the entire range of recall and false-positive rates, indicating more reliable decision boundaries for both high-precision and high-recall operating points. The confusion matrix for Graph2Image predictions in Fig.~\ref{figure:PPathways_Dataset}b shows a strong diagonal structure with only modest off-diagonal mass (full 54-tissue confusion matrix in Supplementary Fig.~S4), suggesting that misclassifications are relatively rare and are largely concentrated between closely related tissues such as subcutaneous vs visceral adipose or between different arterial beds, which are known to share overlapping vascular gene-expression programs\cite{chaudhry2022,gtex_science20}.

Furthermore, as shown in Fig.~\ref{figure:PPathways_Dataset}a, we used SHapley Additive exPlanations (SHAP) on the node images to understand which tissue features drive each class decision. This stands in contrast to GNNs, where applying attribution methods remains a significant challenge~\cite{ying2019gnnexplainer}. While GNN-specific approaches like GNNExplainer have been developed, formally adapting principled methods like SHAP to the graph domain is an active and complex area of research, often requiring specialized approximations to remain computationally feasible~\cite{ying2019gnnexplainer, duval2021graphsvx, jin2022orphicx}. In this setting, SHAP operates on the tissue-expression channel of the Graph2Image representation and yields per-tissue contribution scores for every gene\cite{ying2019,duval2021,lin2022}. For each class (dominant tissue-of-expression), Graph2Image learns a compact set of tissue features with high positive SHAP values that closely match the GTEx label definition (see Supplementary Fig.~\ref{supp_fig:ppathway_shap_heatmap}). Subcutaneous and visceral adipose genes are predominantly supported by their respective adipose tissue features, with secondary support from each other, consistent with the view that these depots share core adipocyte programs but differ in inflammatory and endocrine signatures\cite{kershaw2004,ahima2000}. Artery-labeled genes (aorta, coronary, tibial) show the strongest SHAP contributions from the corresponding arterial tissues, again with cross-support among arterial beds, mirroring the shared smooth-muscle and endothelial expression programs that underpin systemic vascular function\cite{chaudhry2022,gtex_science20}. Genes whose dominant expression lies in the adrenal gland or terminal ileum are primarily driven by their matching tissues, reflecting the specialized endocrine role of the adrenal cortex and medulla in hormone production\cite{dutt2023,adrenal_cancer_ca} and the role of the distal small intestine in nutrient and bile acid absorption\cite{basile2023,collins2024,duca2021}. Finally, breast mammary tissue genes are most strongly influenced by the mammary feature, consistent with the highly specialized transcriptional program of mammary epithelial cells for milk synthesis and secretion\cite{kobayashi2023,shennan2000}.

To examine the broader structure of these feature contributions, we aggregated SHAP scores across all 54 GTEx tissues and performed hierarchical clustering on tissues and classes. The tissue-level dendrogram (Supplementary Fig.~\ref{supp_fig:ppathway_dendrogram}) reveals modules that closely track known organ systems from transcriptomic and proteomic atlases\cite{uhlen2015proteome, gtex_science20}: multiple brain regions cluster together, gastrointestinal tissues (stomach, small intestine, colon) form a coherent group, arterial tissues and heart chambers cluster along a cardiovascular branch, and adipose depots and the mammary gland are placed in close proximity, consistent with their shared involvement in lipid storage, endocrine signaling, and secretory function\cite{kershaw2004,shennan2000,duca2021}. The SHAP clustermap of classes and tissues in supplementary Figs.~\ref{supp_fig:ppathway_dendrogram}, \ref{supp_fig:ppathway_class_dendogram} further shows that genes assigned to different adipose depots, arteries, and gut tissues share partially overlapping but distinct SHAP profiles, capturing both shared pathway usage and tissue-specific refinements. Class-level hierarchical clustering based on SHAP profiles (Supplementary Fig.~\ref{supp_fig:ppathway_class_dendogram}) yields a similar organization, separating vascular, adipose/endocrine, and neural modules in a manner that aligns with system-level patterns seen in GTEx and the Human Protein Atlas\cite{uhlen2015proteome, gtex_science20}. Together, these analyses indicate that Graph2Image not only predicts the dominant tissue-of-expression with high accuracy but also organizes proteins in a feature-importance space that recapitulates major physiological systems and their relationships across the human interactome.

\subsection{Graph2Image recapitulates tissue programs in the HuRI human interactome}
Next, to show the superior capability of  Graph2Image compared to existing methods to recover physiologically meaningful tissue programs directly from a protein–protein interaction network, we analyzed  the Human Reference Interactome (HuRI) annotated with GTEx v8 tissue expression. In this setting, nodes represent genes and edges correspond to experimentally supported human protein--protein interactions from HuRI~\cite{luck2020reference}, and each node is associated with a GTEx gene expression profile across diverse normal tissues~\cite{gtex2017,gtex2020}. We assigned each gene a real biological label equal to its primary GTEx tissue of expression, defined as the tissue with maximal mean TPM across donors~\cite{gtex2017,gtex2020}, and restricted the analysis to nine tissues with sufficient support: whole blood, brain-cerebellum, brain-cerebellar hemisphere, skeletal muscle, thyroid, ovary, testis, EBV-transformed lymphoblastoid cell lines (EBV-LCLs), and cultured fibroblasts. These tissues span major physiological systems-hematopoietic and immune (blood and EBV-LCLs), neuronal (brain regions), contractile muscle, endocrine (thyroid), reproductive (ovary and testis) and stromal/mesenchymal compartments (fibroblasts)~\cite{uhlen2015proteome,kalluri2016fibroblasts,omi2017lcl}. Graph2Image was trained to predict these tissue labels from two-channel images that jointly encode network structure and tissue expression, and we compared its performance against four GNN baselines (GAT, GCN, GIN and GraphSAGE) trained on the same HuRI graph.

\begin{figure}[pt]
    \centering
    \includegraphics[width=0.9\linewidth, page=3]{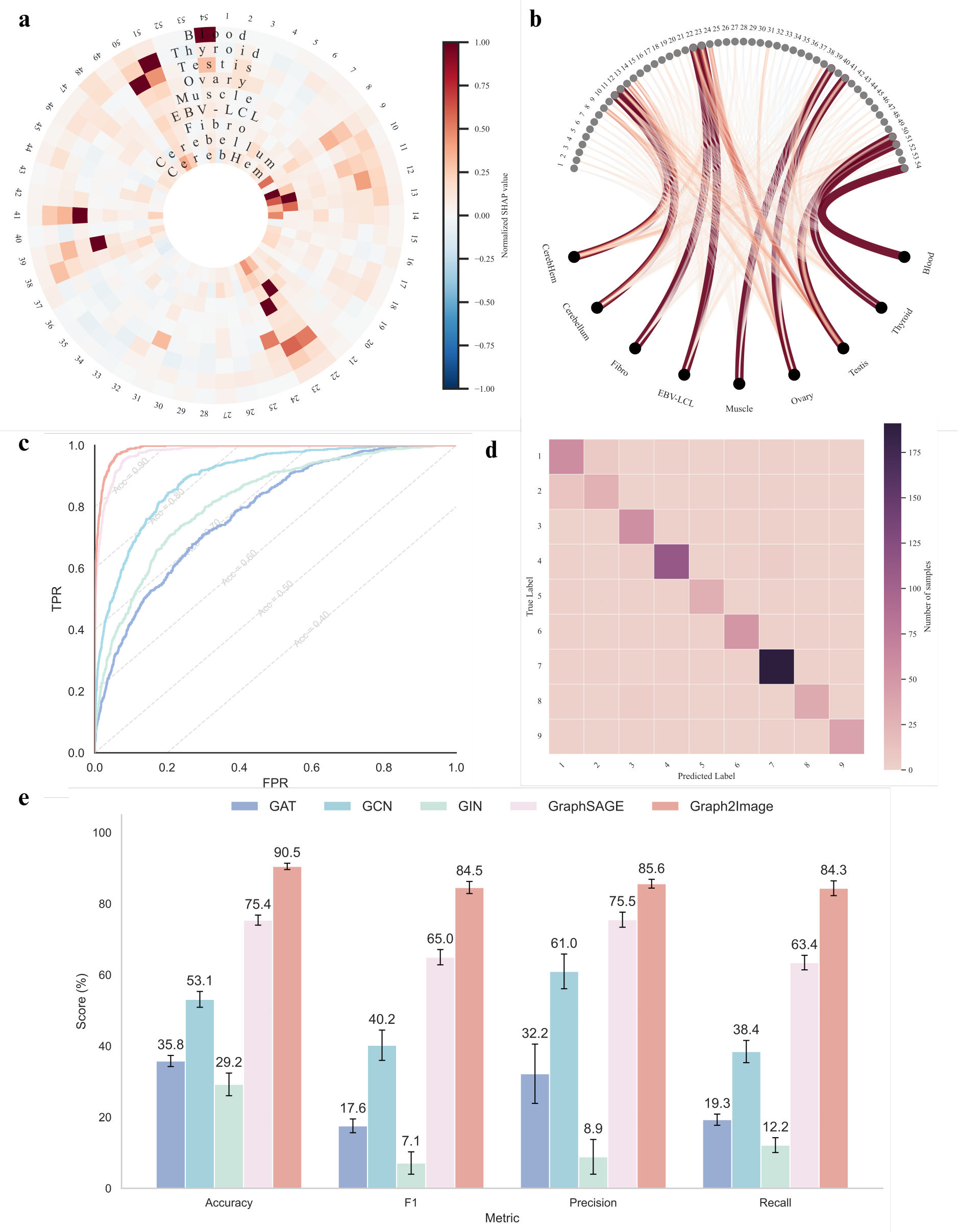}
    \caption{
    Performance and interpretability of Graph2Image on the HuRI + GTEx dataset.
    \textbf{a}, Radial SHAP map summarizing normalized SHAP values for each GTEx tissue feature across the nine primary tissue classes; concentric rings correspond to classes and angular positions to tissues.
    \textbf{b}, SHAP chord diagram highlighting the strongest positive links between tissue classes (bottom) and influential GTEx tissue features (top), illustrating shared and class-specific programs.
    \textbf{c}, Receiver operating characteristic (ROC) curves for Graph2Image and GNN baselines (GAT, GCN, GIN, GraphSAGE) in the nine-way tissue classification task.
    \textbf{d}, Confusion matrix of Graph2Image predictions across the nine primary tissues (counts in linear scale).
    \textbf{e}, Classification comparison (Accuracy, macro F1, macro Precision, macro Recall) between Graph2Image and GNN baselines; bars show mean scores with bootstrap confidence intervals.
    See Supplementary Figs.~\ref{supp_fig:huri_confusion},~\ref{supp_fig:huri_shap_heatmap},~\ref{supp_fig:huri_tissue_dendrogram}, and~\ref{supp_fig:huri_class_dendogram} for the full confusion matrix, SHAP clustermaps, and dendrograms.
    }
    \label{figure:HuRI_Dataset}
\end{figure}

As seen in Fig.~\ref{figure:HuRI_Dataset}e, Graph2Image achieved a classification accuracy of 90.5\% and a macro F1-score of 84.5\%, substantially outperforming all GNN baselines. The best-performing GNN, GraphSAGE, reached 75.4\% accuracy and 65.0\% macro F1, while GAT, GCN and GIN performed markedly worse, particularly on minority tissues. Consistent with these summary metrics, the precision–recall and ROC curves in Fig.~\ref{figure:HuRI_Dataset}c show that Graph2Image dominates the baselines across operating thresholds (see also Supplementary Fig.~\ref{supp_fig:huri_roc_pr}). The confusion matrix of Graph2Image predictions in Fig.~\ref{figure:HuRI_Dataset}d exhibits a strong diagonal structure with only limited off-diagonal mass (se Supplementary Fig.~\ref{supp_fig:huri_confusion} for a detailed confusion matrix), and the remaining errors occur mainly between closely related brain regions or between immune compartments such as whole blood and EBV-LCLs, which are known to share overlapping transcriptional programs and lineage relationships in the hematopoietic system \cite{gtex2017,uhlen2015proteome,omi2017lcl,wroblewski2002lcl}.

To understand what Graph2Image learns from the HuRI interactome and how it uses tissue expression to make predictions, we applied SHAP to the node images, mapping pixel attributions back to GTEx tissues via the optimal-transport–derived layout. The resulting SHAP profiles reveal that Graph2Image has organized the HuRI genes along a set of physiologically meaningful tissue axes that match known systems biology \cite{gtex2017,uhlen2015proteome}. The radial SHAP map in Fig.~\ref{figure:HuRI_Dataset}a shows that each primary tissue class is dominated by its corresponding GTEx tissue feature:  whole blood for blood-labeled genes, skeletal muscle for muscle-labeled genes, and so on, consistent with the strong tissue specificity observed in GTEx and related atlases \cite{gtex2017,gtex2020,uhlen2015proteome} (see Supplementary Fig.~\ref{supp_fig:huri_shap_heatmap}). At the same time, the model exploits systematic combinations of tissues rather than relying on single features. As shown in Fig.~\ref{figure:HuRI_Dataset}a (see Supplementary Figs.~\ref{supp_fig:huri_shap_heatmap} for detailed SHAP heatmap), for genes whose primary label is whole blood, SHAP assigns high positive importance not only to whole blood but also to spleen and EBV-LCLs, reflecting the shared hematopoietic and immune origin of circulating blood cells, splenic immune populations, and EBV-transformed B lymphoblastoid cell lines \cite{janeway2012immunobiology,omi2017lcl,wroblewski2002lcl}. Thyroid-labeled genes are supported by the thyroid feature together with minor salivary gland, ovary, and cortical brain regions, in line with reports that epithelial cells of the salivary gland and thyroid share immune and antigen-presentation programs and that thyroid and reproductive tissues participate in common endocrine and autoimmune axes \cite{fox1986salivaryhla,zhang2023salivarythyroid,lin2025sjogrensthyroid,weigert2025reprohormones}. Genes labeled as ovary show strong SHAP contributions from ovary, cervix (ectocervix) and fallopian tube, capturing a coordinated female reproductive-tract program in which gene expression is shared across these tissues \cite{ulrich2022fallopiantube,lengyel2022fallopianTube,hpa2025fallopian}. Testis-labeled genes receive additional support from the ovary, cerebellar tissues, and immune features, consistent with the tight coupling between the male reproductive tract and the immune system and the immune-privileged status of the testis \cite{hedger2014immunophysiology,li2012testisimmunity}, as well as the broader endocrine and neuroendocrine crosstalk between gonadal and brain tissues \cite{gtex2020,uhlen2015proteome}.

The chord diagram in Fig.~\ref{figure:HuRI_Dataset}b summarizes these relationships by linking each class to its most influential GTEx tissues. Brain-associated classes (cerebellum and cerebellar hemisphere) are predominantly connected to brain tissues, including cortex, frontal and cingulate cortex, and cerebellar regions, highlighting a coherent central nervous system module \cite{gtex2017,uhlen2015proteome}. Muscle-labeled genes are anchored by skeletal muscle but also connect strongly to fibroblasts and vascular or immune tissues, consistent with the well-established role of fibroblasts and immune cells in providing stromal, vascular, and inflammatory support within muscle and other tissues \cite{kalluri2016fibroblasts,janeway2012immunobiology}. Across classes, fibroblasts and lymphocyte-related features (EBV-LCL and whole blood) appear as recurrent high-SHAP channels, suggesting that Graph2Image uses a broadly expressed stromal and immune backbone as a reference frame against which more specialized tissue signatures are contrasted \cite{kalluri2016fibroblasts,janeway2012immunobiology,gtex2017}.

To formally quantify this structure, we performed hierarchical clustering on the tissue-level SHAP profiles as shown in Supplementary Fig.~\ref{supp_fig:huri_tissue_dendrogram}. The tissue dendrogram groups central nervous system tissues together, clusters blood, spleen, and lymphocyte-derived samples into an immune super-group, and places muscle, fibroblast-rich, and epithelial/endocrine tissues into interconnected stromal and organ-specific branches \cite{gtex2017,uhlen2015proteome}. A complementary SHAP clustermap in Supplementary Fig.~\ref{supp_fig:huri_shap_heatmap} and class-level dendrogram in Supplementary Fig.~\ref{supp_fig:huri_class_dendogram} reveal that classes sharing similar SHAP signatures, e.g., the two cerebellar labels or the reproductive tissues, cluster closely in this importance space \cite{gtex2017,uhlen2015proteome,ulrich2022fallopiantube,lengyel2022fallopianTube}. Thus, the HuRI analysis shows that Graph2Image not only achieves superior predictive performance on a realistic tissue-of-expression task, but also learns a structured, interpretable representation in which major tissue systems and their relationships emerge directly from the interactome.

\subsection{Graph2Image accurately classifies diverse cell types in a whole-organism atlas}
\begin{figure}[pt]
    \centering
    \includegraphics[width=\linewidth, page=4]{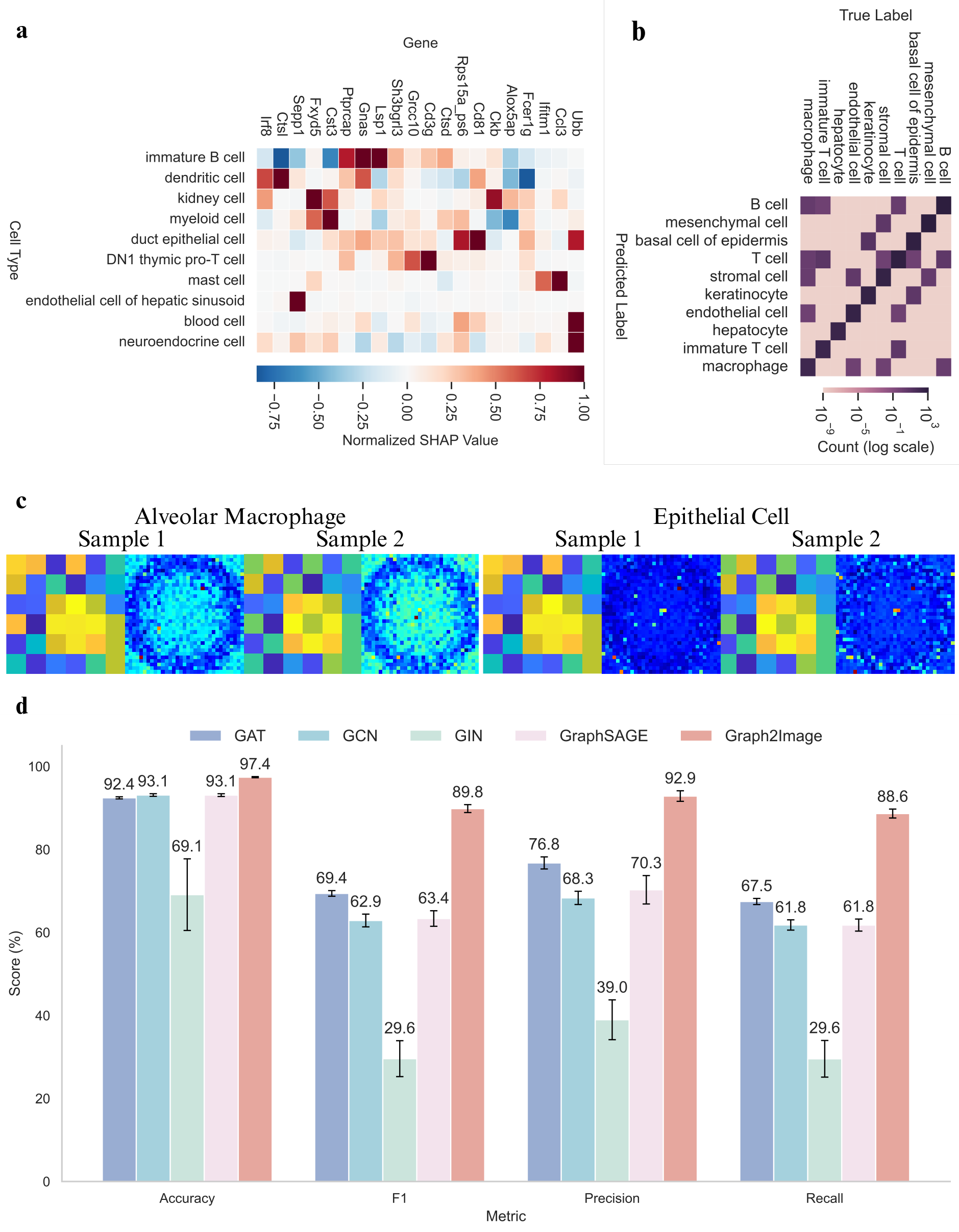}
    \caption{Performance and interpretability on the Tabula Muris dataset.
    \textbf{a}, Class-averaged SHAP heatmap for top marker genes across representative cell types; values are normalized SHAP scores.
    \textbf{b}, Confusion matrix of Graph2Image predictions across 55 cell types (counts in log scale).
    \textbf{c}, Representative Graph2Image outputs from Tabula Muris with samples from Alveolar Macrophage and Epithelial Cell.
    \textbf{d}, Classification comparison (Accuracy, F1, Precision, Recall) between Graph2Image and GNN baselines (GAT, GCN, GIN, GraphSAGE).
    See Supplementary Figs. \ref{supp_fig:tm_confusion}, \ref{supp_fig:tm_dendrogram}, and \ref{supp_fig:tm_shap_heatmap} for full-resolution confusion matrix, dendrogram and comprehensive heatmap.}
    \label{figure:TM_Dataset}
\end{figure}

To demonstrate that our approach can analyze complex, large-scale biological systems with better accuracy and computational efficiency than existing methods, we applied Graph2Image to the Tabula Muris single-cell RNA-sequencing dataset. The resulting graph from the data comprises 54,864 nodes representing cells and over 3 billion edges, a scale that is intractable for standard GNNs. To create a manageable baseline, we were required to threshold the graph and reduce the number of edges by over 99\% to approximately 5.1 million. The task was to classify the nodes representing single cells into 55 different cell types (See Fig.~\ref{supp_fig:class_distribution}c for class distribution). As seen in Fig.~\ref{figure:TM_Dataset}d, compared to existing methods,  Graph2Image, demonstrated a substantial improvement. Note that Graph2Image could operate on the complete graph, whereas the existing methods could be only applied on downsampled graphs. Graph2Image achieved a classification accuracy of 97.4\% and a macro F1-score of 89.8\%, significantly outperforming all benchmark GNNs by at least 4.3\%. These classification results are also shown using the confusion matrix in Fig. \ref{figure:TM_Dataset}b (for detailed confusion matrix see Fig.~\ref{supp_fig:tm_confusion}), where we can see a strong diagonal concentration, indicating minimal misclassification even among closely related cell types.

Beyond these summary metrics, inspection of the Graph2Image outputs confirms that the transformation produces class-specific visual signatures. From Fig.~\ref{figure:TM_Dataset}c we can see that all Alveolar Macrophages are mapped to images with a highly consistent textural pattern, whereas Epithelial Cells give rise to a clearly distinct pattern. Within each cell type, node-specific images are visually similar across samples, but between cell types they differ markedly in both the structural and feature channels, indicating that the learned spatial organization of the images reflects underlying cellular identity rather than an arbitrary encoding.

To show that our approach is more interpretable and to reveal the biological principles learned by the model, we used SHAP to quantify the contribution of each gene to the classification of each cell type on the constructed node images. Leveraging the straightforward applicability of SHAP to our framework, in Fig.~\ref{figure:TM_Dataset}a we found that the model learned a set of biologically meaningful gene expression signatures that correctly recapitulate the fundamental principles of cell lineage and identity (see Fig.~\ref{supp_fig:tm_shap_heatmap} for a detailed heatmap). The most prominent clusters found in our study correspond to hematopoietic, epithelial, and mesenchymal lineages. The hematopoietic cluster, for instance, is known to have a high level of expression of PTPRC (encoding the pan-leukocyte marker CD45) \cite{hermiston2003cd45}. Within this hematopoietic group, the model precisely recapitulated the fundamental lymphoid-myeloid dichotomy. Our SHAP analysis revealed that this was achieved by learning two distinct and opposing gene signatures that defined the two lineages. The B-cell lineage is shown to have high positive SHAP values for Vpreb3, Cd79a, and Cd79b, genes encoding critical components of the B-cell receptor complex \cite{karasuyama1994pre}. Moreover, T-cells have higher SHAP values of hallmark genes like Cd3g, a subunit of the T-cell receptor \cite{clevers1988t}. The myeloid lineage is characterized by a distinct gene expression signature, which includes key markers such as Lyz1 for monocytes and Mpo for granulocytes \cite{faurschou2003neutrophil}. The model also identified key regulators, such as the transcription factor Irf8, which showed high importance for dendritic cells and monocytes, consistent with its known role in myeloid development \cite{tamura2002icsbp}. The model also robustly learned the distinction between epithelial and mesenchymal cells. The expression of Epcam, a classic epithelial adhesion molecule, was a powerful positive feature for epithelial cells \cite{trzpis2007epithelial}, while its absence was key for identifying mesenchymal cells like fibroblasts, which were instead defined by markers such as Vim (Vimentin) \cite{franke1978different}. 

Finally, the analysis highlights the model's ability to identify highly specific biomarkers for terminally differentiated cell types. The most striking example is the gene Alb (Albumin), whose expression was learned as an overwhelmingly powerful and exclusive positive predictor for hepatocytes, reflecting its primary role as the main protein produced by the liver \cite{rothschild1972albumin}. Similarly, the model correctly identified Sftpc (Surfactant Protein C) as the definitive marker for type II pneumocytes, essential for lung function \cite{weaver2001function}. To formally assess this learned structure, we performed hierarchical clustering on the cell types based on their SHAP score profiles. The resulting dendrogram in Fig.~\ref{supp_fig:tm_dendrogram} reveals a clear, high-level separation of major cell lineages that closely mirrors known developmental biology, demonstrating that the model's feature importance space is structured to reflect both broad, lineage-defining gene programs and the highly specific biomarkers that define terminal cell fates.

\subsection{Multi-omic analysis in pan-cancer classification reveals diverse molecular drivers}
To evaluate the ability of Graph2Image to integrate large-scale multi-omic profiles in a clinically heterogeneous setting, we applied it to the TCGA pan-cancer network, comprising 8,314 nodes connected by over 69 million edges. Whereas standard GNN baselines were only computationally manageable after aggressive edge down-sampling to ~3.2 million edges, Graph2Image operated directly on the complete, unfiltered graph.

\begin{figure}[pt]
    \centering
    \includegraphics[width=0.95\linewidth, page=5]{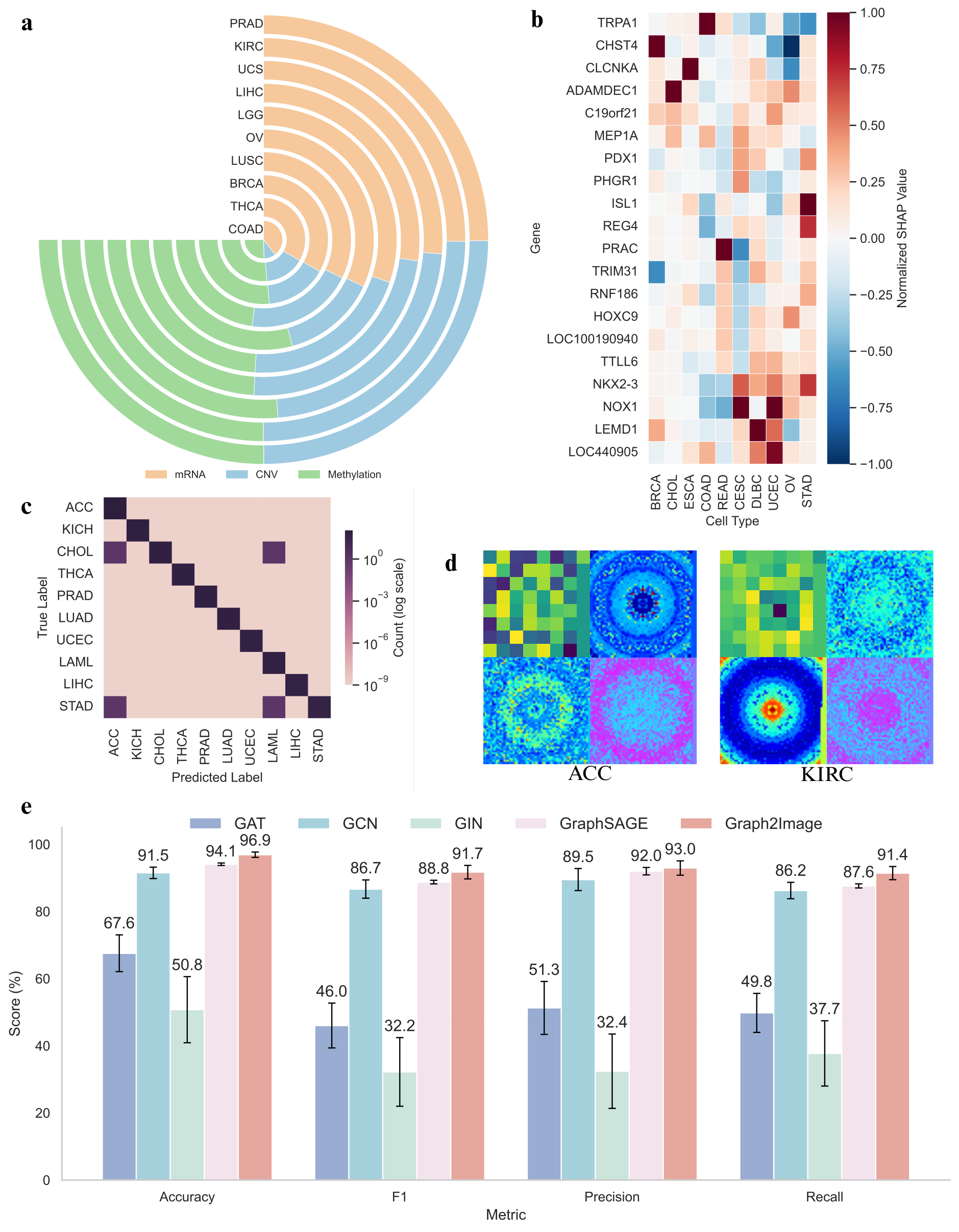}
    \caption{Performance and modality contributions on the pan-cancer cohort.
    \textbf{a}, Relative contribution of each omic layer (mRNA, CNV, DNA methylation) per cancer type, computed by aggregating SHAP importance.
    \textbf{b}, Class-averaged SHAP heatmap for top mRNA features across cancer types (normalized SHAP scores).
    \textbf{c}, Confusion matrix of Graph2Image predictions across 32 TCGA cancer types (log-scale counts).
    \textbf{d}, Representative Graph2Image outputs from the Pan Cancer dataset where images corresponds to different omics types of Adrenocortical carcinoma (ACC) and Kidney renal clear cell carcinoma (KIRC).
    \textbf{e}, Classification comparison (Accuracy, F1, Precision, Recall) between Graph2Image and GNN baselines.
    See Supplementary Figs. \ref{supp_fig:pan_confusion}, \ref{supp_fig:pan_shap_heatmap}, \ref{supp_fig:pan_modal_contrib}, and \ref{supp_fig:pan_dendrogram} for extended confusion matrices, heatmaps, modality breakdowns and clustering analyses.}
    \label{figure:Pan_Dataset}
\end{figure}

The task was to classify nodes into 32 distinct cancer types (see Fig.~\ref{supp_fig:class_distribution}d for class distribution). As seen from Fig. \ref{figure:Pan_Dataset}e, our model again demonstrated superior performance. It achieved 96.9\% accuracy and an F1-score of 91.7\%. Our technique significantly outperformed the next-best GNN models (by 2.8\% in accuracy and 2.9\% in F1-score), which were trained on the sparse graph. These classification results are also depicted using the confusion matrix in Fig.~\ref{figure:Pan_Dataset}c and Fig.\ref{supp_fig:pan_confusion}. The matrix shows a strong diagonal concentration, indicating minimal misclassification even among histologically similar cancer types like Colon (COAD) and Rectal Adenocarcinoma (READ) (see Supplementary Table~\ref{supp_tab:pan_label_abbreviatio} for full name of the cancer type abbreviations).

To reveal the biological principles learned by our model, we used SHAP to analyze the most impactful mRNA features. We visualize these SHAP scores as a clustered heatmap in Fig.~\ref{figure:Pan_Dataset}b (see Supplementary Fig.~\ref{supp_fig:pan_shap_heatmap} for detailed heatmap). From Fig.~\ref{figure:Pan_Dataset}b, we observe that the model's decisions are driven by distinct and biologically coherent gene signatures. The model's interpretability also extends to uncovering the biological relationships between the cancer types. To demonstrate this, we performed hierarchical clustering on the 32 cancers using their learned mRNA SHAP score profiles (Fig.~\ref{supp_fig:pan_dendrogram}). This analysis revealed that the model automatically grouped cancers according to their known developmental origins without any prior information. For example, the heatmap clearly shows a prominent cluster of all squamous cell carcinomas (HNSC, LUSC, CESC), which the model identified using a shared set of predictive genes. Similarly, the clustering correctly grouped cancers of neuroendocrine origin (PCPG, THYM) and gastrointestinal adenocarcinomas (COAD, READ, STAD). The resulting similarity relationships align well with established cancer phylogeny and tissue-of-origin patterns \cite{hoadley2014multiplatform}.

In Fig.~\ref{figure:Pan_Dataset}a we show that our framework can also deconstruct the contributions of different molecular modalities. We aggregated the total SHAP importance from all three omic layers, i.e., mRNA, copy number variation (CNV), and DNA methylation. This allowed us to quantify the relative contribution of each modality to the classification decisions (see Fig.~\ref{supp_fig:pan_modal_contrib} for detailed contribution map). This analysis reveals that the model learned to prioritize different modalities for different cancer types, reflecting their diverse biology. For example, mRNA expression was the dominant factor for cancers with a strong lineage-specific identity. The predictions for Liver Hepatocellular Carcinoma (LIHC), Prostate Adenocarcinoma (PRAD), and Thyroid Carcinoma (THCA) were almost entirely dependent on tissue-defining genes like ALB, KLK3 (PSA), and TG, respectively. In contrast, the model identified CNV as a major factor for cancers known for genomic instability. The classification of Ovarian Cancer (OV) and Uterine Carcinosarcoma (UCS) showed a high reliance on CNV features. This aligns with the known prevalence of gene amplifications and deletions in these diseases \cite{tcga2011ovarian}. Furthermore, the model identified DNA methylation as a critical predictive layer for specific cancer types. As an example, the classification of Low-Grade Glioma (LGG) showed a substantial contribution from methylation features. This finding reflects the central role of the CpG Island Methylator Phenotype (G-CIMP), which is a defining characteristic of a major LGG subtype and is strongly associated with IDH1 mutations \cite{noushmehr2010identification}.

Consistent with these modality-specific attributions, Graph2Image examples from the pan-cancer cohort display reproducible, class-specific visual patterns in Fig.~\ref{figure:Pan_Dataset}d. Tumors of the same type give rise to highly similar multi-channel images, whereas tumors from different types exhibit noticeable differences in both the structural and feature channels. For instance, ACC and KIRC images show distinct arrangements of concentrated, ring-like signals versus more diffuse textures across channels, reflecting their divergent molecular architectures. This visual separability highlights that the Graph2Image representation encodes cancer-type–specific structure rather than arbitrary pixel patterns.

\subsection{Graph2Image accurately classifies primary versus metastatic prostate cancer}
\begin{figure}[pt]
    \centering
    \includegraphics[width=\linewidth, page=6]{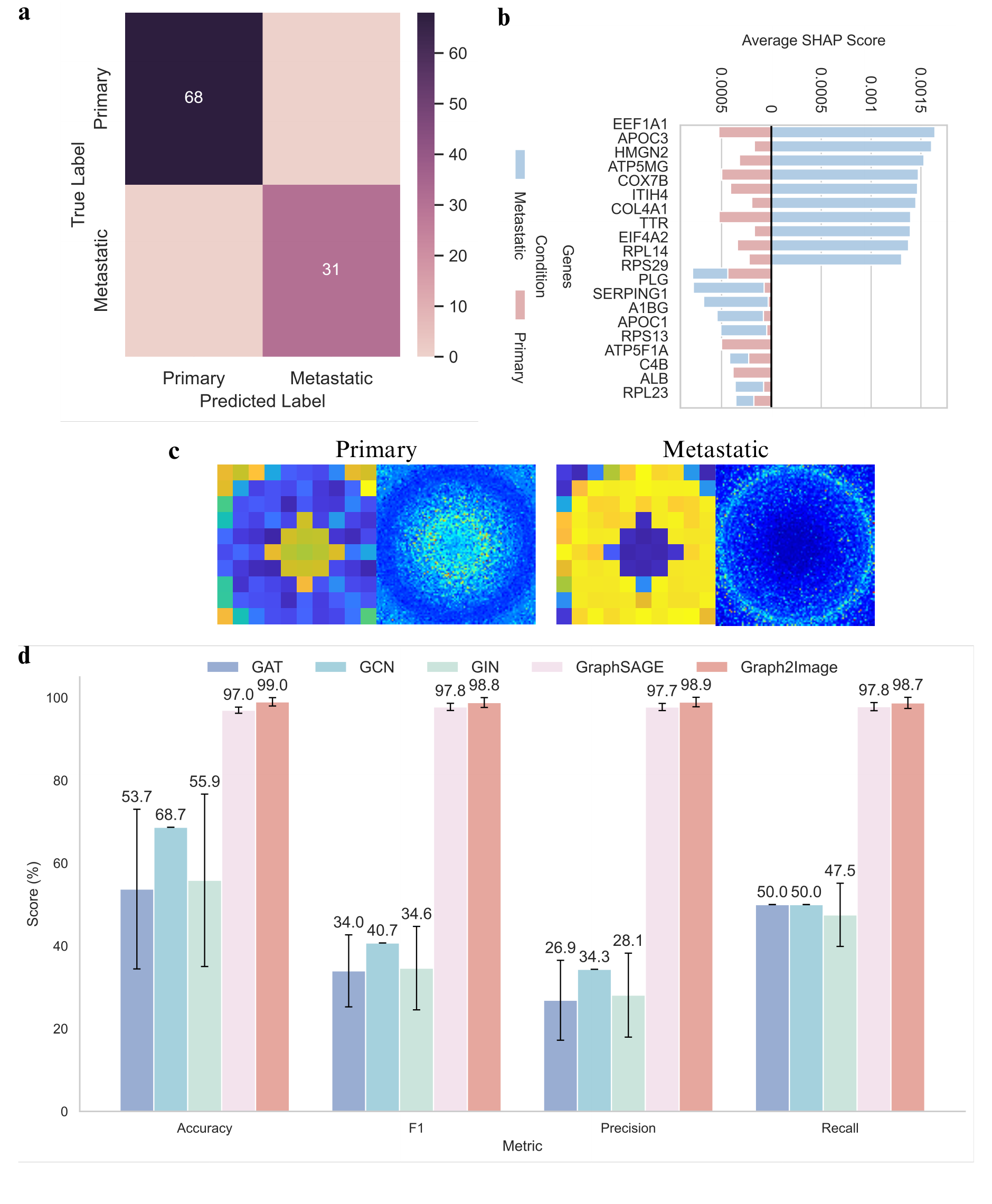}
    \caption{Primary versus metastatic prostate cancer classification and gene-level attributions.
    \textbf{a}, Confusion matrix for Graph2Image predictions on the Prostate Cancer cohort.
    \textbf{b}, Class-averaged SHAP scores for the most predictive genes, with directionality shown for metastatic (blue) and primary (red) classes.
    \textbf{c}, Representative Graph2Image outputs from the Prostate Cancer dataset where images corresponding to Primary and Metastatic tumors.
    \textbf{d}, Classification comparison (Accuracy, F1, Precision, Recall) between Graph2Image and GNN baselines.
    See Supplementary Figs.~\ref{supp_fig:pnet_butterfly} and ~\ref{supp_fig:pnet_gene_dendrogram} for gene-level distributions and clustering of predictive features.}
    \label{figure:Pnet_Dataset}
\end{figure}

To test our model's applicability in clinical settings, we applied it to the Prostate Cancer dataset. The task was to classify primary versus metastatic prostate cancer tumors. From Fig.~\ref{figure:Pnet_Dataset}d, Graph2Image achieved an accuracy of 99.0\% and an F1-score of 98.8\%, significantly outperforming benchmark GNNs. The second best method achieved an accuracy of 97.0\% and F-1 score of 97.8\%. The confusion matrix in Fig.~\ref{figure:Pnet_Dataset}a also shows the model's near-perfect classification, which establishes its robust capability to identify the molecular signatures that differentiate these two critical disease states. Again, we used SHAP analysis and in Fig. \ref{figure:Pnet_Dataset}b showed that the model learned two distinct and biologically coherent molecular signatures corresponding to the different disease states (see Fig.~\ref{supp_fig:pnet_butterfly} for detailed map). The features most predictive of metastatic classification point toward a phenotype of heightened metabolic activity, increased protein synthesis, and active tissue remodeling. These are all hallmarks of aggressive, invasive cancer \cite{hanahan2011hallmarks}.

In Fig. \ref{figure:Pnet_Dataset}b, Graph2Image's top-ranked feature was EEF1A1. This is a crucial translation elongation factor whose overexpression is essential to sustain the high rate of protein synthesis required by rapidly proliferating cancer cells \cite{abbas2015eef1a}. This finding was supported by the high importance of other translation machinery components, such as EIF4A2 and multiple ribosomal proteins (e.g., RPL14, RPL7A). Furthermore, the model identified a signature of metabolic reprogramming. It heavily weighted genes involved in mitochondrial energy production (ATP5MG, COX7B) and altered lipid metabolism (APOC3) \cite{cairns2011regulation}. This anabolic signature was complemented by features essential for physical invasion, such as COL4A1. This gene is critical for remodeling the extracellular matrix during metastatic dissemination \cite{kalluri2009basement}. Conversely, the features predictive of primary, non-metastatic tumors reflected a more stable and differentiated cellular identity. The model correctly identified KLK3 as a key feature for this class. This gene encodes the Prostate-Specific Antigen (PSA) and is consistent with its role as the definitive clinical biomarker for prostate tissue \cite{lilja1985kallikrein}. This was reinforced by other markers of epithelial lineage, such as the cytokeratins KRT8 and KRT18. The model also associated a signature of stable cellular function with primary disease. It highlighted core "housekeeping" genes like RPS27 and RPS23. Their stable expression is indicative of a less transformed state, as these proteins are also involved in P53 tumor suppressor pathways \cite{deisenroth2010ribosome}. Finally, the model highlighted the importance of GPX1, which encodes the antioxidant enzyme glutathione peroxidase 1 \cite{chu2004role}. This suggests the model learned that having a working system to manage oxidative stress is a key feature of less aggressive, localized tumors.

To formally assess whether the model organized these predictive genes into coherent biological group, we performed hierarchical clustering on the top predictive genes based on their SHAP profiles shown in Fig.~\ref{supp_fig:pnet_gene_dendrogram}. From Fig.~\ref{supp_fig:pnet_gene_dendrogram}, we see that the model groups genes into distinct functional modules. For example, the largest cluster was highly enriched for ribosomal proteins and translation factors. This demonstrates that the model's feature importance space is organized according to the biological functions that differentiate primary from metastatic disease.

Visual inspection of the Graph2Image outputs for the Prostate Cancer cohort further supports this separation. Primary tumors and metastatic lesions form two clearly distinct families of images: within each class, the structural channel and accompanying feature channels are highly consistent across patients, but the overall visual appearance differs sharply between the two disease states (Fig.~\ref{figure:Pnet_Dataset}c). Primary tumors exhibit a stable, symmetric pattern, whereas metastatic lesions display altered, more intense and spatially redistributed signals, mirroring the underlying shift toward a highly proliferative and invasive phenotype. These class-specific visual signatures provide an intuitive, image-based view of the molecular transition from localized to metastatic disease.

\section{Discussion}
In this work, we introduced Graph2Image, a biological graph cartography framework that overcomes the critical limitations of scalability and interpretability inherent in current biological network analysis methods. Our approach is highly scalable and have shown to be able to process the complete Tabula Muris cell atlas, a biological network with 54,865 nodes and over 3 billion edges, at a scale where leading GNN models failed. By reframing biological network analysis as an image recognition task, we achieve state-of-the-art performance across multiple biological contexts, including 99.0\% accuracy in distinguishing metastatic from primary prostate cancer. More importantly, our results show that this transformation provides a powerful new lens for scientific discovery, as the model automatically recapitulated principles of developmental biology, acted as a "molecular pathologist" to identify cancer drivers, and learned the molecular signatures of tumor progression. Our model does more than just predict an outcome. It helps us understand the biology behind its decisions. This leads to new scientific discoveries.

In the two tissue-of-expression tasks, Graph2Image shows that the same cartographic machinery can recover stable tissue programs directly from pathway-derived and reference interactomes. On the PP-Pathways network, where nodes are proteins linked by curated pathway interactions and labeled by their dominant GTEx tissue-of-expression, the model accurately predicts the tissue label and, through SHAP, reconstructs a hierarchy of organ systems: adipose depots, arterial beds, adrenal gland, distal small intestine and mammary gland are each supported by coherent sets of tissue features, and their SHAP-based clustering recapitulates cardiovascular, endocrine and gastrointestinal modules that agree with transcriptomic and proteomic atlases~\cite{gtex_science20,uhlen2015proteome}. This shows that Graph2Image can use pathway connectivity plus bulk expression to rediscover large-scale physiological structure without being explicitly told about organ systems.

A complementary picture emerges from the HuRI interactome, which is constructed from high-throughput protein–protein interaction screens rather than pathway aggregation~\cite{luck2020reference}. Here, Graph2Image again learns tissue-of-expression labels with high accuracy and organizes genes along interpretable tissue axes: immune- and blood-labeled genes are supported by a shared hematopoietic module, brain-labeled genes align with a central nervous system module, and reproductive tissues (ovary and testis) cluster with related endocrine and neuroendocrine features~\cite{gtex2017,gtex2020,uhlen2015proteome}. Across PP-Pathways and HuRI, the SHAP-based dendrograms reveal consistent grouping of immune, stromal, neural, cardiovascular and endocrine tissues, suggesting that Graph2Image is capturing reproducible tissue programs that are robust to how the underlying interactome is constructed.

A key advantage of our framework is its inherent scalability and interpretability. As we have shown, GNNs often fail when faced with the massive, densely connected graphs typical of biological systems, requiring a significant loss of information through edge thresholding. Graph2Image bypasses this bottleneck by deconstructing the global biological graph analysis into a set of independent, fixed-size image classification problems—one for each node. This transformation makes the computational complexity linear with respect to the number of nodes and entirely sidesteps the recursive message-passing operations that constrain GNNs, enabling the analysis of biological networks at their true scale. Furthermore, our interpretability analysis of the Tabula Muris dataset in Fig.~\ref{supp_fig:tm_dendrogram} demonstrates that the model's learned feature space is hierarchically structured. This structure mirrors known developmental biology. The model learned to use broad, high-level markers like PTPRC to define an entire lineage, while also capturing fundamental sub-lineages, such as the lymphoid-myeloid dichotomy. Finally, it identified highly specific, exclusive markers for terminally differentiated cells, such as Alb for hepatocytes. The model organized the 55 cell types into coherent super-groups, correctly identifying canonical markers for the major hematopoietic, epithelial, and mesenchymal lineages, and recapitulating the lymphoid-myeloid dichotomy. The same machinery underlies the tissue hierarchies we observe in PP-Pathways and HuRI, indicating that Graph2Image learns a coherent taxonomy of both cell types and tissues from very different graphs. This ability to recover biologically meaningful structure from complex, whole-organism and whole-interactome data helps to validate the model's internal logic and moves beyond traditional "black box" models by providing a clear framework for biological interpretation.

The analysis of the pan-cancer omics dataset further highlights the model's systems-level understanding of cancer pathology. The model learned to dynamically weigh the importance of different molecular data types, i.e., gene expression, copy number variation, and DNA methylation, on a per-cancer basis, effectively learning the primary molecular driver for each disease. Its reliance on mRNA for functionally defined cancers like LIHC and PRAD, on CNV for genomically unstable cancers like OV, and on methylation for epigenetically driven cancers like LGG, demonstrates a subtle understanding that aligns with decades of cancer research. The prostate cancer analysis provides a powerful case study for the model's ability to deconstruct the biology of tumor progression. Our model successfully distinguished primary from metastatic tumors with 99.0\% accuracy by learning their opposing molecular signatures. It correctly identified primary tumors using markers of a stable, differentiated state, such as KLK3 (PSA). Conversely, it identified metastatic tumors by learning their signatures of heightened biosynthetic and metabolic activity, exemplified by genes like EEF1A1.

As seen from Figs.~\ref{supp_figure:ppathway_embeddings},~\ref{supp_figure:huri_embeddings},~\ref{supp_figure:tm_embeddings},~\ref{supp_figure:pan__embeddings}, and~\ref{supp_figure:pnet_embeddings}, across all five datasets, we found that the embedding space learned by Graph2Image aligns more closely with the ground-truth labels than the embeddings produced by standard GNNs such as GCN, GAT, and GIN. From Table~\ref{supp_tab:emb_metrics}, Figs.~\ref{supp_figure:ppathway_embeddings} and \ref{supp_figure:huri_embeddings}, for the PP-Pathways and HuRI interactomes, respectively, Graph2Image is the only method that yields clearly non-trivial agreement between the embedding structure and the tissue labels, increasing ARI, NMI and silhouette scores over all GNN baselines and producing visibly tighter, more label-consistent clusters in t-SNE space. In Table~\ref{supp_tab:emb_metrics} for the pan-cancer dataset, Graph2Image achieves an ARI of 0.87 and an NMI of 0.93, clearly outperforming the GNN baselines (ARI $\approx$ 0.39–0.64, NMI $\approx$ 0.68–0.80). This is reflected in the t-SNE plots in Fig.~\ref{supp_figure:pan__embeddings}, where Graph2Image yields compact, well-separated clusters for each cancer type, whereas the GNN embeddings show elongated manifolds with partially overlapping classes. For the Tabula Muris dataset in Table~\ref{supp_tab:emb_metrics}, the ARI of Graph2Image is comparable to GCN and GAT, but the NMI and homogeneity are substantially higher (NMI 0.83 vs.\ 0.77–0.79; homogeneity 0.91 vs.\ 0.81), indicating that Graph2Image forms purer clusters whose label composition is closer to one-hot, consistent with the visually tighter CNN t-SNE clusters in Fig.~\ref{supp_figure:tm_embeddings}. The advantage is most pronounced in the Prostate Cancer primary–metastasis cohort. In Table~\ref{supp_tab:emb_metrics} Graph2Image attains an ARI of 0.67 and a very high silhouette score of 0.93, while all GNNs are near random (ARI $\approx$ 0.05 or even negative) and show heavy mixing of primary and metastatic samples in their t-SNE embeddings in Fig.~\ref{supp_figure:pnet_embeddings}. In contrast, the Graph2Image embedding separates primary tumors and metastases into clearly disjoint islands. Together, these quantitative metrics and qualitative visualizations show that Graph2Image learns a substantially more discriminative and label-consistent feature space than conventional GNN approaches.

While Graph2Image establishes a powerful new paradigm for graph analysis, it has certain limitations. The feature-mapping step is explicitly driven by how node features co-vary across the graph: it constructs a feature–feature association matrix from pairwise correlations and then uses this matrix to define the two-dimensional layout. Consequently, the approach works best for modalities that exhibit rich, structured co-variation, such as gene expression or DNA methylation profiles, in which groups of features form coherent biological programs. In contrast, for feature sets that are extremely sparse or show little systematic correlation across nodes (for example, nearly binary mutation indicators), the resulting association matrix is less informative, and the corresponding feature maps may be less effective. Furthermore, the initial community detection step is critical; the quality of the structural embedding is sensitive to the number of communities chosen, and selecting a number that is too small or too large can distort the learned layout and adversely affect performance.
 
Graph2Image reframes biological network analysis as an image interpretation problem. This allows dense biological graphs to be processed end-to-end without aggressive sparsification while still providing direct access to feature attributions. We evaluated Graph2Image on two large human interactomes with tissue-of-expression labels, a whole-organism cell atlas, a pan-cancer multi-omics cohort, and a clinically relevant prostate cancer diagnosis task. In all five settings, the method achieved state-of-the-art classification performance. At the same time, it organized its decision space along known biological structures: it recovered established cell lineages in the atlas, captured tissue-of-origin relationships in interactomes and cancer, and identified coherent molecular programs that distinguish primary from metastatic disease. By standardizing biological graphs into multi-channel images, the framework aligns network biology with mature tools from computer vision and SHAP-based attribution, enabling scalable computation and transparent readouts at the level of genes and pathways. Taken together, these results position Graph2Image as a practical approach for analyzing large biological networks at their native scale, while coupling predictive performance with mechanistic interpretability.

\section{Methods}
Let us assume that a graph $G = (V, E, F)$ represents the biological system that we want to analyze, where $V$ is the set of vertices (nodes) with $|V|=n$, $E$ is the set of edges, and $F \in \mathbb{R}^{n \times k}$ is the feature matrix. Graph2Image converts this attributed graph into a set of multi-channel images. First, we partition the $n$ nodes into a set of $P$ communities based on the graph's structure, where the number of communities $P$ is determined by the number of node features $k$, such that $P = \lceil\sqrt{k}\rceil$. We then use these communities to create two distinct 2D embeddings, one for structure and one for features—which are combined into a final multi-channel image. As illustrated in Fig.~\ref{figure:graph2image}, this transformation enables the graph's relational and feature information to be processed by CNNs.

\subsection{Community Detection via Node Connectivity Profiles}
Consider an undirected graph $G=(V,E)$, where $V=\{v_{1},v_{2},\dots ,v_{n}\}$ is the set of $n$ vertices and $E$ is the set of edges. The weighted adjacency matrix $A$ of this graph is an $n \times n$ matrix where each element $A_{ij}$ is defined as:
\begin{equation}
A_{ij} = \begin{cases}
w_{ij} & \text{if } \{v_i, v_j\} \in E \\
0 & \text{if } \{v_i, v_j\} \notin E
\end{cases}
\end{equation}
where $w_{ij}$ is the weight of the edge between vertex $v_{i}$ and vertex $v_{j}$. In our experiments, we used graphs without any self-loops, so the diagonal elements $A_{ii}$ are all 0. Since the graph is undirected, the adjacency matrix is symmetric, i.e., $A_{ij}=A_{ji}$ for all $i,j$.

Once we have the adjacency matrix, we identify the underlying community structure of the network by partitioning the nodes into $P$ distinct groups, where $P$ is set to $\lceil\sqrt{k}\rceil$. We treat the rows of the adjacency matrix $A$ as a set of $n$ vectors in an $n$-dimensional space, $\{a_1, a_2, \dots, a_n\}$, where each vector $a_i$ represents the connectivity profile of node $v_i$.

The partitioning process is initialized by selecting $P$ seed centroids using a probabilistic strategy designed to improve the quality of the final solution \cite{arthur2006k}. The initialization begins by choosing the first centroid, $c_1$, uniformly at random from the set of all node vectors. Subsequent centroids are then selected iteratively. For each node vector $a_i$, we compute the squared Euclidean distance $D(a_i)^2 = \min_{k} \|a_i - c_k\|^2$ to the nearest existing centroid. A new centroid is then chosen from the set of all node vectors $\{a_1, \dots, a_n\}$ with a probability proportional to this squared distance. This weighted selection, expressed as $\text{P}(a_i) = D(a_i)^2 / \sum_{j=1}^{n} D(a_j)^2$, is repeated until all $P$ centroids have been seeded.

After initialization, the algorithm iteratively refines the cluster assignments and centroids to minimize the within-cluster sum of squares. Let $S_k^{(t)}$ be the set of nodes belonging to cluster $k$ at iteration $t$. In each iteration, an assignment step allocates each node vector $a_i$ to the cluster $S_k^{(t)}$ corresponding to its nearest centroid, based on the minimum squared Euclidean distance:
\begin{equation}
    S_k^{(t)} = \{ a_i : \|a_i - c_k^{(t)}\|^2 \le \|a_i - c_j^{(t)}\|^2 \quad \forall j, 1 \le j \le P \}.
\end{equation}
This partitions the $n$ nodes into $P$ clusters, defining the membership for each node $v_i$. Following the assignment, an update step recalculates the centroid of each cluster as the element-wise mean of all node vectors assigned to it:
\begin{equation}
    c_k^{(t+1)} = \frac{1}{|S_k^{(t)}|} \sum_{a_i \in S_k^{(t)}} a_i.
\end{equation}
This iterative process converges when the cluster assignments, and thus the node memberships, no longer change between iterations. The final output is a set of $P$ centroids representing the identified communities and a partition of the nodes into these communities.

Following the partitioning of nodes into their respective communities, we quantify the relationships between these communities. The set of $P$ final centroids, $\{c_1, c_2, \dots, c_P\}$, represents the average connectivity profiles of the identified communities. To measure the dissimilarity between these community profiles, we compute the pairwise Euclidean distance between each pair of centroid vectors. The element $D_{kl}$ of the resulting $P \times P$ distance matrix is calculated as:
\begin{equation}
\label{eqn:euc_dist}
    D_{kl} = \|c_k - c_l\|_2 = \sqrt{\sum_{i=1}^{n}(c_{ki} - c_{li})^2}
\end{equation}
where $c_{ki}$ and $c_{li}$ are the $i$-th elements of the centroid vectors $c_k$ and $c_l$, respectively.

Finally, to normalize the distribution of these distances, we standardize the resulting matrix $D$ using a z-score transformation. Each element $Z_{kl}$ of the final standardized distance matrix $Z$ is computed as:
\begin{equation}
\label{eqn:zscore_dist}
    Z_{kl} = \frac{D_{kl} - \mu_D}{\sigma_D}
\end{equation}
where $\mu_D$ and $\sigma_D$ are the mean and standard deviation, respectively, of all elements in the distance matrix $D$. This transformation yields the final community association matrix used for the subsequent optimal transport mapping.

\subsection{Optimal Transport for Spatial Mapping}
With the inter-community relationships quantified, the next step is to arrange these $P$ communities onto a single two-dimensional grid. The goal is to create a spatial map where the distance between communities on the grid reflects their structural similarity. We frame this as an Optimal Transport (OT) problem, which finds a correspondence between the geometry of the community relationships and the geometry of a physical grid \cite{peyre2019computational}. The method operates on two metric measure spaces: the Community Space, $(C_{\text{struct}}, p)$, which represents the intrinsic geometry of all $P$ discovered communities, and the Grid Space, $(C_{\text{grid}}, q)$, which represents the target 2D topology, with its cost matrix $C_{\text{grid}}$ composed of the pairwise squared Euclidean distances between all $P$ locations on the grid template.

We utilize the Gromov-Wasserstein (GW) discrepancy to align these two geometric structures \cite{peyre2016gromov}. The GW framework finds an optimal transport plan $T$ that minimizes the distortion between the two distance matrices, ensuring that highly associated communities are mapped to proximal locations on the grid. The GW distance is formally defined as:
\begin{equation}
\label{eqn:ot_eqn}
d_{GW}(C_{\text{struct}}, C_{\text{grid}}, p, q) := \min_{T \in \Pi(p,q)} \sum_{i,j,k,l} (C_{\text{struct}}(i, k) - C_{\text{grid}}(j, l))^2 T_{ij} T_{kl}
\end{equation}
where $\Pi(p,q) = \{T \in \mathbb{R}^{P \times P} \mid T\mathbf{1}_P = p, T^T\mathbf{1}_P = q \}$. The optimization yields a single transport plan $T$ that is resolved into a permutation matrix. This matrix dictates a fixed spatial position for each of the $P$ communities on the 2D grid, creating a \textbf{single master structural layout} that preserves the graph's overall relational information.

From this master layout, we then generate a unique image for each individual community. For a given community, its corresponding image is created by populating the grid locations with the z-scored structural similarity values between itself and all other communities in their assigned master positions. This ensures that while all images share the same spatial arrangement of communities, the pixel intensities of each image reflect the specific relational profile of its corresponding community.

In practice, solving for the exact transport plan can be computationally intensive. To improve efficiency, the OT problem can be regularized with an entropy term, $-\epsilon H(T)$~\cite{cuturi2013sinkhorn}. This yields a probabilistic mapping that can be resolved into a discrete assignment using a linear sum assignment algorithm~\cite{crouse2016implementing}. For all main results in this work, we set $\epsilon=0$ to obtain the exact, unregularized transport matrix.

\subsection{Image Construction from Structural and Feature Embeddings}
Biological networks are typically accompanied by features for individual nodes~\cite{camacho2018}. Since each node in the graph was converted to a 2D grid of spatial embedding, the logical step is to convert the features of each node into a 2D feature embedding to ensure that a CNN can be trained with both structural and functional information. We represent the node features as a matrix $F \in \mathbb{R}^{n \times k}$, where $n$ is the number of nodes and $k$ is the number of features per node. To create an image-like representation of these features, we follow a similar OT-based approach as was used for the graph structure.

First, we compute a pairwise feature association matrix, $C_{\text{feat}} \in \mathbb{R}^{k \times k}$. This is achieved by calculating the Pearson correlation between each pair of feature vectors (the columns of the feature matrix $F$), after z-score normalizing each column. Let $f_j$ and $f_l$ be the column vectors for the $j$-th and $l$-th features in $F$, respectively. The feature association matrix $C_{\text{feat}}$ is computed as:
\begin{equation}
    C_{\text{feat}} = \frac{\sum_{i=1}^{n}(f_{ij} - \bar{f}_j)(f_{il} - \bar{f}_l)}{\sqrt{\sum_{i=1}^{n}(f_{ij} - \bar{f}_j)^2}\sqrt{\sum_{i=1}^{n}(f_{il} - \bar{f}_l)^2}}
\end{equation}
where $f_{ij}$ is the value of the $j$-th feature for the $i$-th node, and $\bar{f}_j$ is the mean of the $j$-th feature vector. The OT algorithm (Eq.~\ref{eqn:ot_eqn}) is again applied to find the optimal mapping of these $P$ community feature profiles onto a separate $P \times P$ grid, creating a feature embedding that spatially encodes feature relationships.

This process results in two distinct embeddings. First, we obtain a structural embedding of the $P$ communities, and next a feature embedding of the $k$ features. To ensure these two embeddings are compatible for concatenation, the smaller of the two is zero-padded to match the dimensions of the larger one. For instance, the structural embedding of the $P$ communities is mapped in the center of a matrix with zeros onto a grid of size $P_s \times P_s$, where $P_s = \lceil\sqrt{P}\rceil$. Since $P_s \le P$, the structural embedding is padded with zeros to match the $P \times P$ dimensions of the feature embedding. Finally, the two compatible $P \times P$ embeddings are concatenated as separate channels to form the final Graph2Image, a multi-channel image of shape $P \times P \times 2$. Furthermore, this channel-based approach naturally allows for the integration of multi-omics data; a separate feature embedding can be created for each modality and concatenated as an additional channel in the final image. This final representation, which synergistically encodes both structural and feature information, is then used to train deep learning models.

\subsection{CNN Architecture and Model Training}
Once the node-specific images are constructed, we train a CNN to perform the classification task. The network architecture is designed to process the multi-channel image inputs, each of size $P \times P$. The model consists of a feature extraction backbone and a subsequent classifier head. The feature extractor comprises four sequential 2D convolutional layers, each employing a $5 \times 5$ kernel with $16$ filters. To maintain spatial resolution throughout the feature extraction process, we use same padding. A Rectified Linear Unit (ReLU) activation function is applied after each convolutional layer to introduce non-linearity. Following the convolutional blocks, the resulting feature maps are flattened into a vector and passed to a classifier head, which consists of two fully connected layers of $768$ and $512$ neurons, respectively. The output of the network is a final fully connected layer with a dimension equal to the number of classes, followed by a softmax activation function to produce a class probability distribution. The model is trained by minimizing the cross-entropy loss. While this study employs a standard CNN architecture to demonstrate the fundamental efficacy of the Graph2Image transformation, the framework is flexible and could readily incorporate more advanced convolutional designs.

For model training, the dataset was partitioned into training (70\%), validation (15\%), and testing (15\%) sets. The network was trained using the Stochastic Gradient Descent with Momentum (SGDM) optimizer with an initial learning rate of $3 \times 10^{-4}$ and a mini-batch size of 32. The model was trained for a maximum of 20 epochs, with the training data shuffled at the beginning of each epoch. To prevent overfitting and select the best-performing model, the network's performance on the validation set was evaluated at regular intervals during training. The model that achieved the lowest loss on the validation set was selected for final evaluation on the held-out test set. Performance was assessed using accuracy, precision, F1-score, and recall.

\subsection{Model Interpretability with SHapley Additive exPlanations}
To deconstruct the biological principles learned by the trained CNN and identify the features driving its predictions, we employed SHapley Additive exPlanations (SHAP), a game-theoretic approach for explaining the output of any machine learning model \cite{lundberg2017unified}. For a given prediction, SHAP assigns each feature an importance value, or Shapley value, representing its contribution to pushing the model's output from a baseline value, defined as the average prediction over a background dataset. The Shapley value for a feature $j$, $\phi_j$, is calculated by considering all possible subsets of features $S \subseteq F \setminus \{j\}$, where $F$ is the set of all features. The value is the weighted average of the feature's marginal contribution across all these subsets:
\begin{equation}
    \phi_j(f,x) = \sum_{S \subseteq F \setminus \{j\}} \frac{|S|!(|F| - |S| - 1)!}{|F|!} [f_x(S \cup \{j\}) - f_x(S)]
\end{equation}
where $f_x(S)$ is the model's prediction for input $x$ given only the subset of features $S$.

In our implementation, we first computed local, per-sample SHAP values for a subset of the 1000 most informative features for each sample in the test set. To select these features, we identified the most highly variable features from the feature matrix $F$. This was achieved using a variance-stabilizing transformation, which models the mean-variance relationship in the data to identify features with higher-than-expected variance. This standard approach ensures that the subsequent, computationally intensive SHAP analysis is focused on the features most likely to contribute to the biological differences between classes. For a specific sample $x_i$ and a target class $c$, we defined a prediction function that returns the model's softmax score for that class. The \texttt{shapley} explainer was then used to compute the contribution of each pixel in the multi-channel Graph2Image to this prediction score, using the full dataset as a background reference to establish the baseline prediction. To obtain a global understanding of the model's decision-making for a particular class, we then averaged these local SHAP values for each feature across all test samples belonging to that class. This resulted in a class-specific global importance score for every feature. Finally, using the permutation matrices generated by the OT algorithm, these pixel-level importance scores were mapped back to their original biological identities (i.e., gene names for the feature channel), allowing us to identify the specific genes that positively or negatively influenced the classification.

\subsection{Datasets}
\subsubsection{PP-Pathways tissue-of-expression interactome}
To evaluate Graph2Image on a pathway-derived protein--protein interaction network, we used the PP-Pathways human interactome from the SNAP collection \cite{pppathways_snap}. This resource aggregates experimentally supported physical interactions from multiple curated pathway databases into a heterogeneous network of human proteins, providing a systems-level view of signaling and metabolic pathways. To obtain biologically meaningful node features and labels, we overlaid GTEx v8 bulk RNA-seq profiles onto the network \cite{gtex_science20, gtex_portal}. For each gene present in both PP-Pathways and GTEx, we computed log-transformed expression features by averaging Transcripts Per Million (TPM) across all samples from the same tissue and taking $\log_2(\mathrm{TPM}+1)$, yielding a gene-by-tissue feature matrix across 54 GTEx tissues. We then assigned each gene a tissue-of-expression label defined as the tissue in which it has the highest mean expression, resulting in a 54-class tissue-of-expression prediction task (see Supplementary Table~\ref{supp_tab:pppathways_classes}).

\subsubsection{HuRI and GTEx tissue interactome}
To test whether our approach can recover tissue programs directly from a high-quality reference interactome, we used the Human Reference Interactome (HuRI), which compiles systematically mapped, high-confidence human protein--protein interactions \cite{luck2020reference}. HuRI encompasses thousands of proteins connected by tens of thousands of binary interactions, capturing a broad range of cellular pathways in a tissue-agnostic manner. As in the PP-Pathways analysis, we annotated HuRI genes with GTEx v8 bulk RNA-seq expression across normal tissues \cite{gtex2017, gtex2020, gtex_portal}. For each gene, we summarized expression as $\log_2(\mathrm{TPM}+1)$ averaged over donors within each tissue and defined its primary tissue label as the tissue with maximal mean expression. For downstream benchmarking, we focused on nine tissues with robust representation and clear physiological roles—whole blood, brain-cerebellum, brain-cerebellar hemisphere, skeletal muscle, thyroid, ovary, testis, EBV-transformed lymphoblastoid cell lines, and cultured fibroblasts—yielding a nine-way tissue-of-expression classification problem on the HuRI network.

\subsubsection{Tabula Muris}
To assess our method's performance on a complex cellular landscape, we used the Tabula Muris collection, a foundational single-cell atlas of the model organism \textit{Mus musculus} \cite{schaum2018tabula}. This dataset captures the transcriptomic state of diverse and interacting cell populations across 20 organs, providing a system-wide view of cellular heterogeneity. We utilized a pre-processed version containing 54,865 cells across 55 distinct cell types (see Supplementary Table~\ref{supp_table:tm_class_name}), selecting the 1,089 most highly variable genes to model the core cellular interaction network (see Supplementary Tables~\ref{supp_table:tm_shap_genes_part_1},~\ref{supp_table:tm_shap_genes_part_2},~\ref{supp_table:tm_shap_genes_part_3}).

\subsubsection{Pan-Cancer Multi-Omics (MLOmics)}
To challenge our model's ability to detect patterns across heterogeneous cancer systems, we utilized the Pan-Cancer dataset from the MLOmics benchmark \cite{yang2025mlomics}. This dataset provides a systems-level view of tumor biology by integrating molecular profiles (gene expression, DNA methylation, and copy number variation) from 8314 tumor samples across 32 distinct cancer types from The Cancer Genome Atlas (TCGA) (see Supplemntary Table~\ref{supp_table:pan_class_name} for class names). The task was to classify tumors based on their integrated molecular state, reflecting the underlying regulatory networks that define each cancer type.

\subsubsection{Prostate Cancer}
To evaluate our method on a clinically critical task, we used transcriptomic data from the P-Net study, which profiled a large cohort of prostate cancer patients \cite{elmarakeby2021biologically}. The dataset comprises 1,013 patient samples, divided into two distinct classes: 680 primary tumors and 333 metastatic tumors. The goal was to distinguish between these two disease states, a key challenge in clinical oncology. For our analysis, we constructed the network using the top 5,000 most highly variable genes from the expression data before applying the Graph2Image pipeline.

\bibliography{main}

\clearpage
\appendix
\setcounter{figure}{0}
\renewcommand{\thefigure}{S\arabic{figure}}
\setcounter{table}{0}
\renewcommand{\thetable}{S\arabic{table}}
\input{supplementary_document_main}  

\end{document}

%% file: supplementary_document_main.tex
\section*{Supporting Information}

\begin{figure}[pt]
    \centering
    \includegraphics[width=0.8\linewidth, page=1]{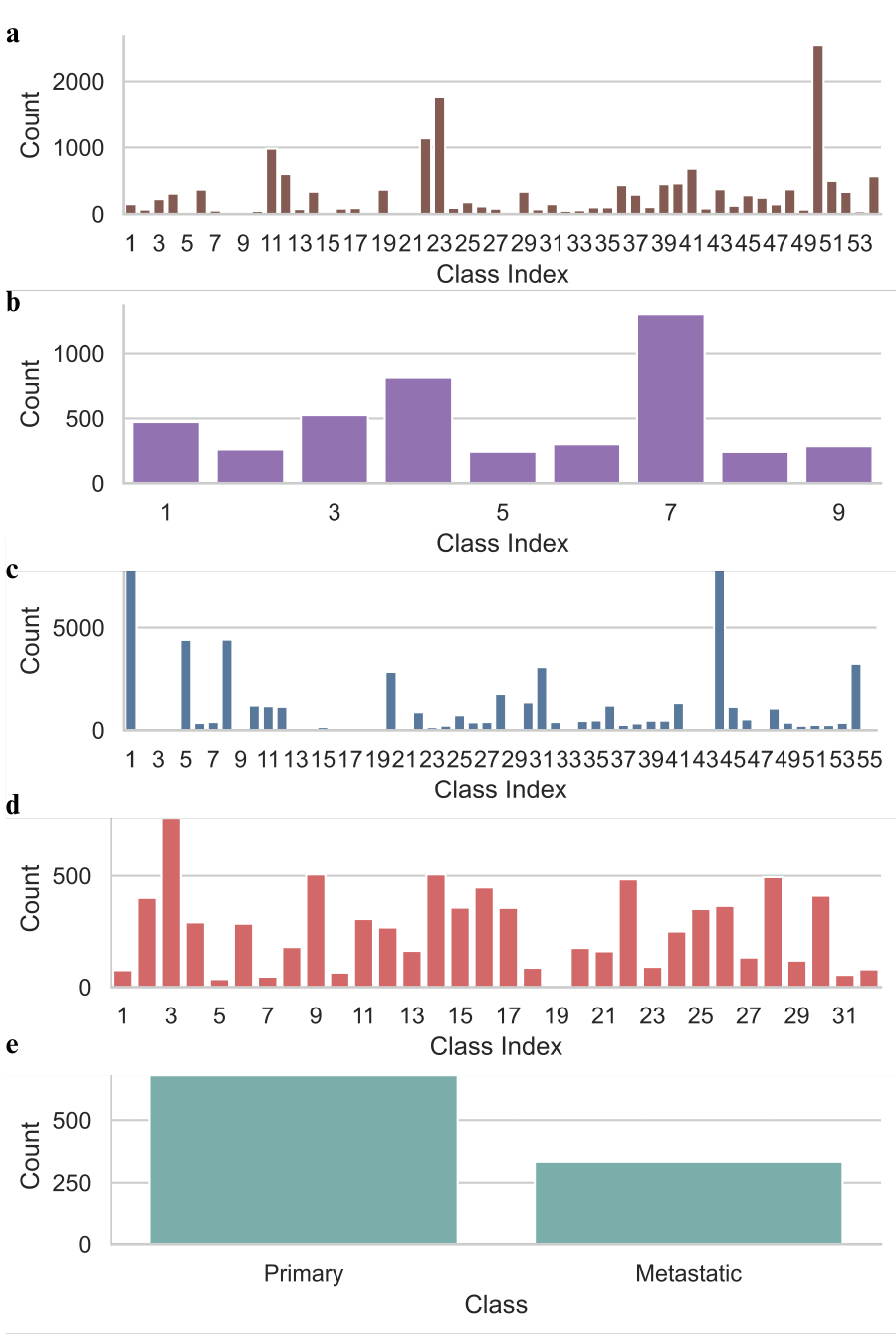}
    \caption{Class Distribution of different datasets of this study.
    \textbf{a}. PP-Pathways (see Table~\ref{supp_tab:pppathways_classes} for class names) \textbf{b}. HuRI (see Table~\ref{supp_tab:huri_classes} for class names) \textbf{c}. Tabula Muris (see Table~\ref{supp_table:tm_class_name} for class names)
    \textbf{b}. Pan Cancer (Please )see Table~\ref{supp_table:pan_class_name} for class names.
    \textbf{c}. Prostate Cancer.}
    \label{supp_fig:class_distribution}
\end{figure}

\begin{figure}[pt]
    \centering
    \includegraphics[width=\linewidth, page=2]{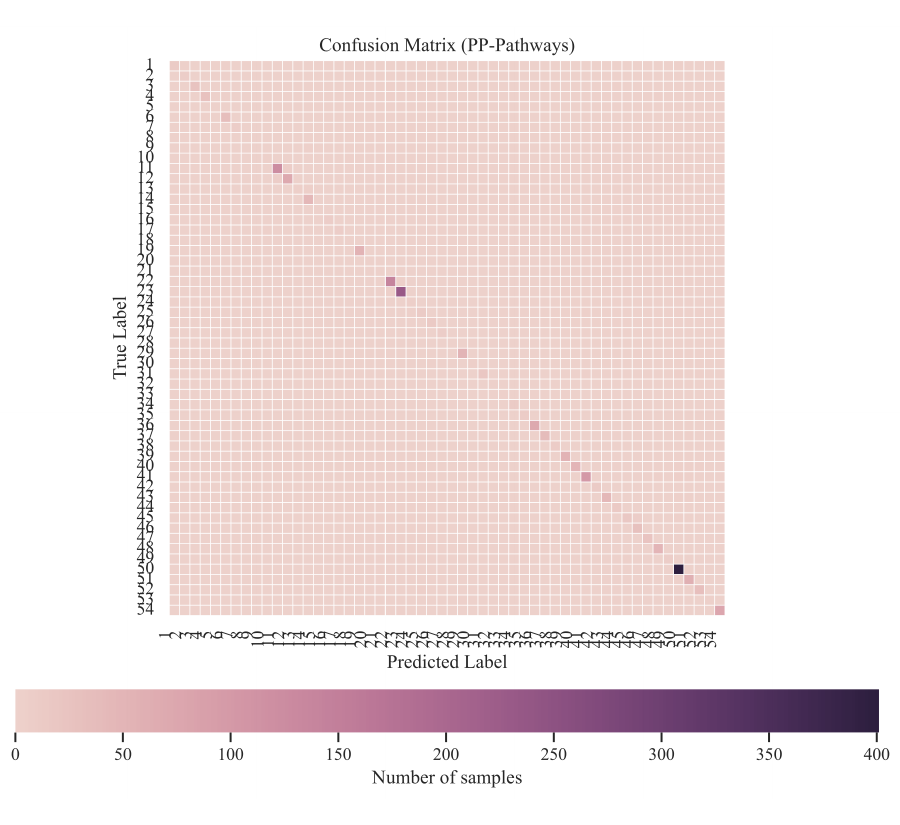}
    \caption{PP-Pathways confusion matrix. Confusion matrix of Graph2Image predictions across 54 cell types with counts shown on a logarithmic colour scale. The prominent diagonal and sparse off-diagonal entries indicate high per-class accuracy with limited cross-class confusion. Class indices and names follow Supplementary Table~\ref{supp_table:tm_class_name}.}
    \label{supp_fig:ppathway_confusion}
\end{figure}

\input{files/ppathway_class_names}

\begin{figure}[pt]
    \centering
    \includegraphics[width=0.9\linewidth, page=4]{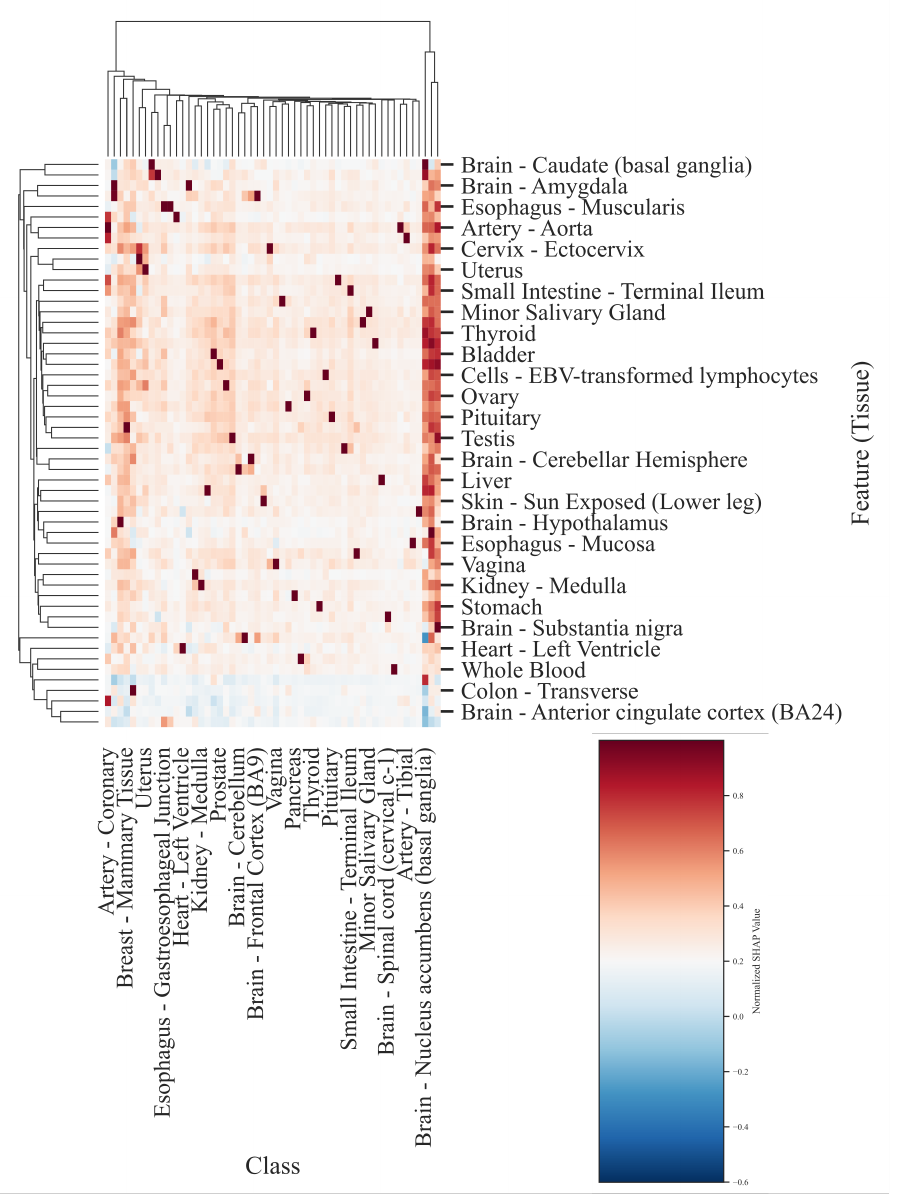}
    \caption{Clustered SHAP heatmap revealing tissue-level feature importance structure in the PP-Pathways dataset.
    Clustered heatmap showing normalized SHAP values for the top positively and negatively influential tissues across all PP-Pathways classes. Hierarchical clustering on both tissues (rows) and classes (columns) highlights coherent blocks of shared feature-importance structure. Brain-related tissues form several tight subclusters, reflecting shared expression patterns across related neuroanatomical regions, whereas digestive, reproductive, endocrine, and vascular tissues display distinct SHAP signatures. Columns correspond to PP-Pathways classes, and rows correspond to tissues selected by class-specific SHAP ranking. Color intensity indicates the normalized SHAP magnitude, with red denoting strong positive contributions and blue denoting strong negative contributions. The dual dendrogram structure provides an interpretable overview of how different tissues contribute to Graph2Image model predictions and reveals biologically meaningful relationships among tissue-specific expression profiles.}
    \label{supp_fig:ppathway_shap_heatmap}
\end{figure}

\begin{figure}[pt]
    \centering
    \includegraphics[width=0.9\linewidth, page=3]{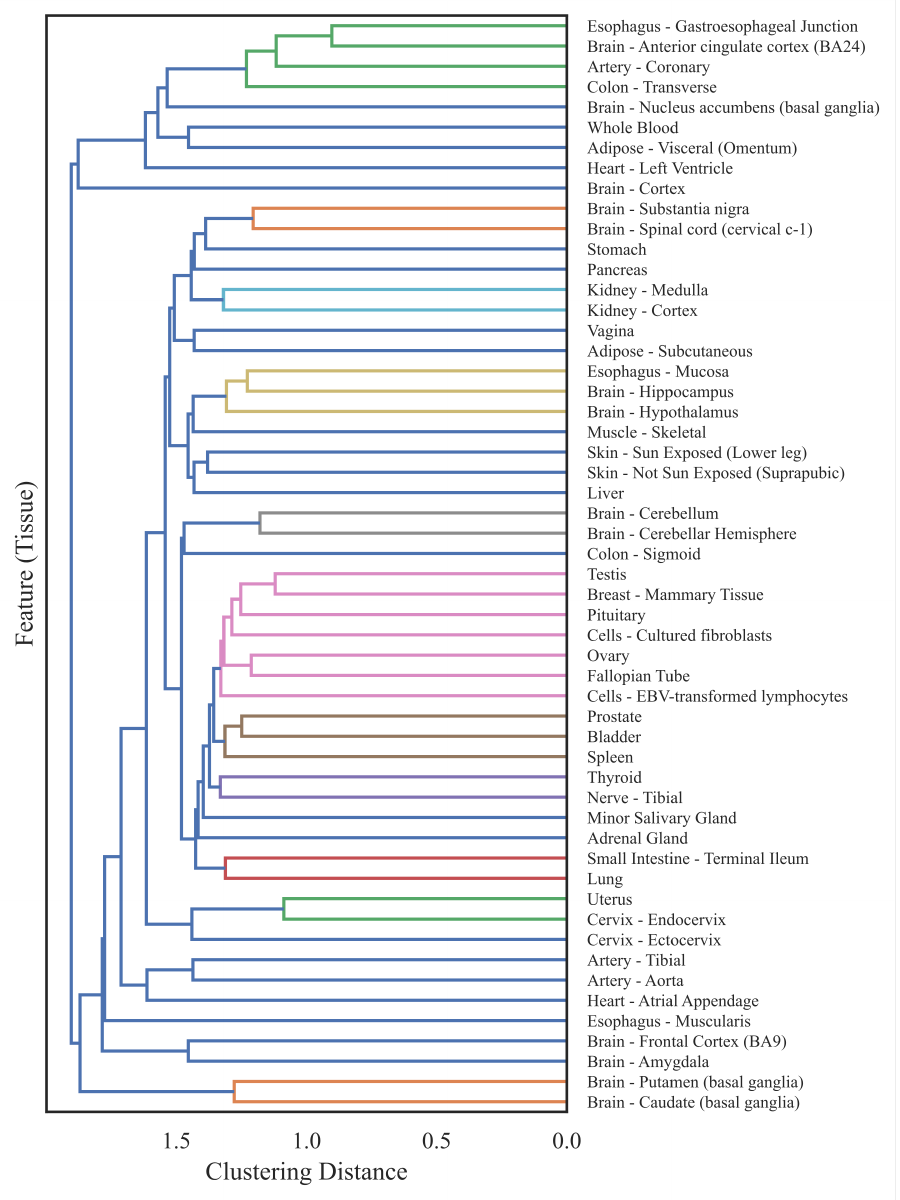}
    \caption{Hierarchical clustering of tissue-level SHAP importance profiles in the PP-Pathways dataset.
    Dendrogram showing the hierarchical relationships among tissues based on their normalized SHAP value profiles across all PP-Pathways classes. Each branch represents a tissue whose SHAP contribution reflects its discriminative importance for class-specific Graph2Image patterns. Tissues with similar SHAP signatures cluster together, revealing shared regulatory or expression landscapes, e.g., anatomically related brain regions and gastrointestinal tissues form coherent subtrees. The clustering is computed using average linkage and Euclidean distance on the top SHAP-ranked features per class. This analysis highlights interpretable groupings of functionally related tissues and provides a global view of how distinct anatomical systems contribute to Graph2Image model predictions.}
    \label{supp_fig:ppathway_dendrogram}
\end{figure}

\begin{figure}[pt]
    \centering
    \includegraphics[width=0.9\linewidth, page=5]{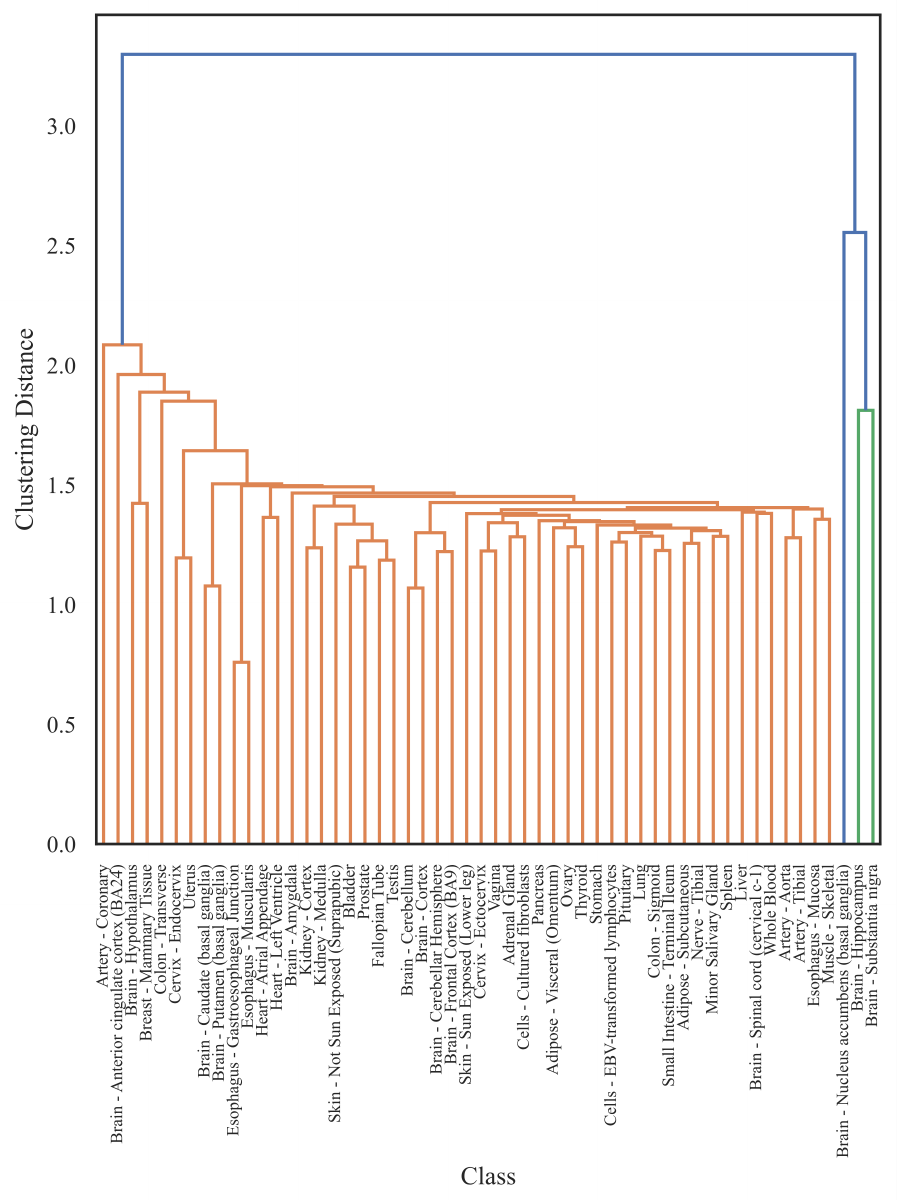}
    \caption{Hierarchical clustering of PP-Pathways classes based on normalized SHAP importance profiles.
    Dendrogram illustrating the hierarchical relationships among PP-Pathways classes derived from their normalized SHAP feature-importance signatures. Each leaf corresponds to a tissue class, and branch structure reflects similarity in the SHAP contribution patterns that drive Graph2Image predictions. Closely clustered tissues show highly similar SHAP profiles, revealing biologically coherent groupings, e.g., related brain regions cluster tightly, while immune-associated and gastrointestinal tissues form distinct subtrees. Clustering was performed using average linkage and Euclidean distance on SHAP-selected top features. This visualization provides a global view of class-level similarity in model-derived tissue importance patterns.}
    \label{supp_fig:ppathway_class_dendogram}
\end{figure}

\begin{figure}[pt]
    \centering
    \includegraphics[width=0.9\linewidth, page=6]{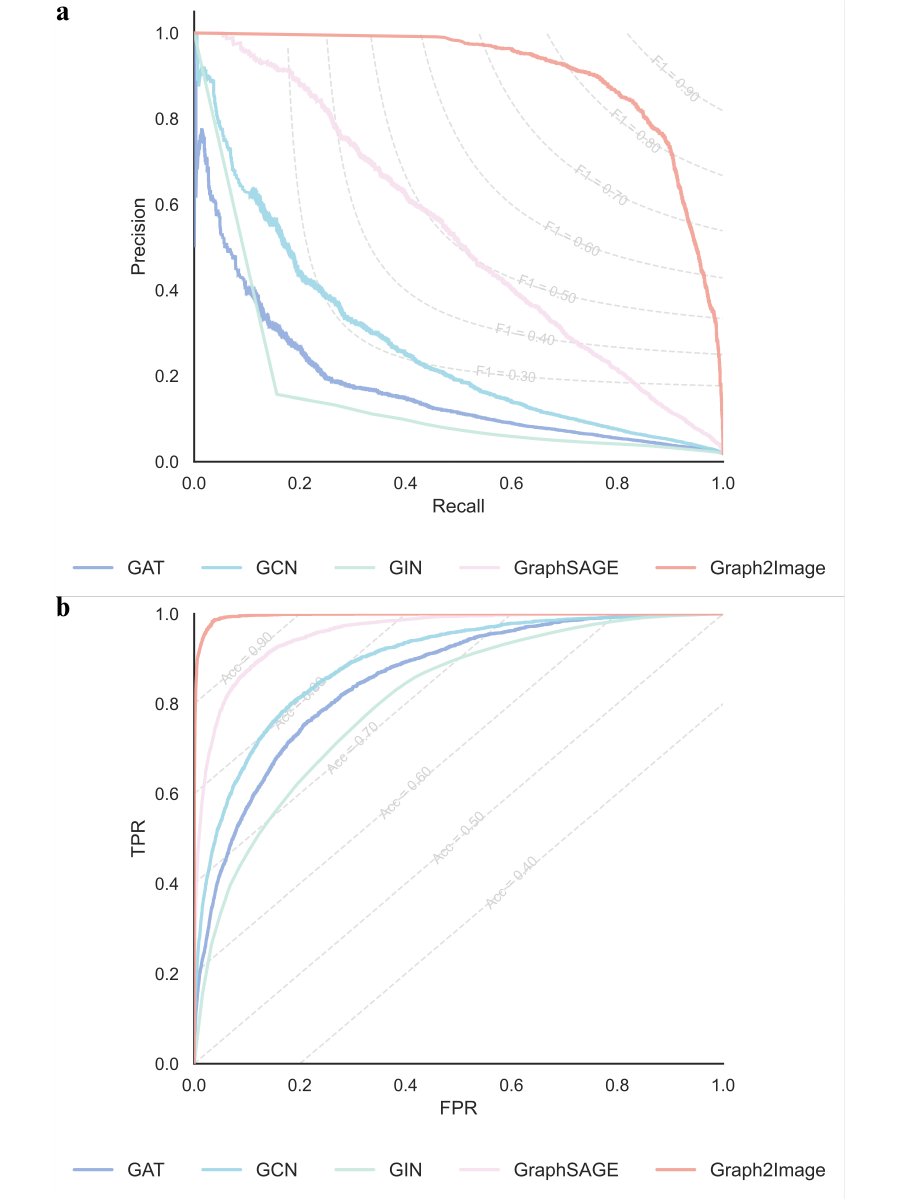}
    \caption{Precision–recall and ROC performance of Graph2Image versus GNN baselines on the PP-Pathways dataset.
    a, Macro-averaged precision–recall curves for Graph2Image and graph neural network baselines (GCN, GAT, GIN, and GraphSAGE). Dashed curves denote F1 contours. Graph2Image achieves substantially higher precision across the full recall range, reflecting stronger discrimination between tissue classes in the PP-Pathways dataset.
    b, ROC curves for the same models. Dashed lines indicate AUC contours. Graph2Image approaches near-perfect sensitivity across false-positive rates, outperforming all GNN baselines. Together, these results highlight the superior separability of Graph2Image embeddings for complex multi-tissue expression signatures.}
    \label{supp_fig:ppathway_roc_pr}
\end{figure}

\begin{figure}[pt]
    \centering
    \includegraphics[width=\linewidth, page=7]{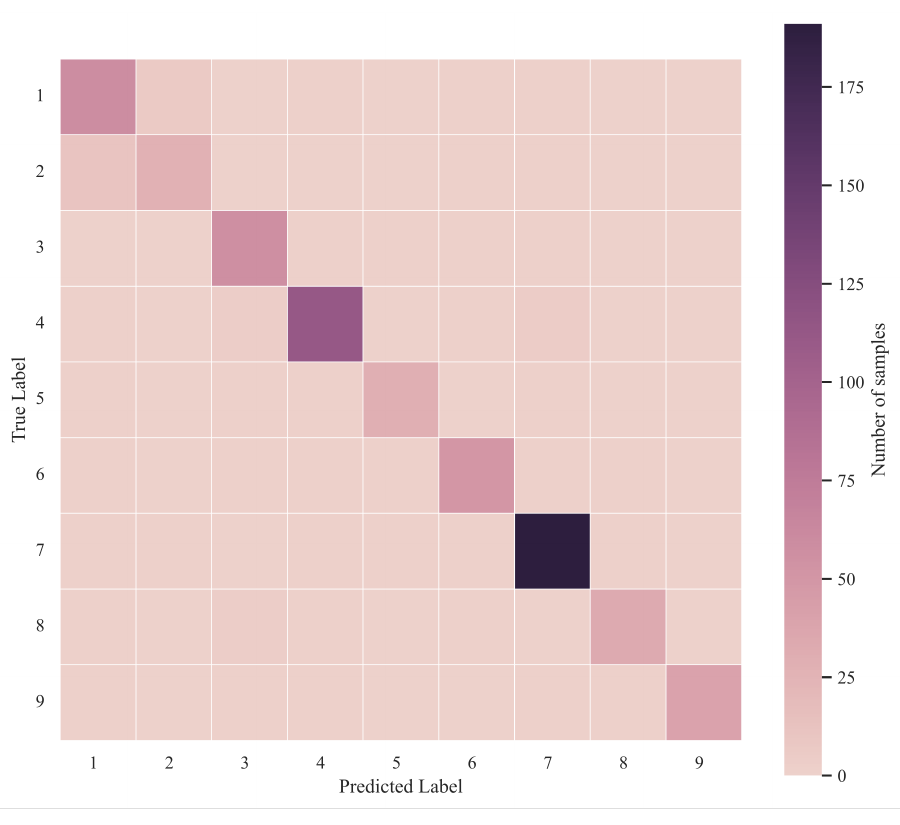}
    \caption{Confusion matrix for Graph2Image predictions on the HuRI dataset.
    Confusion matrix showing classification performance across the nine GTEx-derived tissue classes in the HuRI protein–protein interaction dataset. Graph2Image exhibits strong diagonal dominance, indicating highly accurate tissue-of-origin prediction from gene-level embeddings. Off-diagonal values are minimal across all classes, demonstrating that tissues with distinct expression signatures, i.e., Thyroid, Muscle, Testis and Whole Blood are cleanly separated by the model. Color intensity reflects the number of samples assigned to each true–predicted label pair. Class indices and names follow Supplementary Table~\ref{supp_tab:huri_classes}.}
    \label{supp_fig:huri_confusion}
\end{figure}

\input{files/huri_class_names}

\begin{figure}[pt]
    \centering
    \includegraphics[width=0.9\linewidth, page=8]{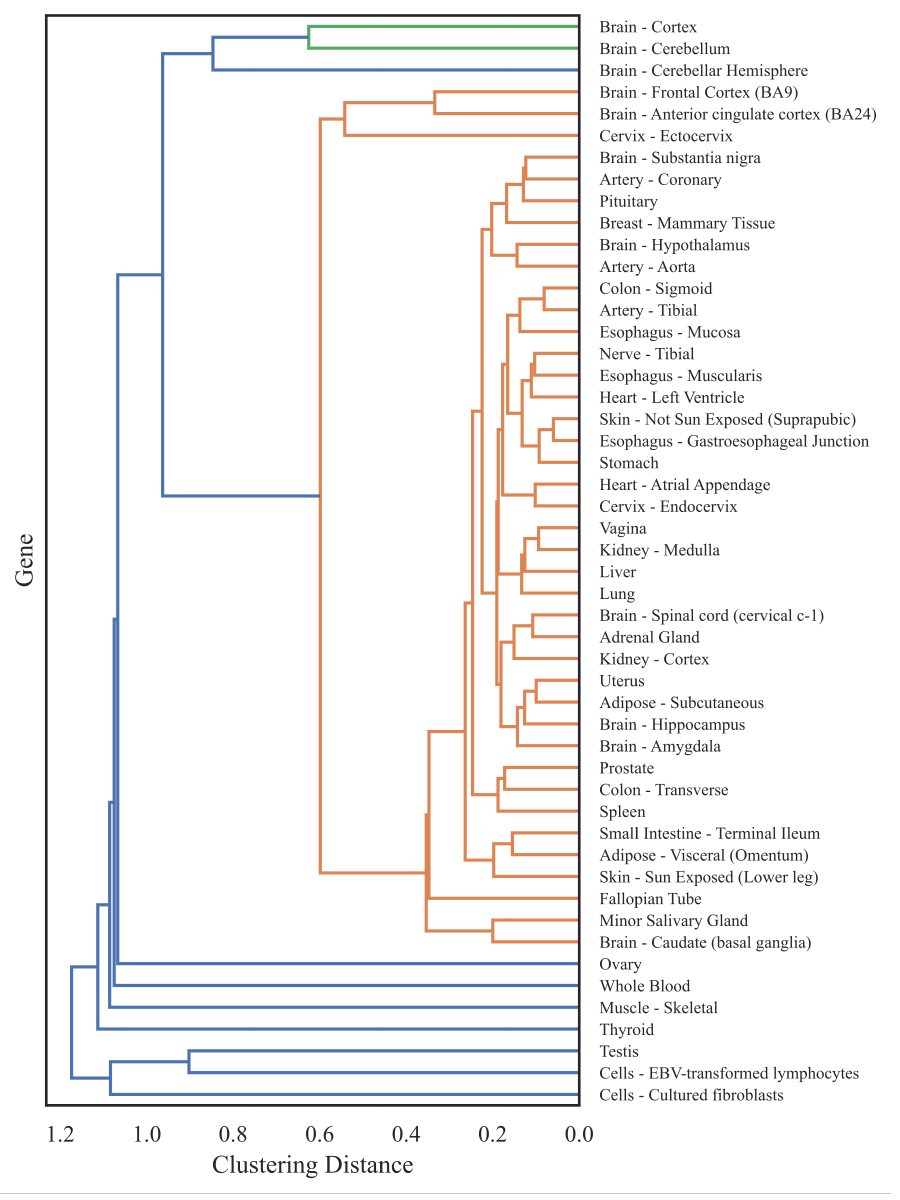}
    \caption{Hierarchical clustering of tissue-level SHAP feature importance in the HuRI dataset.
    Dendrogram showing the hierarchical relationships among GTEx tissue features based on their normalized SHAP importance profiles for the HuRI Graph2Image classifier. Each leaf corresponds to a tissue whose SHAP values summarize how strongly that feature contributes to distinguishing the nine HuRI tissue classes. Closely related tissues, such as multiple cortical and cerebellar brain regions or adipose and skin tissues, form coherent subclusters, whereas immune, endocrine and gastrointestinal tissues separate into distinct branches. Distances were computed from SHAP-derived importance vectors using Euclidean distance with average-linkage agglomeration. This analysis indicates that the model’s feature attributions recover biologically meaningful similarity structure among tissues in the HuRI dataset.}
    \label{supp_fig:huri_tissue_dendrogram}
\end{figure}

\begin{figure}[pt]
    \centering
    \includegraphics[width=0.9\linewidth, page=9]{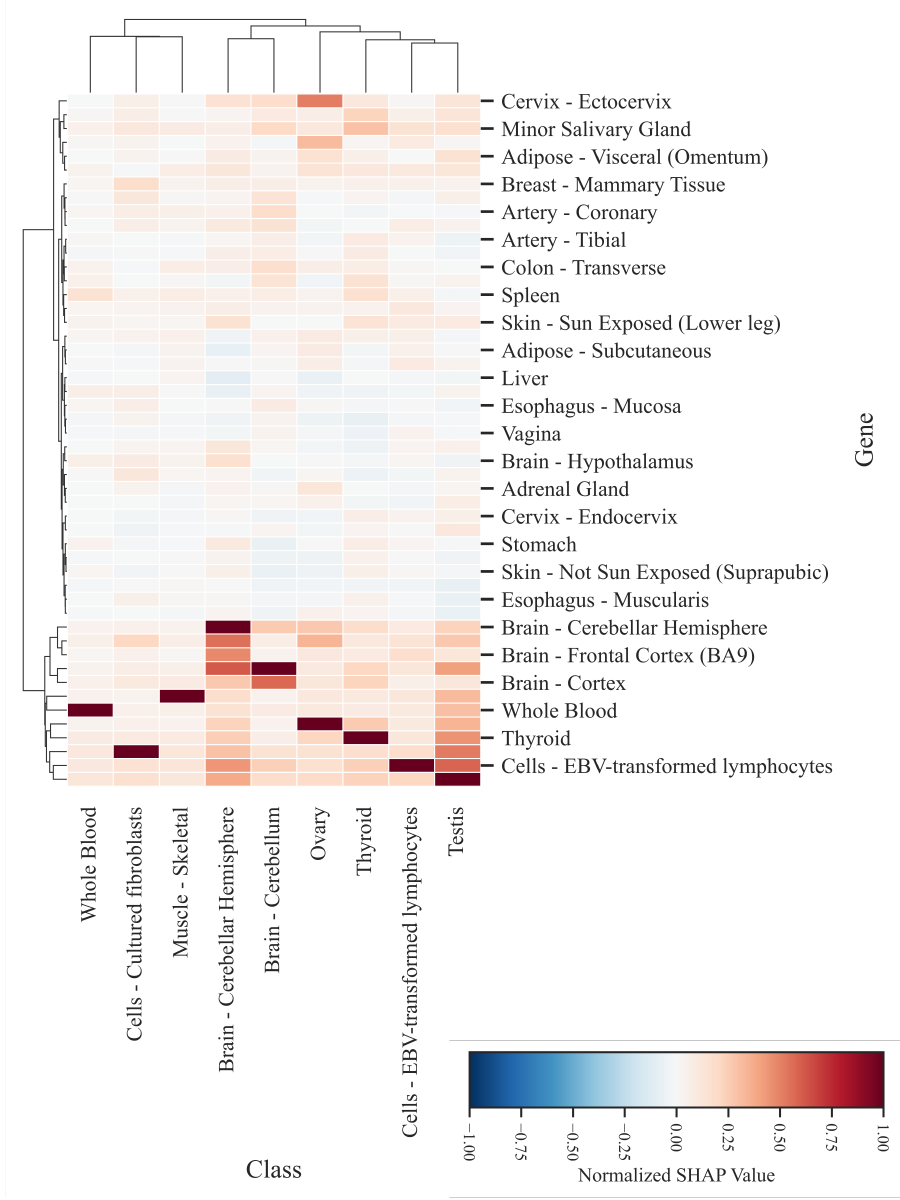}
    \caption{SHAP-based interpretation of Graph2Image predictions on the HuRI interactome.
    Hierarchically clustered heatmap showing the normalized SHAP values for the top predictive genes across the nine GTEx tissues represented in the HuRI–GTEx joint dataset. Rows denote genes selected based on the highest absolute class-specific SHAP contributions, and columns correspond to HuRI tissue classes. Positive SHAP values (red) indicate features that strongly drive the model toward a given tissue prediction, whereas negative values (blue) represent features that oppose that prediction. Dendrograms along both axes reveal that Graph2Image uncovers biologically meaningful co-regulation structure, grouping related tissues (e.g., brain tissues and immune-derived cell types) and clustering coherent gene sets associated with tissue-specific expression programs. The scale bar represents normalized SHAP values.}
    \label{supp_fig:huri_shap_heatmap}
\end{figure}

\begin{figure}[pt]
    \centering
    \includegraphics[width=0.9\linewidth, page=5]{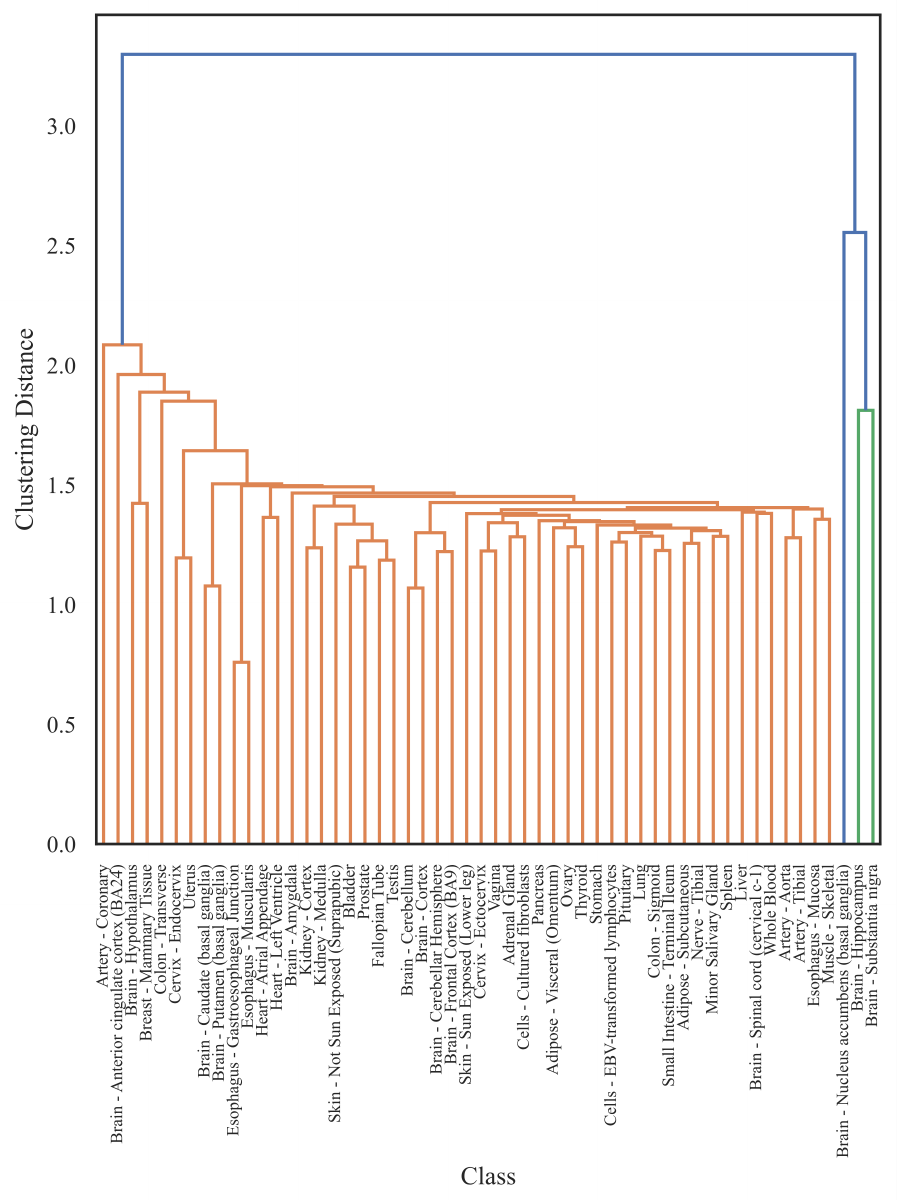}
    \caption{Hierarchical clustering of tissue-specific SHAP profiles in the HuRI interactome.
    Dendrogram showing hierarchical clustering of the nine HuRI tissue classes based on their normalized SHAP importance profiles derived from Graph2Image embeddings. Distances reflect similarity in feature attribution patterns across tissues, revealing biologically meaningful groupings such as the close proximity of brain regions (Cerebellum and Cerebellar Hemisphere), immune-related cell types (EBV-transformed lymphocytes and cultured fibroblasts), and reproductive tissues (Ovary and Testis). Clustering was performed using average linkage and Euclidean distance.}
    \label{supp_fig:huri_class_dendogram}
\end{figure}

\begin{figure}[pt]
    \centering
    \includegraphics[width=0.9\linewidth, page=11]{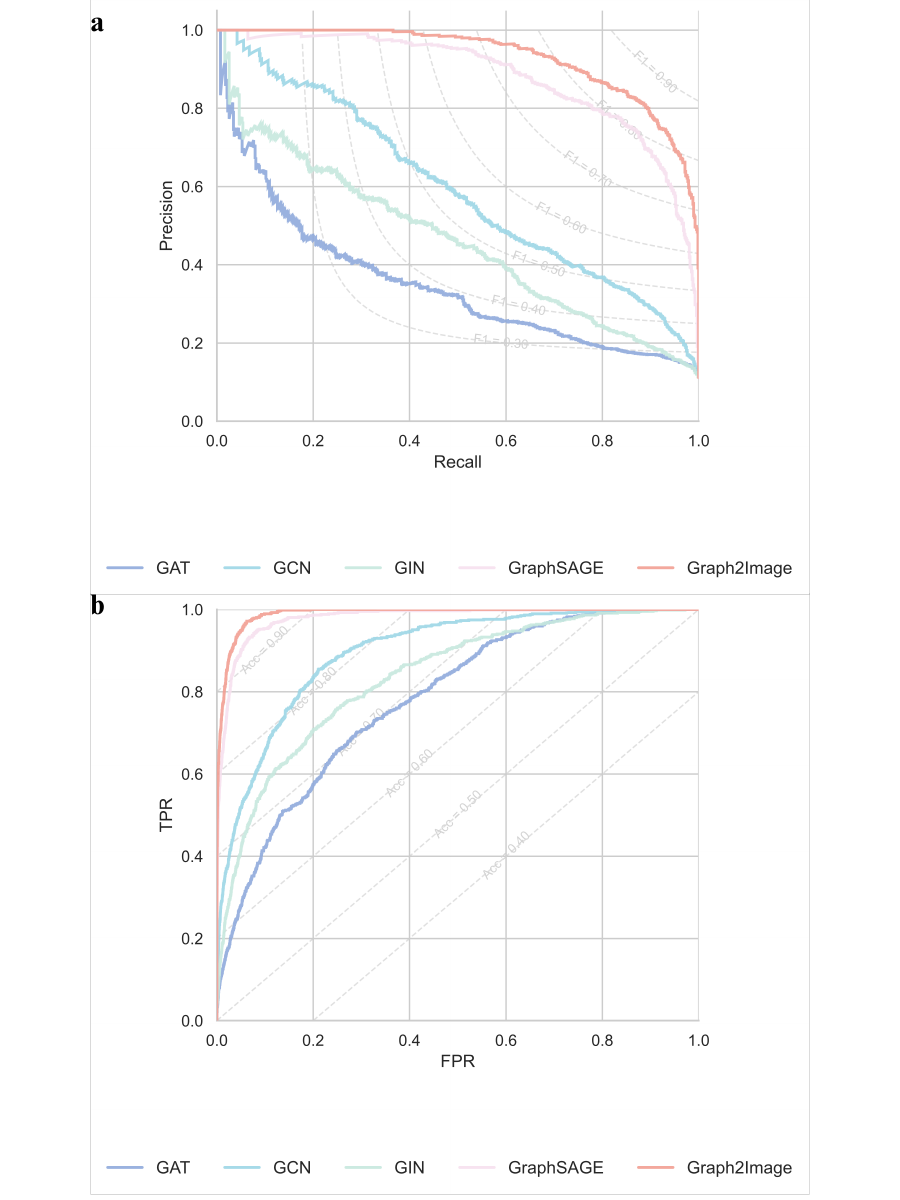}
    \caption{Performance comparison of Graph2Image and graph neural network baselines on the HuRI interactome.
    a, Micro-averaged precision–recall curves for Graph2Image and baseline GNNs (GCN, GAT, GIN, GraphSAGE) on the nine-class HuRI tissue classification task. Graph2Image achieves consistently higher precision across the full recall range, indicating stronger discriminative ability in this sparse interactome setting.
    b, ROC curves showing that Graph2Image attains markedly higher true-positive rates at low false-positive rates, reflecting improved sensitivity. Dashed lines denote F1 (panel a) and AUROC (panel b) reference contours for visual calibration.}
    \label{supp_fig:huri_roc_pr}
\end{figure}

\input{files/huri_gene_names}

\begin{figure}[pt]
    \centering
    \includegraphics[width=0.75\linewidth, page=12]{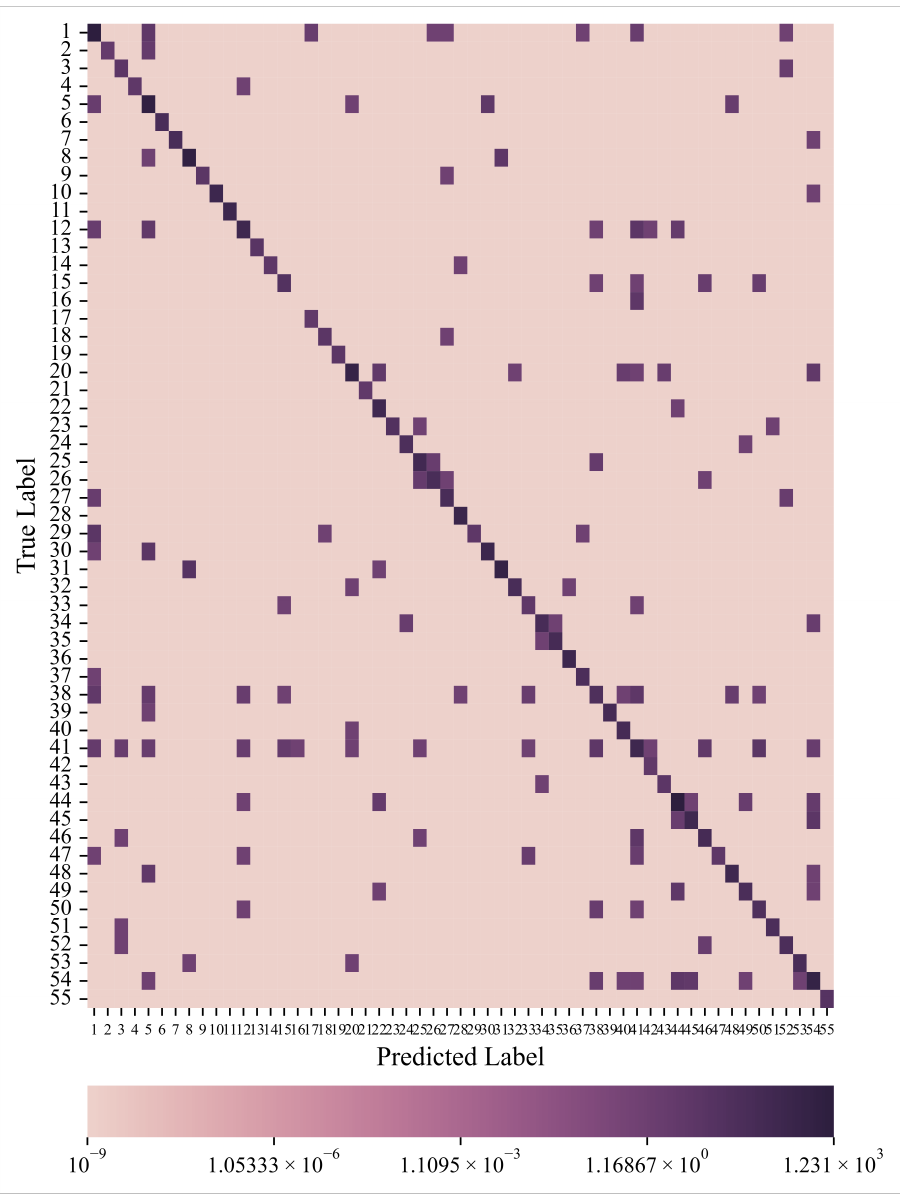}
    \caption{Tabula Muris confusion matrix. Confusion matrix of Graph2Image predictions across 55 cell types with counts shown on a logarithmic colour scale. The prominent diagonal and sparse off-diagonal entries indicate high per-class accuracy with limited cross-class confusion. Class indices and names follow Supplementary Table~\ref{supp_table:tm_class_name}.}
    \label{supp_fig:tm_confusion}
\end{figure}

\begin{figure}[pt]
    \centering
    \includegraphics[width=0.75\linewidth, page=13]{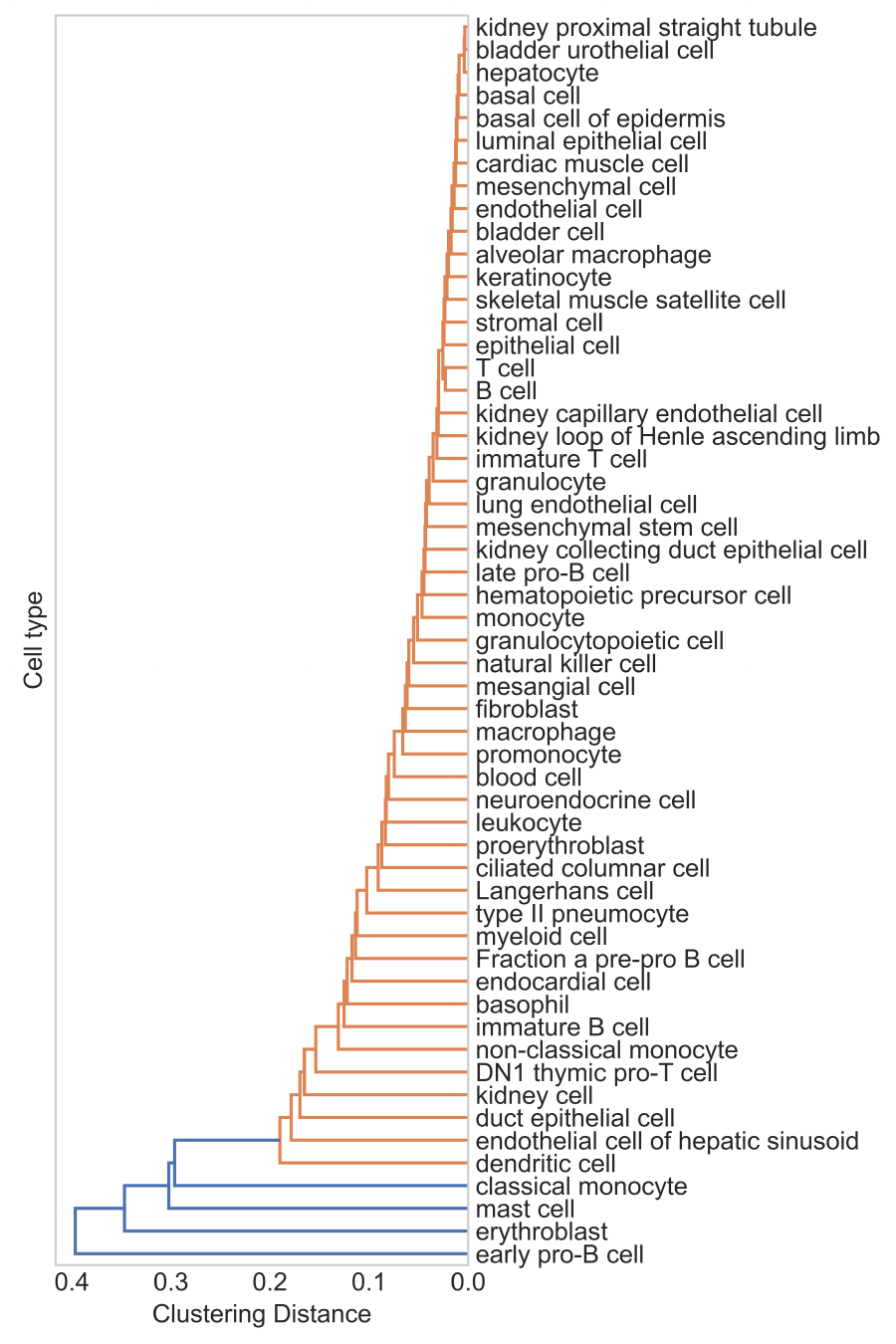}
    \caption{Tabula Muris dendrogram from SHAP profiles. Hierarchical clustering of the 55 cell types using class-averaged, gene-level SHAP profiles (average-linkage, Euclidean distance). The branching recapitulates expected haematopoietic and epithelial lineages and reveals coherent groupings discussed in the main text.}
    \label{supp_fig:tm_dendrogram}
\end{figure}

\begin{figure}[pt]
    \centering
    \includegraphics[width=0.75\linewidth, page=14]{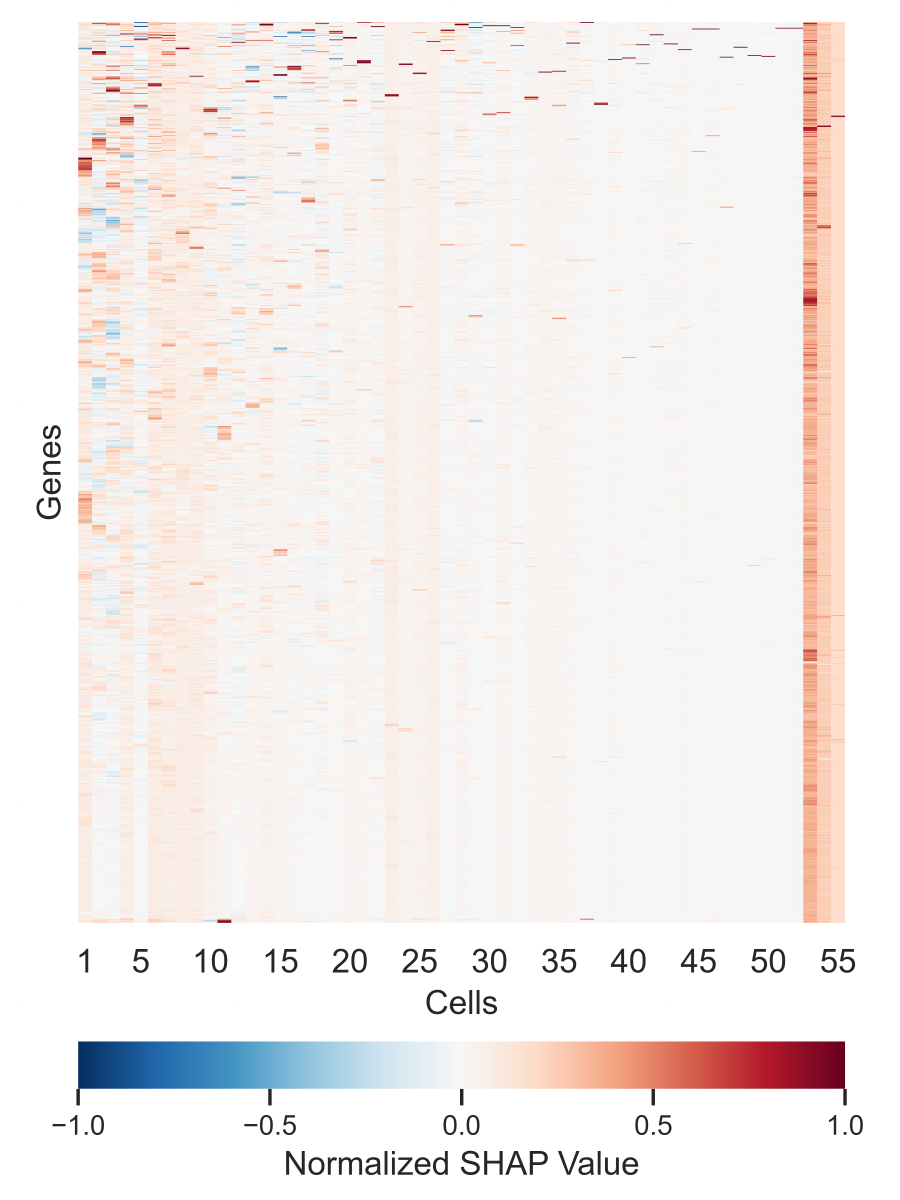}
    \caption{Tabula Muris class-averaged SHAP heatmap. Heatmap of class-averaged SHAP values for top marker genes across all 55 cell types. SHAP values are normalised within each class to the maximum absolute value; warm colours denote features increasing class probability and cool colours the converse. Gene names are listed in Supplementary Table~\ref{supp_table:tm_shap_genes_part_1}, \ref{supp_table:tm_shap_genes_part_2}, and \ref{supp_table:tm_shap_genes_part_3}; cell-type labels follow Supplementary Table~\ref{supp_table:tm_class_name}.}
    \label{supp_fig:tm_shap_heatmap}
\end{figure}

\input{files/tm_class_names}
\input{files/tm_gene_dendogram}

\begin{figure}[pt]
    \centering
    \includegraphics[width=0.9\linewidth, page=15]{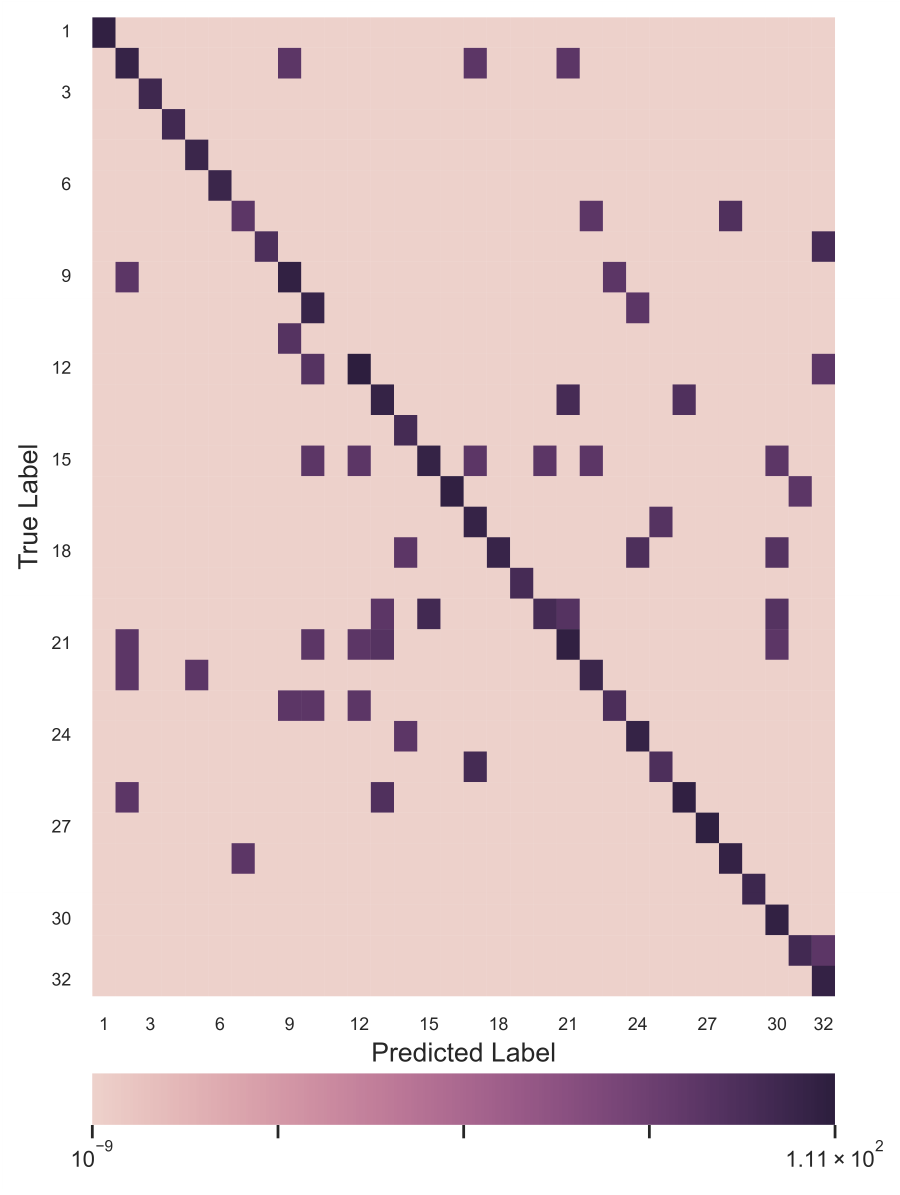}
    \caption{Pan-cancer confusion matrix. Confusion matrix of Graph2Image predictions across 32 TCGA cancer types with a logarithmic colour scale. The strong diagonal highlights accurate class assignment, while rare off-diagonal entries capture confusions among related tissues. Cancer-type names and order follow Supplementary Table~\ref{supp_table:pan_class_name}.}
    \label{supp_fig:pan_confusion}
\end{figure}

\begin{figure}[pt]
    \centering
    \includegraphics[width=0.8\linewidth, page=16]{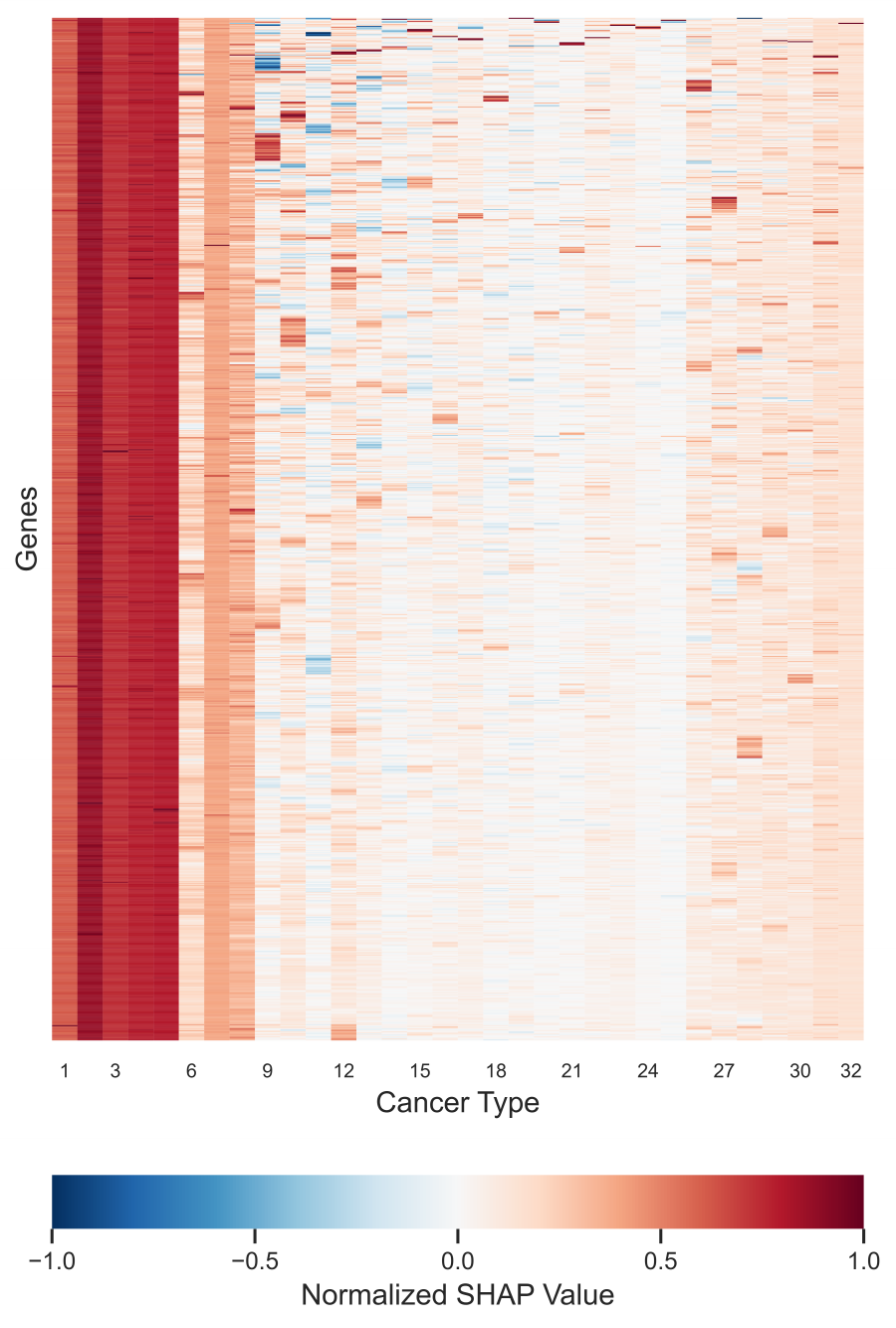}
    \caption{Pan-cancer class-averaged SHAP heatmap. Heatmap of class-averaged SHAP values for mRNA features across 32 cancer types, normalised within class (max $|$SHAP$|=1$). Distinct blocks reflect shared transcriptional programmes and cancer-type-specific signatures described in the main text. Gene names are provided in Supplementary Table~\ref{supp_table:pan_shap_genes_part_1} and \ref{supp_table:pan_shap_genes_part_2}; labels follow Supplementary Table~\ref{supp_table:pan_class_name}.}
    \label{supp_fig:pan_shap_heatmap}
\end{figure}

\begin{figure}[pt]
    \centering
    \includegraphics[width=0.8\linewidth, page=17]{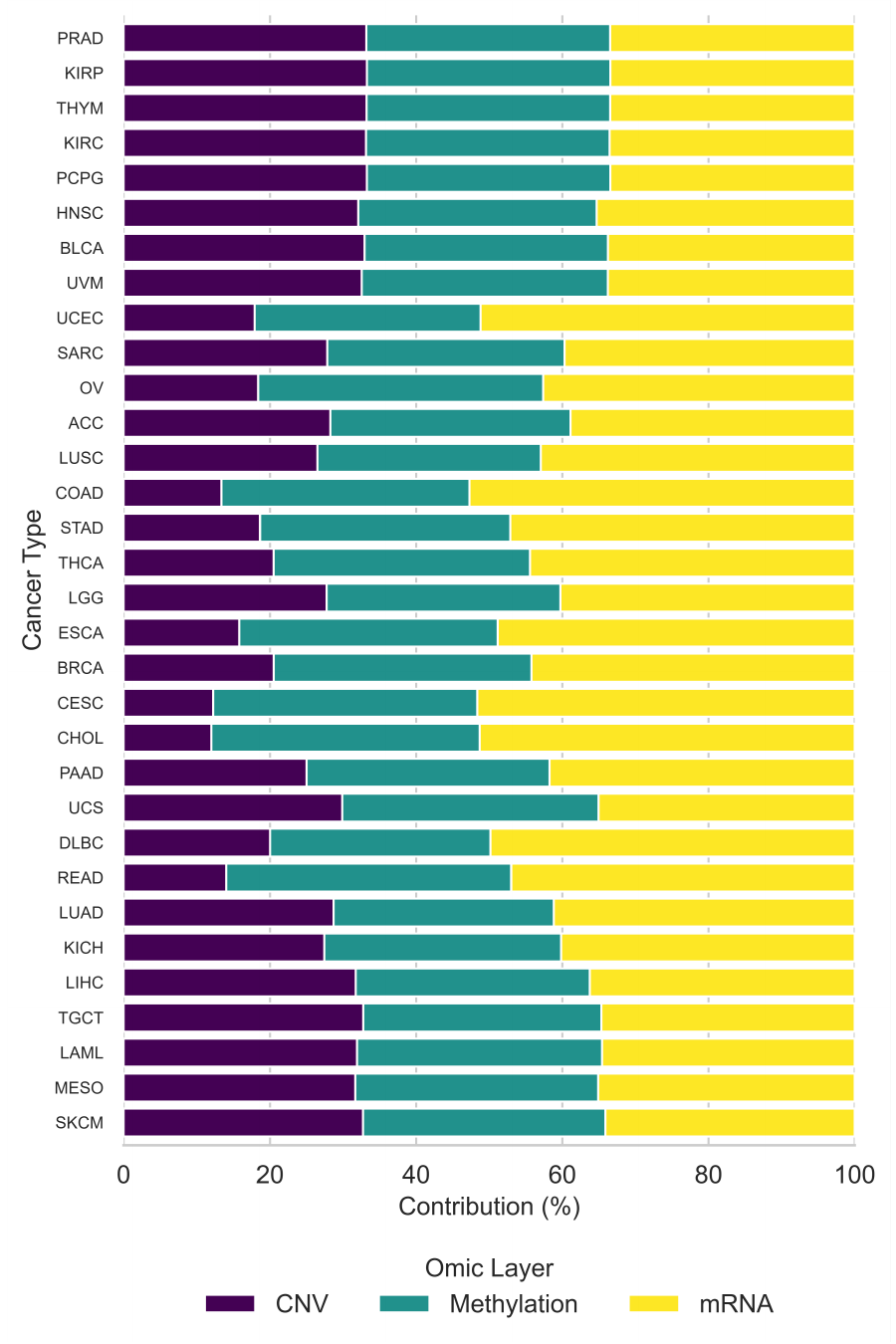}
    \caption{Relative contribution of omics modalities. Per-class mean absolute SHAP partitioned by mRNA, copy-number variation (CNV) and DNA methylation for each cancer type. Modality usage mirrors tissue of origin and known driver biology. Cancer-type order matches Supplementary Table~\ref{supp_table:pan_class_name}.}
    \label{supp_fig:pan_modal_contrib}
\end{figure}

\begin{figure}[pt]
    \centering
    \includegraphics[width=0.8\linewidth, page=18]{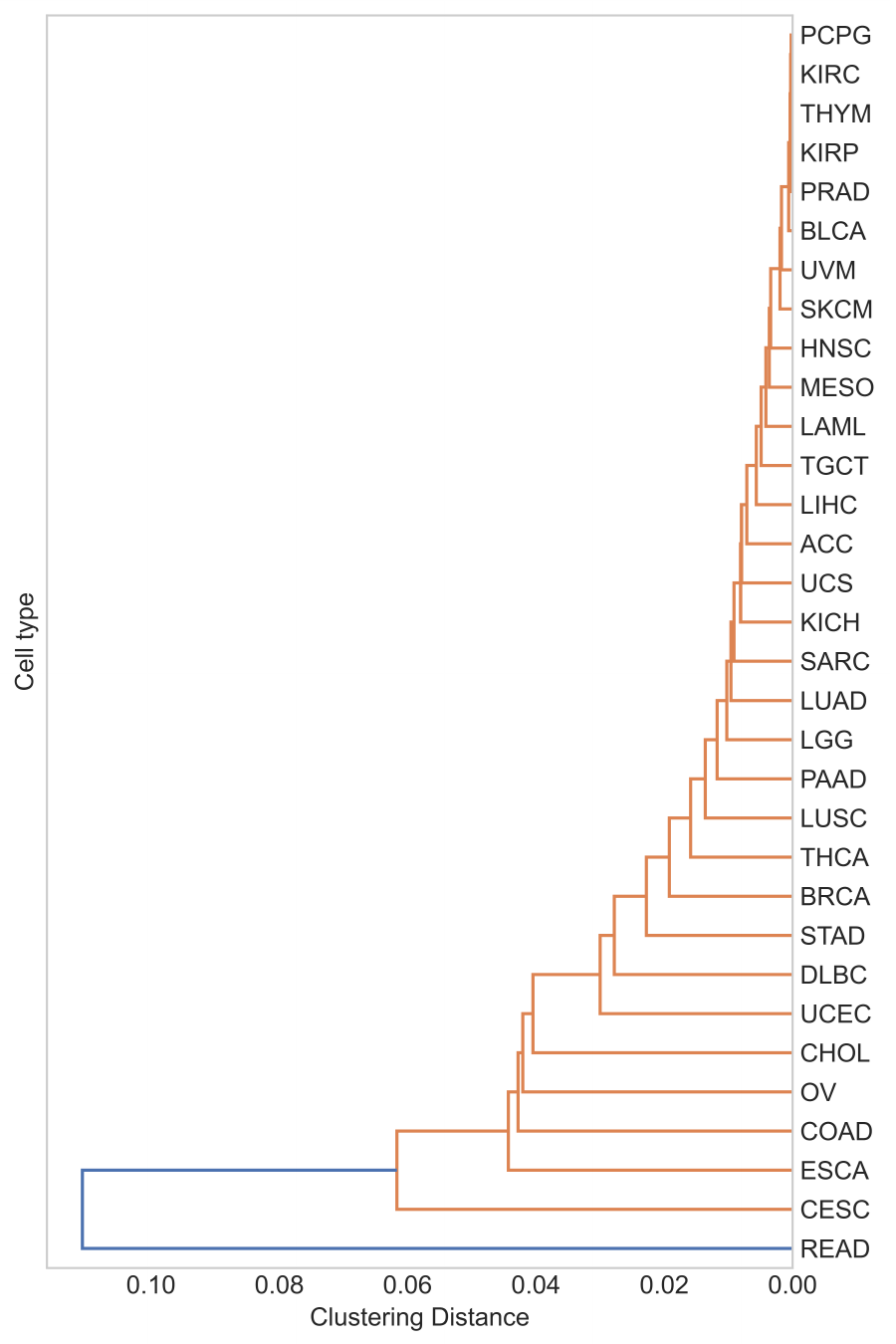}
    \caption{Pan-cancer dendrogram from SHAP profiles. Hierarchical clustering of 32 cancer types using class-level SHAP profiles (average-linkage, Euclidean distance). The arrangement groups tumours with shared molecular programmes and anatomical origin; salient clades are discussed in the main text.}
    \label{supp_fig:pan_dendrogram}
\end{figure}

\input{files/pan_label_abbreviation}
\input{files/pan_labels}
\input{files/pan_gene_list}

\begin{figure}[pt]
    \centering
    \includegraphics[width=0.9\linewidth, page=9]{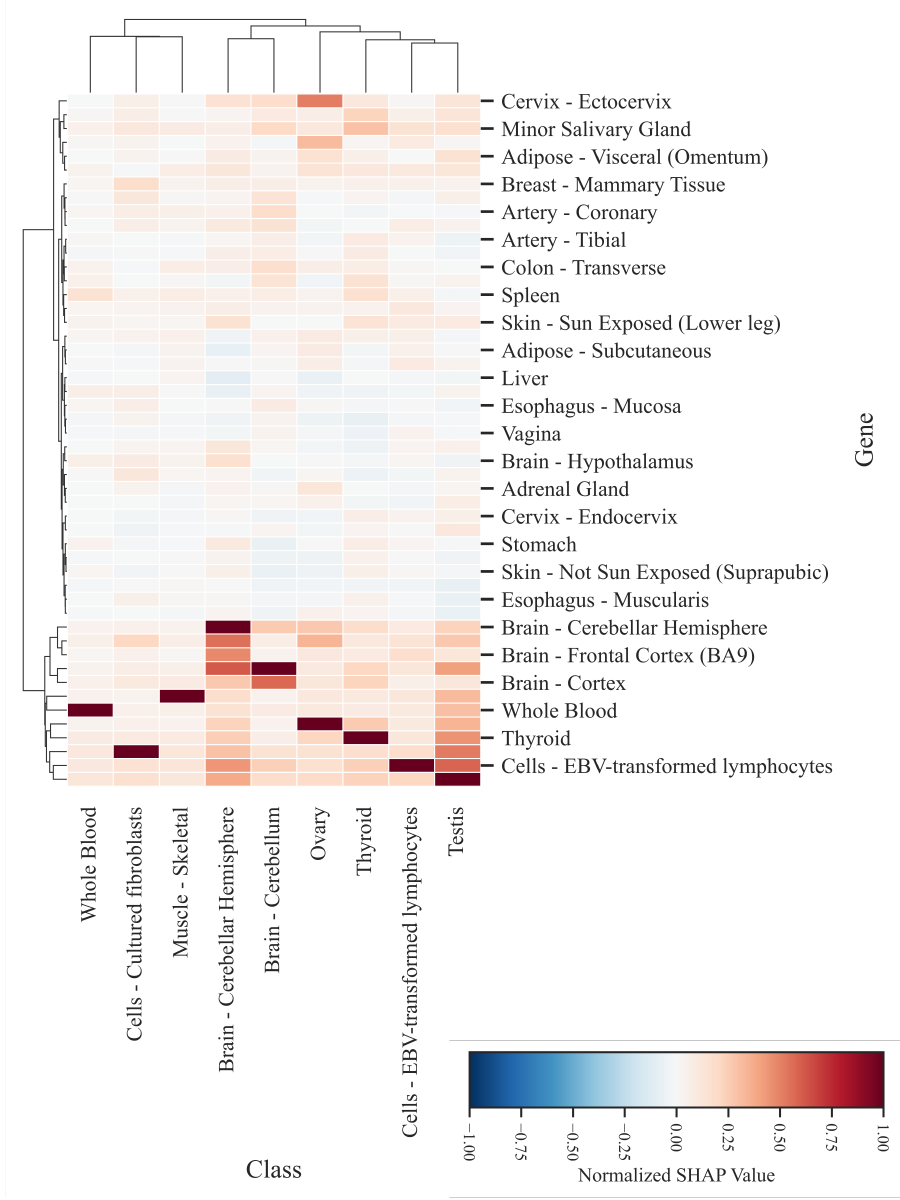}
    \caption{‘Butterfly’ plots of gene-level SHAP in prostate cancer. Mirrored kernel-density estimates of SHAP distributions for selected genes in primary versus metastatic tumours highlight features that preferentially drive each state. Genes are ordered by cohort-wide importance.}
    \label{supp_fig:pnet_butterfly}
\end{figure}

\begin{figure}[pt]
    \centering
    \includegraphics[width=0.85\linewidth, page=20]{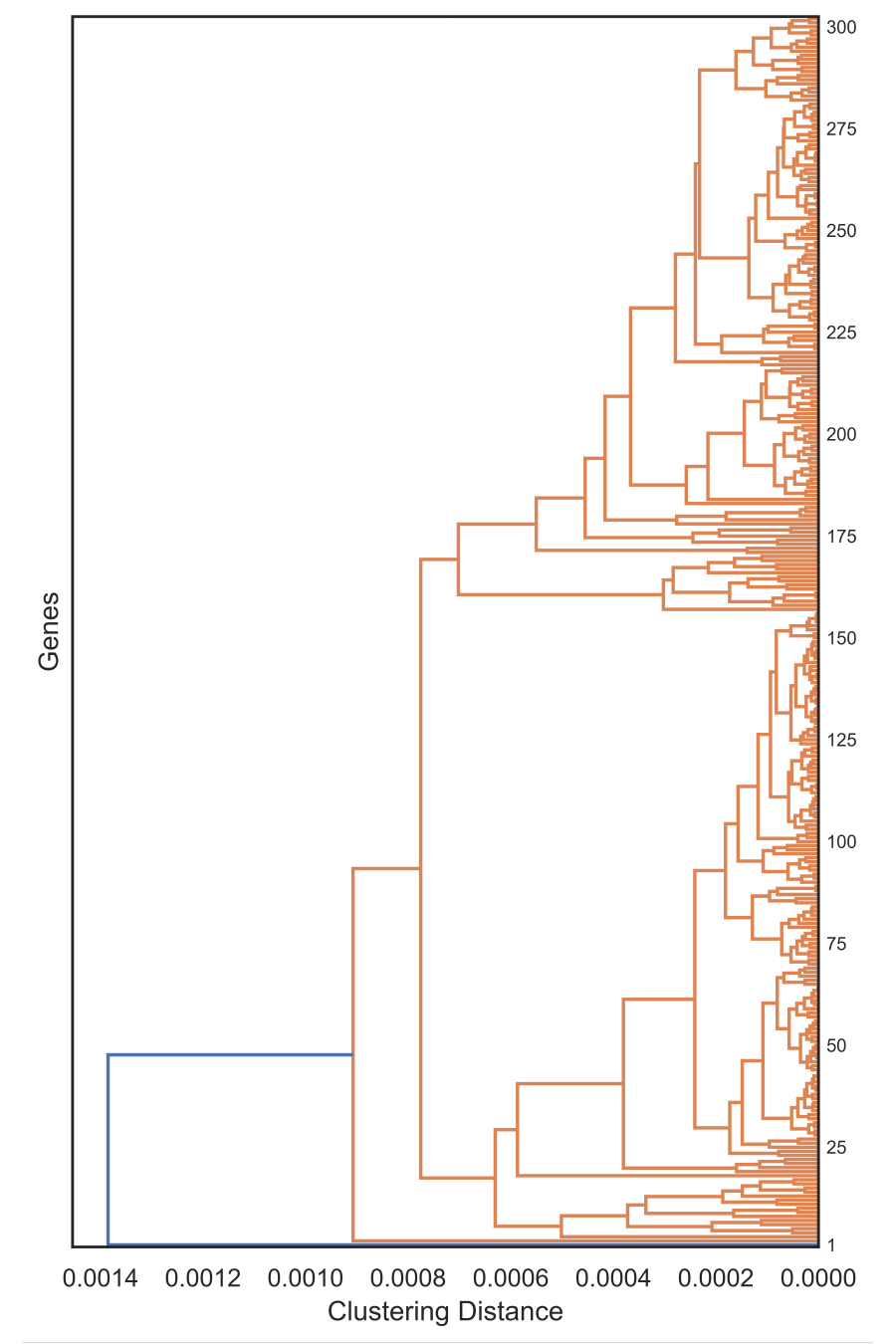}
    \caption{Prostate cancer gene dendrogram from SHAP profiles. Average-linkage clustering (Euclidean distance) of top predictive genes using their SHAP profiles across samples identifies modules associated with primary or metastatic disease states. Gene names are the supplementary Table~\ref{supp_table:pnet_gene_dendrogram_key}.}
    \label{supp_fig:pnet_gene_dendrogram}
\end{figure}

\input{files/pnet_gene_dendogram}

\FloatBarrier
\section{Clustering Metrics}
To assess the quality of the embedding spaces learned by each model, we used a set of standard clustering metrics that compare the learned clusters to the known biological labels. The adjusted Rand index (ARI) measures how often pairs of samples are grouped together or separated in the same way as the ground-truth labels, while correcting for random agreement. Normalized mutual information (NMI) quantifies how much information about the true labels is preserved by the clustering, and is insensitive to the absolute number of clusters.

We further report homogeneity and completeness to characterize the purity of clusters from two complementary perspectives. Homogeneity measures whether each predicted cluster contains samples from only a single true class, whereas completeness measures whether all samples from a given class are assigned to the same cluster. The V-measure is the harmonic mean of homogeneity and completeness, and it summarizes the trade-off between these two properties.

Finally, we compute the silhouette coefficient directly in the embedding space. This metric compares the average distance of each sample to points in the same cluster versus points in the nearest different cluster. High silhouette scores indicate compact, well-separated clusters in the underlying feature space, independent of the label information. Taken together, these metrics allow us to evaluate both label alignment (ARI, NMI, homogeneity, completeness, V-measure) and geometric separation (silhouette) of the learned embeddings.

Let $\mathcal{U} = \{U_1,\dots,U_R\}$ be the ground–truth partition of $N$ samples
and $\mathcal{V} = \{V_1,\dots,V_C\}$ be the clustering obtained from an
embedding (e.g.\ Graph2Image or a GNN).  
Define the contingency table
\[
n_{ij} = |U_i \cap V_j|, \qquad
a_i = \sum_{j} n_{ij}, \qquad
b_j = \sum_{i} n_{ij}, \qquad
N = \sum_{i,j} n_{ij},
\]
and denote the binomial coefficient by
$\binom{n}{2} = \frac{n(n-1)}{2}$.

\subsection*{Adjusted Rand Index (ARI)}

The Rand index measures the agreement between two partitions in terms of
pairwise co–assignment.  The adjusted Rand index corrects this quantity
for chance:
\begin{equation}
\mathrm{ARI}(\mathcal{U},\mathcal{V})
=
\frac{
    \displaystyle
    \sum_{i,j} \binom{n_{ij}}{2}
    -
    \frac{
        \left(\displaystyle\sum_{i} \binom{a_i}{2}\right)
        \left(\displaystyle\sum_{j} \binom{b_j}{2}\right)
    }{\binom{N}{2}}
}{
    \displaystyle
    \tfrac{1}{2}
    \left[
        \sum_{i} \binom{a_i}{2}
        +
        \sum_{j} \binom{b_j}{2}
    \right]
    -
    \frac{
        \left(\displaystyle\sum_{i} \binom{a_i}{2}\right)
        \left(\displaystyle\sum_{j} \binom{b_j}{2}\right)
    }{\binom{N}{2}}
}.
\end{equation}
$\mathrm{ARI}=1$ indicates perfect agreement and $\mathrm{ARI}\approx 0$
corresponds to random labeling (negative values are possible when the
agreement is worse than random).

\subsection*{Normalized Mutual Information (NMI)}

The mutual information between $\mathcal{U}$ and $\mathcal{V}$ is
\begin{equation}
I(\mathcal{U};\mathcal{V})
=
\sum_{i=1}^{R}\sum_{j=1}^{C}
\frac{n_{ij}}{N}
\log\left(
\frac{n_{ij}/N}{(a_i/N)(b_j/N)}
\right),
\end{equation}
with entropies
\begin{equation}
H(\mathcal{U}) = -\sum_{i=1}^{R} \frac{a_i}{N}\log\frac{a_i}{N},
\qquad
H(\mathcal{V}) = -\sum_{j=1}^{C} \frac{b_j}{N}\log\frac{b_j}{N}.
\end{equation}
We use the arithmetic–mean normalized mutual information
(as in \texttt{sklearn}):
\begin{equation}
\mathrm{NMI}(\mathcal{U},\mathcal{V})
=
\frac{2 \, I(\mathcal{U};\mathcal{V})}{H(\mathcal{U}) + H(\mathcal{V})}.
\end{equation}
$\mathrm{NMI}$ ranges in $[0,1]$, where $1$ indicates that the two
partitions carry identical label information.

\subsection*{Homogeneity and Completeness}

Homogeneity measures whether each estimated cluster contains only members
of a single ground–truth class.  It is defined via the conditional entropy
of $\mathcal{U}$ given $\mathcal{V}$:
\begin{equation}
H(\mathcal{U}\mid\mathcal{V})
=
- \sum_{j=1}^{C}\sum_{i=1}^{R}
\frac{n_{ij}}{N}
\log
\frac{n_{ij}}{b_j},
\end{equation}
and
\begin{equation}
\mathrm{homogeneity}
=
1 - \frac{H(\mathcal{U}\mid\mathcal{V})}{H(\mathcal{U})}.
\end{equation}
Completeness measures whether all members of a given class are assigned
to the same cluster.  Using the conditional entropy of
$\mathcal{V}$ given $\mathcal{U}$,
\begin{equation}
H(\mathcal{V}\mid\mathcal{U})
=
- \sum_{i=1}^{R}\sum_{j=1}^{C}
\frac{n_{ij}}{N}
\log
\frac{n_{ij}}{a_i},
\end{equation}
we define
\begin{equation}
\mathrm{completeness}
=
1 - \frac{H(\mathcal{V}\mid\mathcal{U})}{H(\mathcal{V})}.
\end{equation}
Both scores lie in $[0,1]$; higher values indicate purer and more
coherent clusters, respectively.

\subsection*{V-measure}

The V-measure is the harmonic mean of homogeneity $h$ and completeness
$c$:
\begin{equation}
\mathrm{V\mbox{-}measure}
=
\frac{2 h c}{h + c},
\end{equation}
with $\mathrm{V\mbox{-}measure}=1$ if and only if both $h=1$ and $c=1$.

\subsection*{Silhouette Coefficient}

Let $d(\mathbf{x}_p,\mathbf{x}_q)$ be a distance between embedded points
(e.g.\ Euclidean distance).  For a point $i$ assigned to the cluster
$C(i)$, define
\begin{align}
a(i) &= \frac{1}{|C(i)|-1}
\sum_{\substack{j \in C(i)\\ j \neq i}}
d(\mathbf{x}_i,\mathbf{x}_j),
\\
b(i) &= \min_{C' \neq C(i)}
\frac{1}{|C'|}
\sum_{j \in C'} d(\mathbf{x}_i,\mathbf{x}_j),
\end{align}
where $a(i)$ is the average distance to points in the same cluster and
$b(i)$ is the minimum average distance to points in any other cluster.
The silhouette of point $i$ is
\begin{equation}
s(i)
=
\frac{b(i) - a(i)}{\max\{a(i), b(i)\}},
\end{equation}
and the overall silhouette coefficient is the mean over all points:
\begin{equation}
\mathrm{Silhouette}
=
\frac{1}{N} \sum_{i=1}^{N} s(i).
\end{equation}
The silhouette lies in $[-1,1]$, with values close to $1$ indicating
well-separated, compact clusters.

\FloatBarrier

\begin{table*}[pb]
    \centering
    \caption{Clustering quality of embedding spaces learned by Graph2Image (CNN-based) and graph neural networks (GNNs) on five datasets. Higher is better for all metrics.}
    \label{supp_tab:emb_metrics}
    \begin{tabular}{llrrrrrr}
        \toprule
        Dataset & Model & ARI & NMI & Homogeneity & Completeness & V-measure & Silhouette \\
        \midrule
        \multirow{4}{*}{PP-Pathways} 
            & GCN          & 0.032 & 0.137 & 0.146 & 0.129 & 0.137 & 0.115 \\
            & GAT          & 0.020 & 0.104 & 0.111 & 0.097 & 0.104 & 0.094 \\
            & GIN          & 0.009 & 0.037 & 0.022 & 0.113 & 0.037 & \textbf{0.850} \\
            & Graph2Image  & \textbf{0.176} & \textbf{0.497} & \textbf{0.527} & \textbf{0.470} & \textbf{0.497} & 0.054 \\
        \midrule
        \multirow{4}{*}{HuRI} 
            & GCN          & 0.092 & 0.184 & 0.176 & 0.193 & 0.184 & 0.145 \\
            & GAT          & 0.016 & 0.035 & 0.035 & 0.035 & 0.035 & 0.156 \\
            & GIN          & 0.035 & 0.018 & 0.015 & 0.022 & 0.018 & \textbf{0.523} \\
            & Graph2Image  & \textbf{0.281} & \textbf{0.475} & \textbf{0.480} & \textbf{0.469} & \textbf{0.475} & 0.140 \\
        \midrule
        \multirow{4}{*}{Pan} 
            & GCN          & 0.642 & 0.805 & 0.810 & 0.800 & 0.805 & 0.617 \\
            & GAT          & 0.626 & 0.768 & 0.776 & 0.760 & 0.768 & 0.425 \\
            & GIN          & 0.394 & 0.684 & 0.655 & 0.717 & 0.684 & 0.586 \\
            & Graph2Image  & \textbf{0.867} & \textbf{0.926} & \textbf{0.935} & \textbf{0.917} & \textbf{0.926} & 0.504 \\
        \midrule
        \multirow{4}{*}{TM} 
            & GCN          & \textbf{0.635} & 0.775 & 0.811 & 0.741 & 0.775 & 0.364 \\
            & GAT          & \textbf{0.634} & 0.786 & 0.812 & 0.761 & 0.786 & 0.320 \\
            & GIN          & 0.455 & 0.613 & 0.547 & 0.698 & 0.613 & 0.435 \\
            & Graph2Image  & 0.519 & \textbf{0.828} & \textbf{0.911} & \textbf{0.752} & \textbf{0.824} & 0.270 \\
        \midrule
        \multirow{4}{*}{Prostate Cancer} 
            & GCN          & 0.057 & 0.075 & 0.044 & 0.251 & 0.075 & 0.733 \\
            & GAT          & -0.008 & 0.000 & 0.000 & 0.000 & 0.000 & 0.483 \\
            & GIN          & 0.051 & 0.050 & 0.053 & 0.047 & 0.050 & 0.614 \\
            & Graph2Image  & \textbf{0.669} & \textbf{0.606} & \textbf{0.561} & \textbf{0.655} & \textbf{0.604} & \textbf{0.926} \\
        \bottomrule
    \end{tabular}
\end{table*}

\clearpage

\begin{figure}[pt]
    \centering
    \includegraphics[width=0.75\linewidth, page=21]{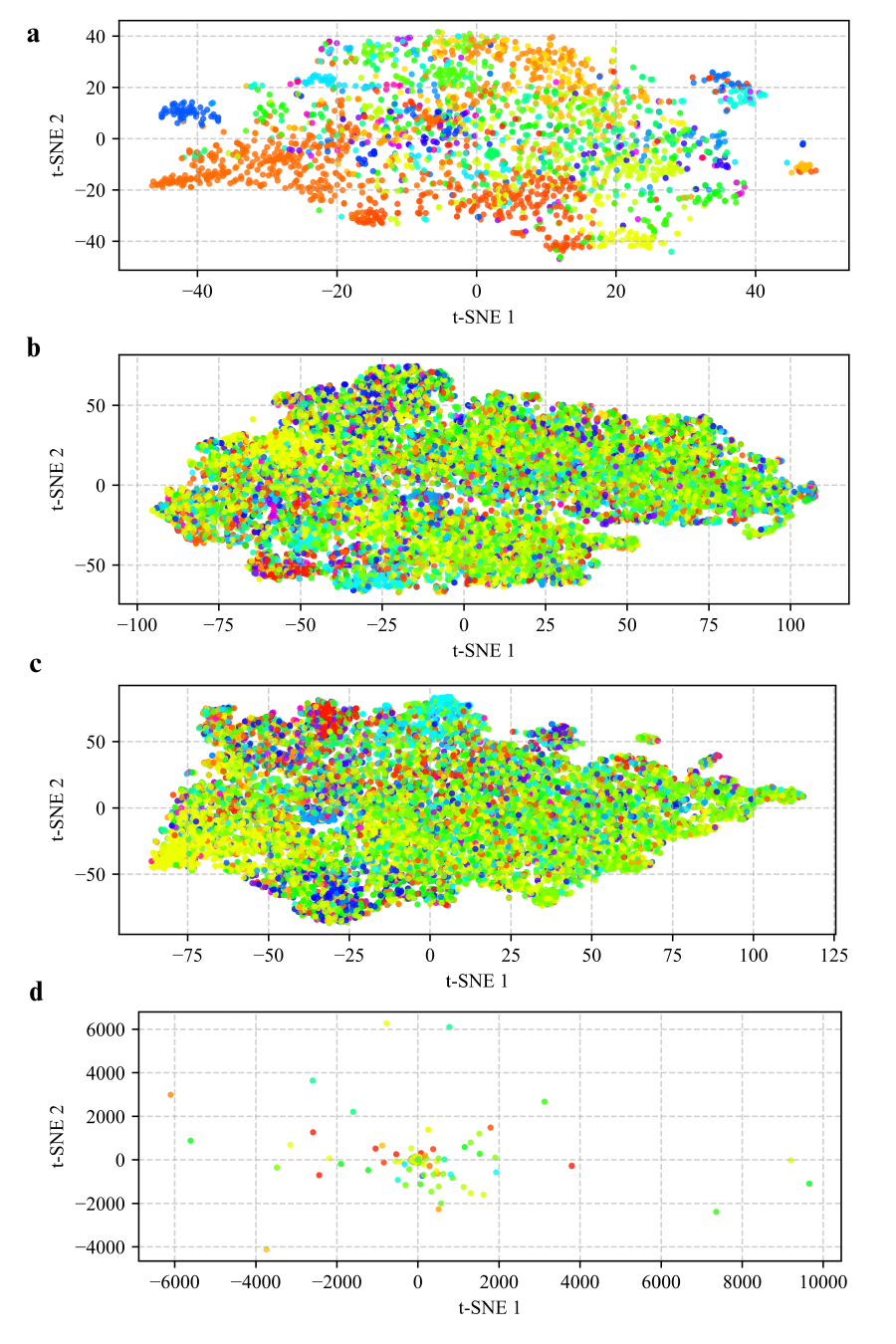}
    \caption{
    t-SNE visualisation of embedding spaces learned by Graph2Image and graph neural network (GNN) baselines on the PP-Pathways interactome.
    Two-dimensional t-SNE projections of (a) Graph2Image (CNN-based) embedding, (b) GAT, (c) GCN and (d) GIN, coloured by GTEx tissue label. Graph2Image yields a markedly more structured and compact embedding space with discernible grouping of related tissues, whereas GNN baselines produce diffuse or overlapping manifolds with weaker separation. These differences indicate that Graph2Image captures tissue-specific expression patterns more coherently in the PP-Pathway protein–protein interaction network. The legend is provided in Figure~\ref{supp_figure:ppathway_embeddings_legend}.
    }
    \label{supp_figure:ppathway_embeddings}
\end{figure}

\begin{figure}[pt]
    \centering
    \includegraphics[width=0.8\linewidth, page=22]{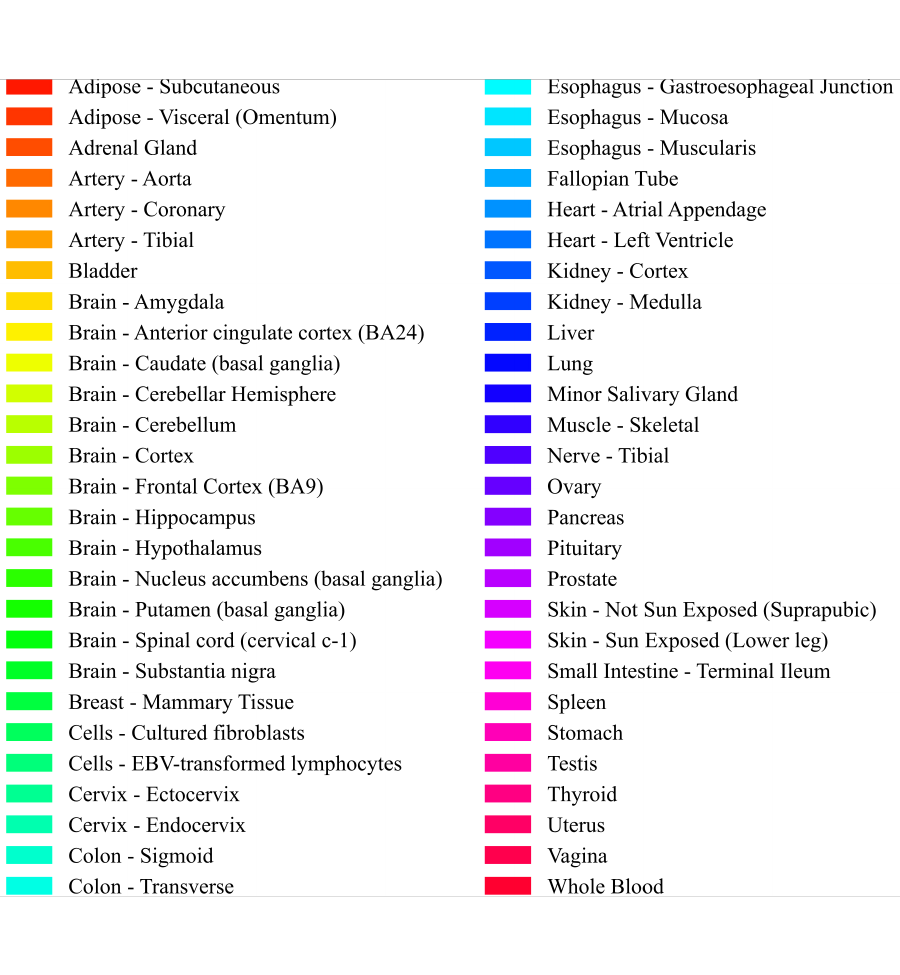}
    \caption{
    Legend for t-SNE visualisation of embedding spaces learned by Graph2Image and graph neural network for the PP-Pathways dataset.
    }
    \label{supp_figure:ppathway_embeddings_legend}
\end{figure}

\begin{figure}[pt]
    \centering
    \includegraphics[width=0.8\linewidth, page=23]{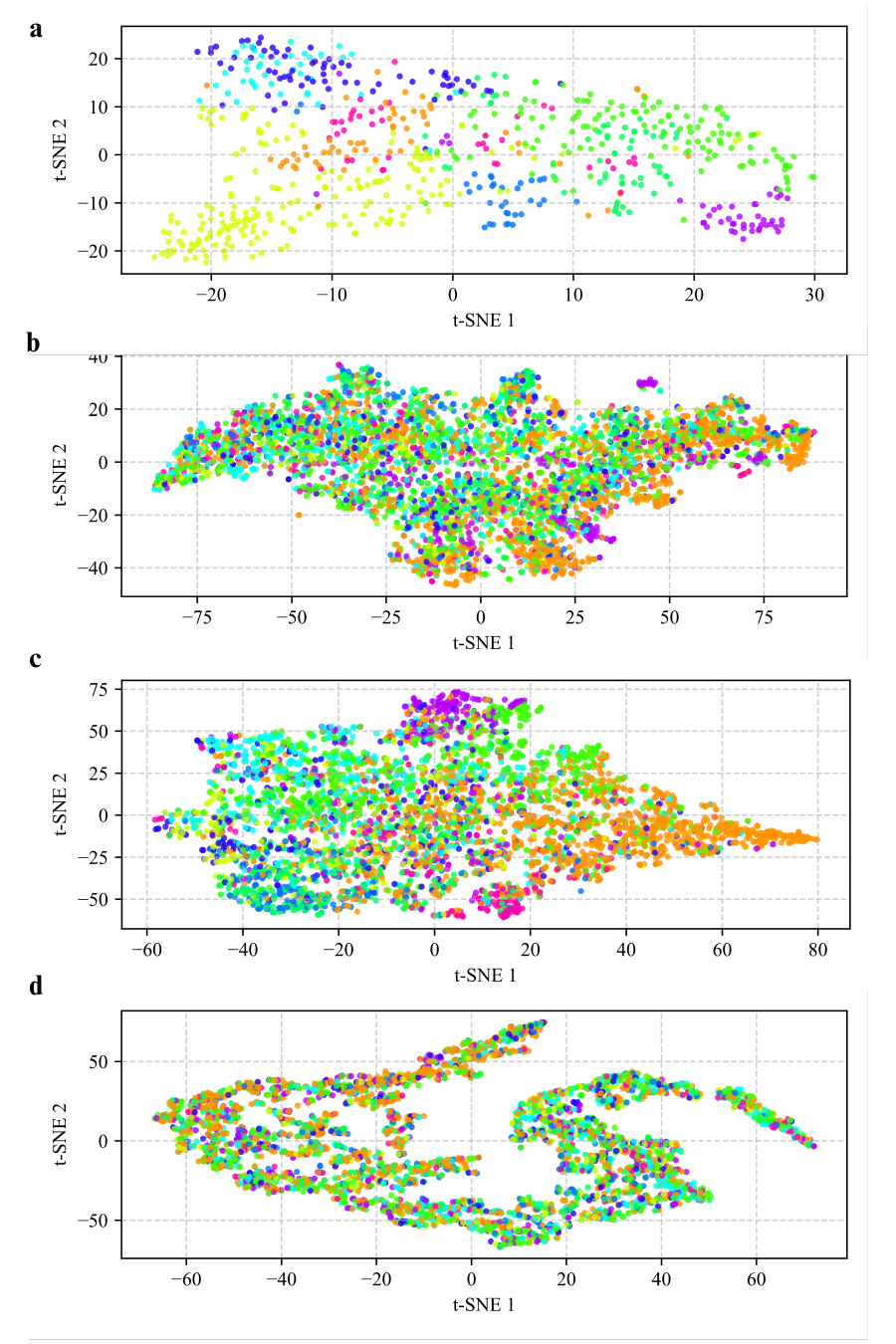}
    \caption{
    t-SNE visualisation of embedding spaces learned by Graph2Image and graph neural network (GNN) baselines on the HuRI tissue dataset.
    Two-dimensional t-SNE projections of (a) Graph2Image (CNN-based) embedding, (b) GAT, (c) GCN and (d) GIN, coloured by GTEx tissue label. Graph2Image produces a substantially more organised and tissue-coherent embedding space, with smoother manifolds and clearer separation between major GTEx tissue groups. In contrast, the GNN baselines generate diffuse, intermingled, or fragmented structures, suggesting weaker preservation of tissue-specific transcriptomic signatures within the HuRI protein–protein network. The legend is provided in Figure~\ref{supp_figure:huri_embeddings}.
    }
    \label{supp_figure:huri_embeddings}
\end{figure}

\begin{figure}[pt]
    \centering
    \includegraphics[width=0.8\linewidth, page=24]{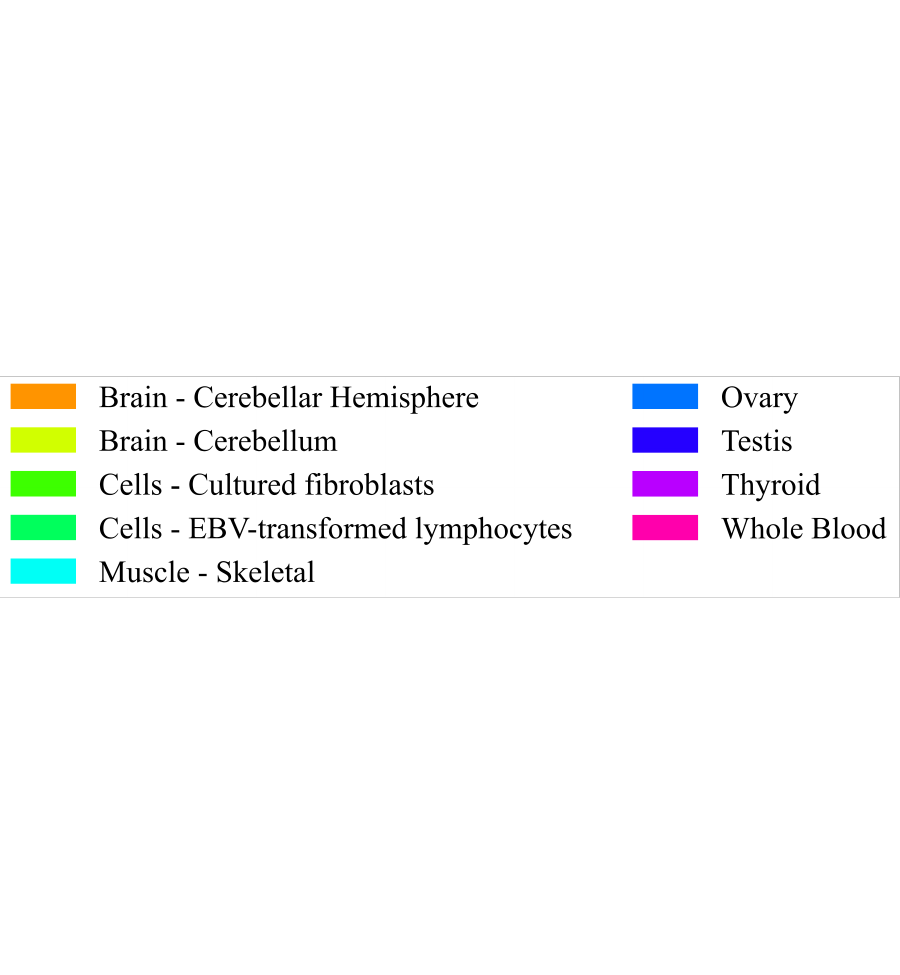}
    \caption{
    Legend for t-SNE visualisation of embedding spaces learned by Graph2Image and graph neural network for the HuRI dataset.
    }
    \label{supp_figure:huri_embeddings_legend}
\end{figure}

\begin{figure}[pt]
    \centering
    \includegraphics[width=0.8\linewidth, page=25]{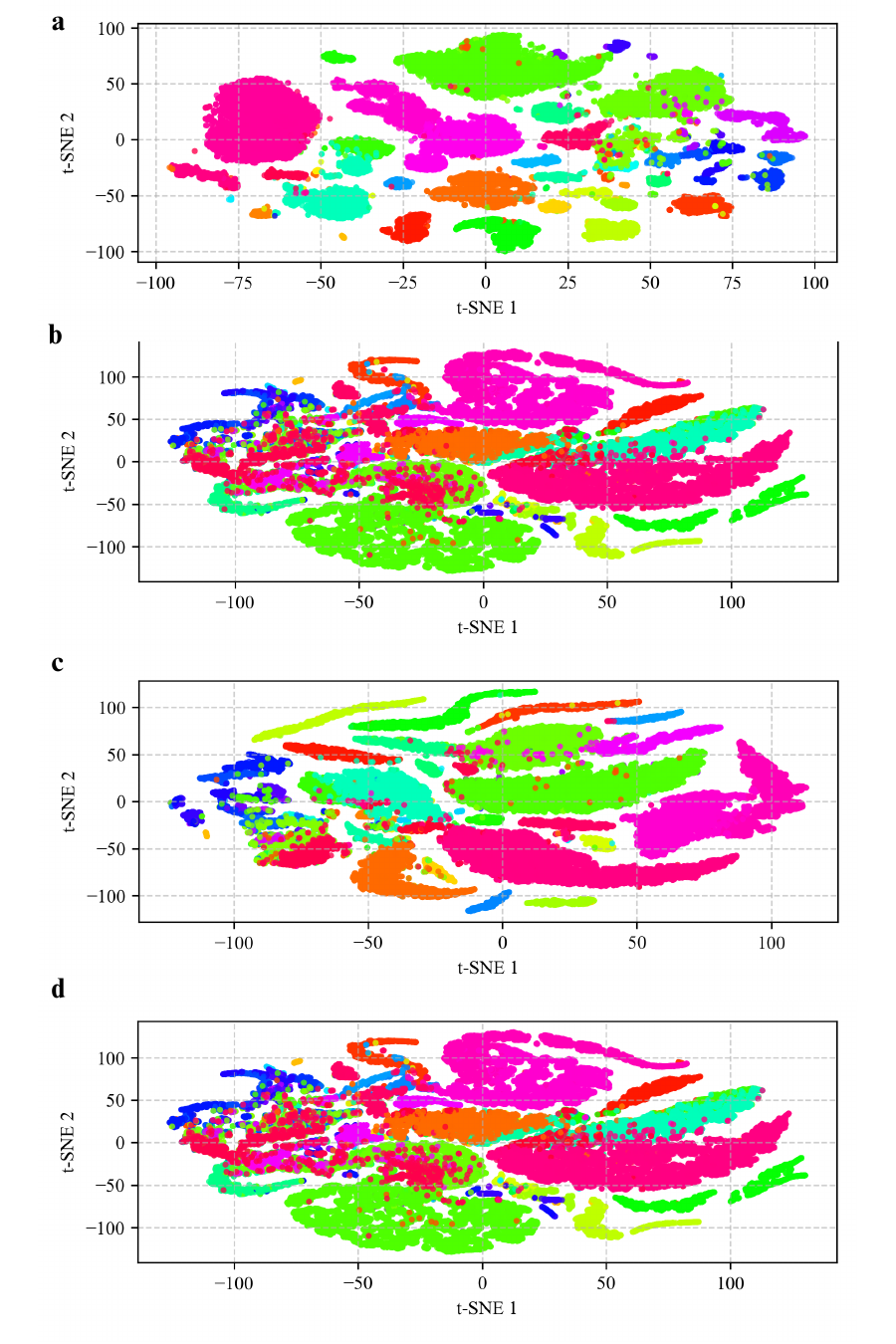}
    \caption{
    t-SNE visualisation of embedding spaces learned by Graph2Image and graph neural network (GNN) baselines on the TM single-cell dataset. Two-dimensional t-SNE projections of (a) Graph2Image (CNN-based) embedding, (b) GAT, (c) GCN and (d) GIN, coloured by annotated cell type. Graph2Image produces compact, well-separated clusters that closely follow the biological annotations, whereas GNN embeddings often form elongated or partially overlapping manifolds, indicating a less label-consistent feature space. The legend is in Figure~\ref{supp_figure:tm_embeddings_legend}
    }
    \label{supp_figure:tm_embeddings}
\end{figure}

\begin{figure}[pt]
    \centering
    \includegraphics[width=0.8\linewidth, page=26]{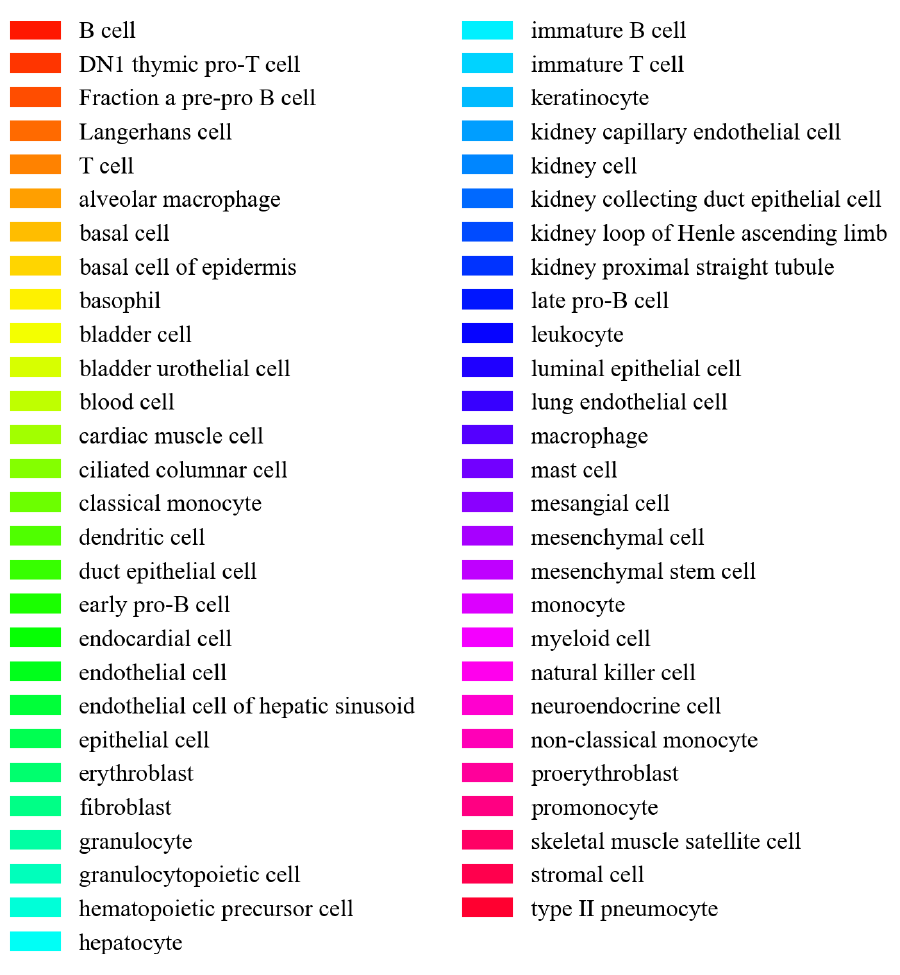}
    \caption{
    Legend for t-SNE visualisation of embedding spaces learned by Graph2Image and graph neural network for the TM dataset.
    }
    \label{supp_figure:tm_embeddings_legend}
\end{figure}

\begin{figure}[pt]
    \centering
    \includegraphics[width=0.8\linewidth, page=27]{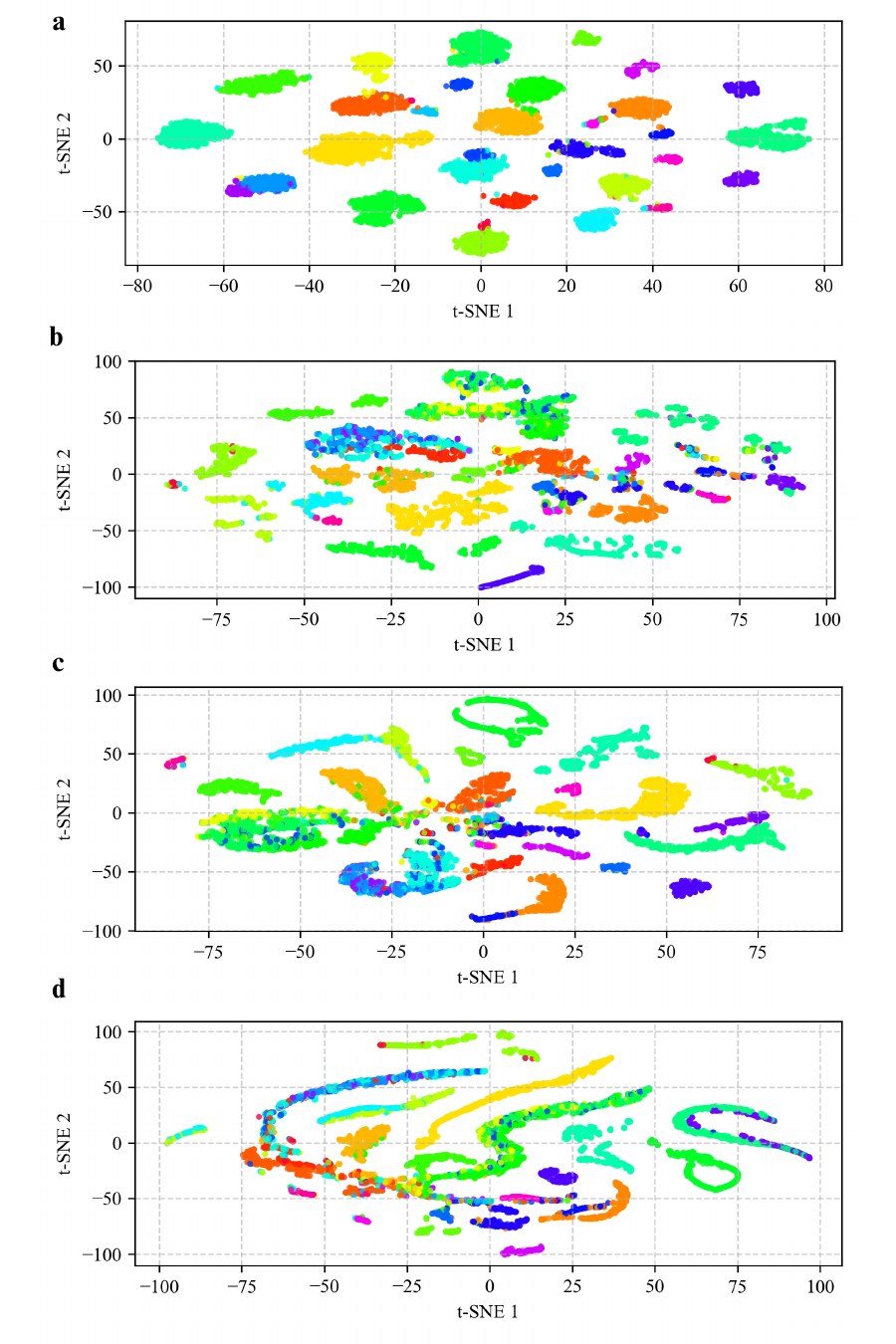}
    \caption{
    t-SNE visualisation of embedding spaces learned by Graph2Image and graph neural network (GNN) baselines on the Pan-cancer (Pan) cohort. Two-dimensional t-SNE projections of (a) Graph2Image (CNN-based) embedding, (b) GAT, (c) GCN and (d) GIN, coloured by tumour type. Graph2Image yields discrete, well-separated clusters for individual cancer types, while GNN embeddings exhibit stretched or interwoven manifolds with substantial overlap between tumours, consistent with reduced clustering purity. The legend is in Figure~\ref{supp_figure:pan_embeddings_legend}
    }
    \label{supp_figure:pan__embeddings}
\end{figure}

\begin{figure}[pt]
    \centering
    \includegraphics[width=0.5\linewidth, page=28]{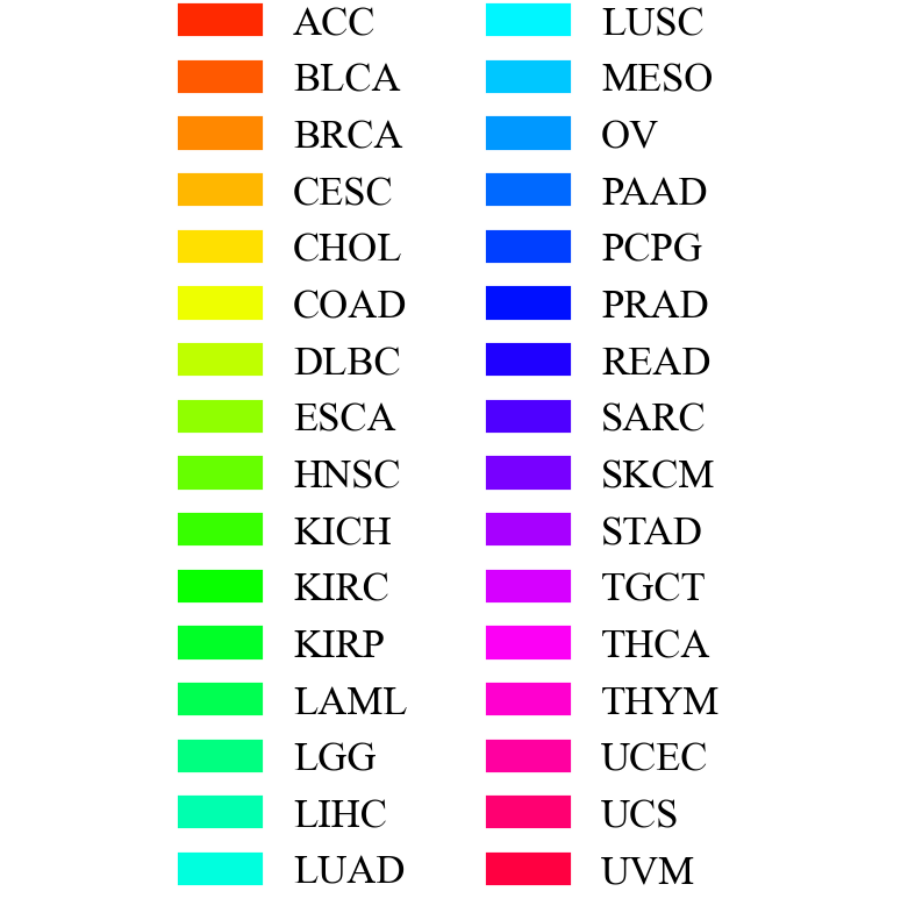}
    \caption{
    Legend for t-SNE visualisation of embedding spaces learned by Graph2Image and graph neural network for the Pan Cancer dataset.
    }
    \label{supp_figure:pan_embeddings_legend}
\end{figure}

\begin{figure}[pt]
    \centering
    \includegraphics[width=0.8\linewidth, page=29]{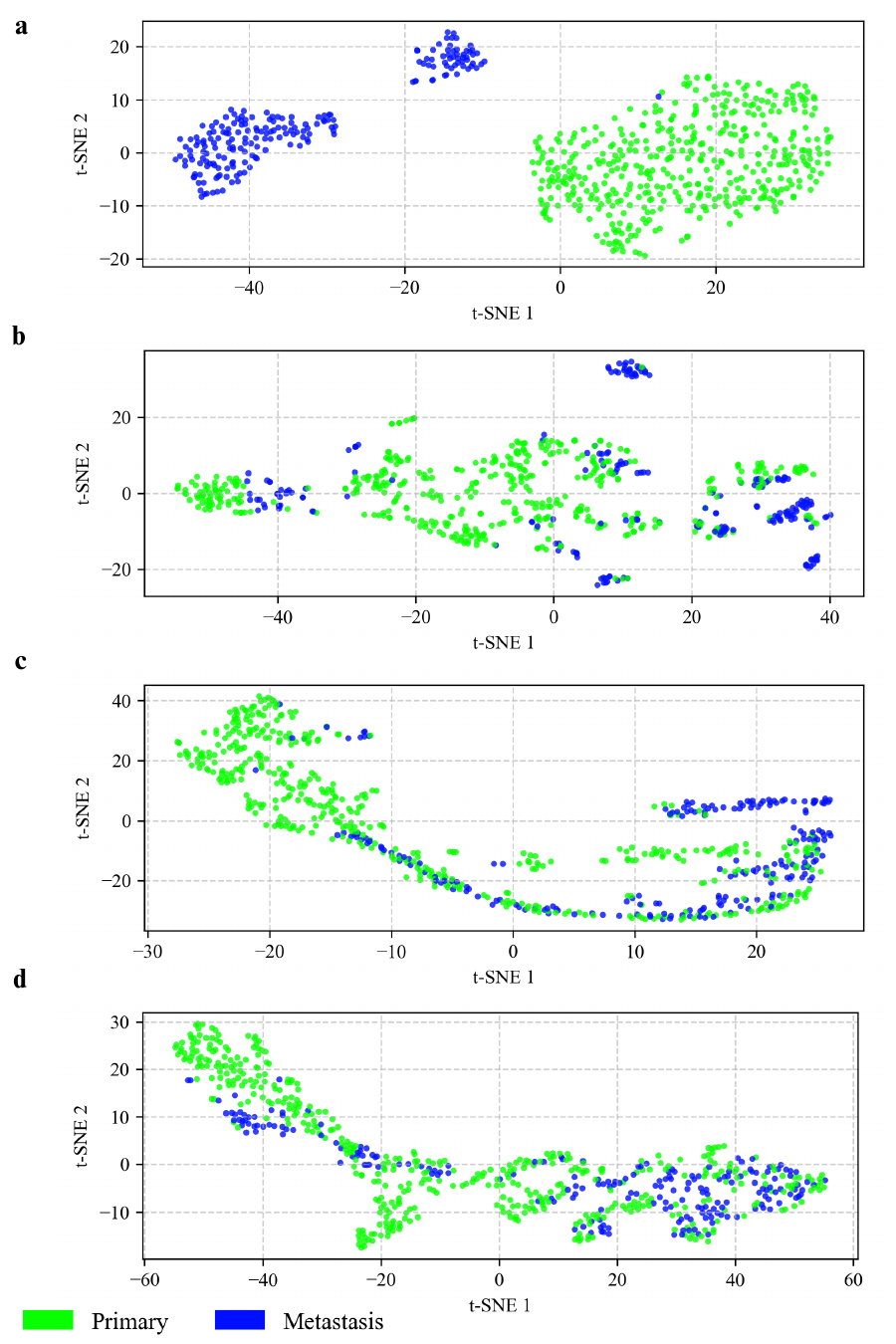}
    \caption{
    t-SNE visualisation of embedding spaces learned by Graph2Image and graph neural network (GNN) baselines on the Prostate Cancer primary--metastasis cohort. Two-dimensional t-SNE projections of (a) Graph2Image (CNN-based) embedding, (b) GAT, (c) GCN and (d) GIN, coloured by clinical label (Primary vs.\ Metastasis). Graph2Image clearly separates primary tumours and metastatic lesions into disjoint clusters, whereas GNN embeddings show substantial mixing of the two labels and fragmented manifolds, indicating weaker alignment between the learned feature space and clinical outcomes.
    }
    \label{supp_figure:pnet_embeddings}
\end{figure}

%% file: files/ppathway_class_names.tex
\begin{table}[p]
  \centering
    \caption{GTEx tissue classes for the PP-Pathways dataset}
    \label{supp_tab:pppathways_classes}
    \begin{tabular}{cl}
        \toprule
        \textbf{Index} & \textbf{Tissue name} \\
        \midrule
         1 & Adipose -- Subcutaneous \\
         2 & Adipose -- Visceral (Omentum) \\
         3 & Adrenal Gland \\
         4 & Artery -- Aorta \\
         5 & Artery -- Coronary \\
         6 & Artery -- Tibial \\
         7 & Bladder \\
         8 & Brain -- Amygdala \\
         9 & Brain -- Anterior cingulate cortex (BA24) \\
        10 & Brain -- Caudate (basal ganglia) \\
        11 & Brain -- Cerebellar Hemisphere \\
        12 & Brain -- Cerebellum \\
        13 & Brain -- Cortex \\
        14 & Brain -- Frontal Cortex (BA9) \\
        15 & Brain -- Hippocampus \\
        16 & Brain -- Hypothalamus \\
        17 & Brain -- Nucleus accumbens (basal ganglia) \\
        18 & Brain -- Putamen (basal ganglia) \\
        19 & Brain -- Spinal cord (cervical c-1) \\
        20 & Brain -- Substantia nigra \\
        21 & Breast -- Mammary Tissue \\
        22 & Cells -- Cultured fibroblasts \\
        23 & Cells -- EBV-transformed lymphocytes \\
        24 & Cervix -- Ectocervix \\
        25 & Cervix -- Endocervix \\
        26 & Colon -- Sigmoid \\
        27 & Colon -- Transverse \\
        28 & Esophagus -- Gastroesophageal Junction \\
        29 & Esophagus -- Mucosa \\
        30 & Esophagus -- Muscularis \\
        31 & Fallopian Tube \\
        32 & Heart -- Atrial Appendage \\
        33 & Heart -- Left Ventricle \\
        34 & Kidney -- Cortex \\
        35 & Kidney -- Medulla \\
        36 & Liver \\
        37 & Lung \\
        38 & Minor Salivary Gland \\
        39 & Muscle -- Skeletal \\
        40 & Nerve -- Tibial \\
        41 & Ovary \\
        42 & Pancreas \\
        43 & Pituitary \\
        44 & Prostate \\
        45 & Skin -- Not Sun Exposed (Suprapubic) \\
        46 & Skin -- Sun Exposed (Lower leg) \\
        47 & Small Intestine -- Terminal Ileum \\
        48 & Spleen \\
        49 & Stomach \\
        50 & Testis \\
        51 & Thyroid \\
        52 & Uterus \\
        53 & Vagina \\
        54 & Whole Blood \\
        \bottomrule
    \end{tabular}
\end{table}

%% file: files/huri_class_names.tex
\begin{table}[pt]
    \centering
    \caption{GTEx tissue classes for the HuRI dataset.}
    \label{supp_tab:huri_classes}
    \begin{tabular}{cl}
        \toprule
        \textbf{Index} & \textbf{Tissue name} \\
        \midrule
         1 & Brain -- Cerebellar Hemisphere \\
         2 & Brain -- Cerebellum \\
         3 & Cells -- Cultured fibroblasts \\
         4 & Cells -- EBV-transformed lymphocytes \\
         5 & Muscle -- Skeletal \\
         6 & Ovary \\
         7 & Testis \\
         8 & Thyroid \\
         9 & Whole Blood \\
        \bottomrule
    \end{tabular}
\end{table}

%% file: files/huri_gene_names.tex
\begin{table}[pt]
    \centering
    \caption{Index mapping for HuRI tissues.}
    \label{tab:pppathways_index_mapping}
    \begin{tabular}{cl}
        \toprule
        \textbf{Index} & \textbf{Tissue name} \\
        \midrule
         1 & Adipose -- Subcutaneous \\
         2 & Adipose -- Visceral (Omentum) \\
         3 & Adrenal Gland \\
         4 & Artery -- Aorta \\
         5 & Artery -- Coronary \\
         6 & Artery -- Tibial \\
         7 & Bladder \\
         8 & Brain -- Amygdala \\
         9 & Brain -- Anterior cingulate cortex (BA24) \\
        10 & Brain -- Caudate (basal ganglia) \\
        11 & Brain -- Cerebellar Hemisphere \\
        12 & Brain -- Cerebellum \\
        13 & Brain -- Cortex \\
        14 & Brain -- Frontal Cortex (BA9) \\
        15 & Brain -- Hippocampus \\
        16 & Brain -- Hypothalamus \\
        17 & Brain -- Nucleus accumbens (basal ganglia) \\
        18 & Brain -- Putamen (basal ganglia) \\
        19 & Brain -- Spinal cord (cervical c-1) \\
        20 & Brain -- Substantia nigra \\
        21 & Breast -- Mammary Tissue \\
        22 & Cells -- Cultured fibroblasts \\
        23 & Cells -- EBV-transformed lymphocytes \\
        24 & Cervix -- Ectocervix \\
        25 & Cervix -- Endocervix \\
        26 & Colon -- Sigmoid \\
        27 & Colon -- Transverse \\
        28 & Esophagus -- Gastroesophageal Junction \\
        29 & Esophagus -- Mucosa \\
        30 & Esophagus -- Muscularis \\
        31 & Fallopian Tube \\
        32 & Heart -- Atrial Appendage \\
        33 & Heart -- Left Ventricle \\
        34 & Kidney -- Cortex \\
        35 & Kidney -- Medulla \\
        36 & Liver \\
        37 & Lung \\
        38 & Minor Salivary Gland \\
        39 & Muscle -- Skeletal \\
        40 & Nerve -- Tibial \\
        41 & Ovary \\
        42 & Pancreas \\
        43 & Pituitary \\
        44 & Prostate \\
        45 & Skin -- Not Sun Exposed (Suprapubic) \\
        46 & Skin -- Sun Exposed (Lower leg) \\
        47 & Small Intestine -- Terminal Ileum \\
        48 & Spleen \\
        49 & Stomach \\
        50 & Testis \\
        51 & Thyroid \\
        52 & Uterus \\
        53 & Vagina \\
        54 & Whole Blood \\
        \bottomrule
    \end{tabular}
\end{table}

%% file: files/tm_class_names.tex

\begin{table}[p]
  \centering
  \caption{Tabula Muris dataset cell names}
  \label{supp_table:tm_class_name}
  \begin{tabular}{llll}
    \hline
    1: keratinocyte & 2: neuroendocrine cell & 3: duct epithelial cell & 4: skeletal muscle satellite cell \\
    5: dendritic cell & 6: endothelial cell & 7: mesenchymal cell & 8: basal cell of epidermis \\
    9: luminal epithelial cell & 10: stromal cell & 11: mesenchymal stem cell & 12: immature B cell \\
    13: hematopoietic precursor cell & 14: T cell & 15: granulocytopoietic cell & 16: promonocyte \\
    17: myeloid cell & 18: blood cell & 19: Fraction a pre-pro B cell & 20: kidney capillary endothelial cell \\
    21: natural killer cell & 22: macrophage & 23: cardiac muscle cell & 24: epithelial cell \\
    25: alveolar macrophage & 26: basal cell & 27: kidney cell & 28: B cell \\
    29: leukocyte & 30: DN1 thymic pro-T cell & 31: immature T cell & 32: Langerhans cell \\
    33: granulocyte & 34: bladder cell & 35: kidney loop of Henle ascending limb & 36: monocyte \\
    37: non-classical monocyte & 38: type II pneumocyte & 39: kidney collecting duct epithelial cell & 40: basophil \\
    41: mast cell & 42: mesangial cell & 43: endothelial cell of hepatic sinusoid & 44: lung endothelial cell \\
    45: endocardial cell & 46: fibroblast & 47: ciliated columnar cell & 48: classical monocyte \\
    49: erythroblast & 50: proerythroblast & 51: early pro-B cell & 52: late pro-B cell \\
    53: hepatocyte & 54: bladder urothelial cell & 55: kidney proximal straight tubule \\
    \hline
  \end{tabular}
\end{table}

%% file: files/tm_gene_dendogram.tex

\begin{table}[p]
  \centering
  \caption{Tabula Muris SHAP gene list (part 1 of 3; first 400 genes).}
  \label{supp_table:tm_shap_genes_part_1}
  \begin{tabular}{llllllll}
    \hline
    Fcer1g & Ubb & Cd79a & Cd79b & Cd3g & Bst2 & Il1b & Vpreb3 \\
    Cfh & Cst3 & Fxyd5 & Tmsb10 & Laptm5 & Lsp1 & Gnas & Gng11 \\
    S100a4 & Alox5ap & Ly6c1 & Plvap & Ctsl & Irf8 & H2\_K1 & Tpm2 \\
    Klf2 & Lyz2 & Sepp1 & Ccl3 & Vim & Socs3 & Lyz1 & Igfbp2 \\
    Krt14 & Ramp2 & Ckb & Oaz1 & Gm1821 & Marcksl1 & Arpc2 & Arpc3 \\
    Hbb\_b2 & Car2 & Scgb1a1 & Prss34 & Ccl9 & Defb1 & Tyrobp & Gzma \\
    AW112010 & Nkg7 & Ifitm1 & Tg & H2\_D1 & Emp3 & Hmgb2 & Lgals1 \\
    Dek & Ly6c2 & Coro1a & Slc12a1 & Fbln1 & Ccl6 & Ctsd & Hmgn2 \\
    S100a8 & Krt19 & Shisa5 & Rps15a\_ps6 & Clic1 & Rps20 & B2m & Cmtm7 \\
    Igj & Crip1 & Igfbp7 & Ifitm3 & Fabp4 & Sparcl1 & Selk & Cd81 \\
    Jund & Capzb & Krt7 & Serp1 & Cd52 & A130040M12Rik & Ctsb & Sln \\
    Myl4 & Nppa & Ifitm6 & Retnlg & S100a9 & x1100001G20Rik & Laptm4a & Rhob \\
    Tmsb4x & Cxcl15 & Sftpb & Sftpc & Ly6a & Atp1b1 & Ly6e & Ppap2b \\
    Tuba1a & Clec3b & Sod3 & Pcolce2 & Tnfaip6 & Dntt & Ptprcap & Grcc10 \\
    Sh3bgrl3 & Pck1 & Slc34a1 & Errfi1 & Mt2 & Btg1 & Hspa1a & Ccnl1 \\
    Meg3 & Vcam1 & Id3 & Ndufa2 & x1110003E01Rik & Upk1a & Upk1b & Ifitm2 \\
    Cfl1 & Rpl37 & Gadd45b & Zfp36 & Jun & Rps6 & Postn & Rpl13 \\
    Rps16 & Dpt & Ppp1r15a & Rpl7 & Srsf5 & Npc2 & Cryab & Gatm \\
    Pabpc1 & Rps26 & Atp5k & Cyc1 & Tceb2 & Sec61g & x2410015M20Rik & Actr3 \\
    Ubl5 & Mdh2 & Vdac1 & Atox1 & Aplp2 & Cox6c & Lamp1 & Cdc42 \\
    x1190002H23Rik & Cks2 & Ccl7 & Cyr61 & Aqp3 & Perp & Cd9 & Atp5d \\
    Tbca & Mif & Pim3 & Eif6 & Trim29 & Jup & Krt6a & Vsig8 \\
    Rab25 & Fabp5 & Ly6d & Rtn4 & Nfkbia & D14Ertd449e & Rpl9 & Rpl27a \\
    Rps10 & Eif4a1 & Lgals3 & x2010001M09Rik & Chchd10 & Ndufc2 & Psap & Ier2 \\
    Sepw1 & Psmb8 & Rhoa & Gabarap & Arpc5 & Tmed10 & Mal & Tpt1 \\
    Lum & Csrp1 & H3f3b & Cd24a & S100a11 & Cotl1 & Taldo1 & Ndufa7 \\
    Actb & Rpl37a & Rplp2 & Serinc3 & Tmem176a & Gdi2 & H2\_Aa & Rps15a\_ps4 \\
    Btg2 & Ccl17 & Tmem176b & Atf4 & Fosb & Gadd45g & Rnase4 & Cbr2 \\
    Ptpn18 & Rps2 & Rps11 & Rps17 & Atp5b & Rpl8 & Rplp1 & Txn1 \\
    Rpl18a & Rps8 & Sat1 & Ndufb10 & Chchd2 & Ccdc72 & Cox8a & Eif4g2 \\
    Atp5a1 & Btf3 & Ndufv3 & Myl12a & Rnasek & Foxq1 & Ivl & S100a6 \\
    Wfdc2 & Rplp0 & Pdlim1 & Rpl19 & Sdc1 & Igfbp3 & Gltscr2 & Krt5 \\
    Hspa8 & Rpl15 & S100a10 & Mrfap1 & Pkm2 & Psmb3 & Sfn & Csnk1a1 \\
    Rbm3 & Rps5 & Txndc17 & Cldn5 & H2\_Ab1 & Cd74 & H2\_Eb1 & Lmna \\
    Cd164 & Krt8 & H3f3a & Romo1 & Hmgn1 & Rps28 & H2afv & Mdh1 \\
    Eif5a & Rpl23 & Rps24 & Aldh2 & Rpl22l1 & Tomm6 & Fos & Gpx1 \\
    Arpc1b & Shfm1 & Klf9 & Cox7a2l & Rpl27 & Rpl18 & x1500012F01Rik & Eef1a1 \\
    Atf3 & Cct4 & Adamts1 & Pa2g4 & Atp6v0e & Timp3 & Rpl32 & Rpl36al \\
    Rps15 & Rpl41 & Rps27 & Hnrnpk & Ran & Pcbp1 & Usmg5 & Cycs \\
    Cnbp & Eif2s2 & Map1lc3a & Myl6 & Cyba & Sec11c & Hint1 & Birc5 \\
    Anp32e & Cenpa & Eif1 & Zfp706 & Atp1a1 & Bag1 & Rpl5 & Rpl35 \\
    Ifi27l2a & Atp6v0b & Slc38a2 & Ier5 & Pnrc1 & Tpm4 & Vtn & Apoa1 \\
    Akr1c6 & Gnmt & Srp9 & Nfe2l2 & x2310003F16Rik & Gsr & Hp & Cyp2f2 \\
    S100a1 & x1810037I17Rik & Rac2 & Brp44 & Nr4a1 & Egr1 & Arf1 & Ndufb4 \\
    Hist1h2bc & Hras1 & Sepx1 & Apoe & Ifrd1 & Prdx5 & Umod & Clta \\
    Hnrnpf & Rpl4 & Rps27a & Use1 & Tmed2 & Bsg & Tomm7 & Tpm3 \\
    Uqcrc1 & Ctla2a & Gmfg & Gas6 & Rpl35a & Pcbp2 & Sgk1 & Rps9 \\
    Psca & x2700060E02Rik & Rps14 & Gnb2l1 & Rpl29 & Eef2 & Lsm4 & Dusp1 \\
    Junb & Nme2 & Rpl11 & C3 & Igfbp4 & Tmem123 & Bgn & Wif1 \\
    Myl9 & Elane & Ms4a3 & Anxa1 & Glrx & H2afx & Mki67 & Pfn1 \\
    \hline
  \end{tabular}
\end{table}

\begin{table}[p]
  \centering
  \caption{Tabula Muris SHAP gene list (part 2 of 3; next 400 genes).}
  \label{supp_table:tm_shap_genes_part_2}
  \begin{tabular}{llllllll}
    \hline
    Park7 & Rpl17 & Sub1 & Ube2i & Erp29 & Mcpt8 & Icam1 & Tsc22d1 \\
    Atp5e & Fxyd3 & Rps18 & Ndufb9 & x3110003A17Rik & Dmkn & Uqcrfs1 & Rps19 \\
    Rps25 & Ltb & Hnrnpab & Plac9 & Dnaja1 & Itm2b & Id1 & Ier3 \\
    Rarres2 & Csrp2 & Mmp3 & Calm1 & Gnai2 & Pfdn5 & Sumo2 & Ctnnb1 \\
    Atp5j & Cox4i1 & Rpsa & Slc25a4 & Cct7 & Ddx5 & Atp5h & Rpl23a \\
    Snrpd2 & Tagln & Chmp2a & Ndufa4 & Ccl5 & Nbl1 & Rps21 & Uqcrb \\
    Rpl10 & Arpc1a & Rpl10a & Akr1a1 & Etfb & Pomp & Srgn & Ubc \\
    Ftl1 & Rpl28 & Rps27l & Stmn1 & Tubb5 & H2afy & Hist1h2ae & Ssr4 \\
    Ctsg & Top2a & Aes & Rpl21 & Rpl3 & Reep5 & Cct5 & Tmbim6 \\
    Rps4x & Ywhaq & Atp6v1g1 & Arl6ip1 & Aqp1 & Ldhb & Cd36 & Cxcl12 \\
    Cd14 & C1qa & C1qb & Ndufa8 & Edf1 & Hprt & Fkbp1a & Top1 \\
    Mfap5 & Mmp2 & Ccl11 & Pi16 & Col6a1 & Ugdh & Dcn & Serpinf1 \\
    Mt1 & Myoc & Smoc2 & Pcolce & Cebpd & Serping1 & Cxcl1 & Fgl2 \\
    Zfp36l1 & Rps29 & Tmem134 & D8Ertd738e & Slc25a5 & Smc4 & x2410006H16Rik & Chmp4b \\
    Hk2 & Ahnak & Sdc4 & Atp5o & Dynlrb1 & Dad1 & Aldoa & Map1lc3b \\
    Hdgf & Uqcrh & Cox6b1 & Nme1 & Rpl7a & Rps12 & Rps7 & Nhp2l1 \\
    Polr1d & Cdkn1a & Ywhaz & Rps3 & Maff & Ndufb8 & Klf6 & Mgst3 \\
    Rps23 & Gnb2 & Hspe1 & Rpl12 & Pgk1 & Crlf1 & Hist1h1c & Klf5 \\
    Phb2 & Arf5 & Avpi1 & Pebp1 & Psmb5 & Crip2 & Ptma & Prdx1 \\
    Psmb6 & Snrpb & Rpl6 & Uba52 & Cox6a1 & Nap1l1 & Hnrnpa2b1 & Ndufa12 \\
    Rps3a & Zfand5 & Eef1b2 & Eif3h & x1110008F13Rik & Vamp8 & Xist & Rpl31 \\
    H2afj & Tst & Dstn & Dbi & Slc25a3 & Pycard & Rnh1 & Pir \\
    Ppp1r14b & Serbp1 & x2610528A11Rik & Rpl14 & Arf6 & Dnajb1 & Cstb & Emp2 \\
    Rpl22 & Fosl1 & Pkp1 & Ovol1 & Hbegf & Nhp2 & Ndufa4l2 & Capg \\
    Calml3 & Them5 & Capns2 & Gpx2 & S100a14 & Ptgr1 & Lypd3 & Krt16 \\
    x2310033E01Rik & Slpi & Spink5 & Actg1 & Eef1d & Nedd8 & Ndufa3 & Plac8 \\
    Fau & Ndufb11 & Sqstm1 & Acta2 & Atp5l & Pdia3 & Spint2 & Gsto1 \\
    Ost4 & Rbx1 & Ppp1ca & Rps13 & Polr2l & Actn4 & Mgp & Swi5 \\
    Hsp90aa1 & Snrpg & Limd2 & Pkp3 & Srsf3 & Slc3a2 & Arpp19 & Tuba1b \\
    Hsp90b1 & Sec61b & Glrx3 & Rac1 & Ccnd3 & Ltf & Ngp & Ap3s1 \\
    Tkt & Gpi1 & Chi3l3 & Camp & Fcnb & Atpif1 & Minos1 & Ndufc1 \\
    Prdx2 & Gm11428 & Hmox1 & Atp5f1 & Ccl4 & Napsa & Ctss & x1500032L24Rik \\
    Alas2 & Pglyrp1 & Eif3i & Nop10 & Clic4 & Lgals7 & Metap2 & Ppia \\
    Eif3m & Tubb4b & Atp5g3 & Gja1 & Ezr & Ldha & Hspd1 & Anxa8 \\
    Gm94 & Ube2d3 & Uqcrq & Thy1 & Tm4sf1 & Glul & F3 & Krt18 \\
    Snrpe & Anp32b & H2afz & Cltb & Gnb1 & Fth1 & Lpl & S100a16 \\
    Cyb5 & Odc1 & Tacstd2 & Psma7 & Rpl36a & Timp2 & Mrpl52 & Serf2 \\
    Srsf2 & Higd1a & Ndufb5 & Tuba4a & Myeov2 & Cd8a & Phlda1 & Cd8b1 \\
    Rpl31\_ps12 & Skp1a & Tmem59 & Blvrb & Uqcr10 & Psma2 & Rps15a & Hspa5 \\
    Pcna & Oat & Ncl & Ybx1 & Gpx3 & x2200002D01Rik & Gsta4 & Hes1 \\
    Sparc & Cpe & Fmod & Eif3k & Chad & Prg4 & Srsf7 & Pdcd4 \\
    Serpine2 & Gcat & Ppp1r2 & Rpl39 & Cox5a & Tpi1 & Rab10 & Chit1 \\
    Ndufs6 & Uqcr11 & Ndufa6 & Ucp2 & Ndufa13 & Pgam1 & Id2 & Atp5g2 \\
    Rcan1 & Arhgdib & Hnrnpu & x2810417H13Rik & Krtcap2 & Hist1h1e & Rrm2 & Manf \\
    Timm13 & Tspo & Naca & Fam162a & Sfr1 & Nrp1 & Acsm2 & Apob \\
    Hpx & Rabac1 & Cdo1 & Atp6v1f & Acaa1b & Fxyd2 & Akr1c21 & Slc27a2 \\
    Hpd & Apoh & Rgn & Serpina1b & Fabp1 & Mup2 & Ndrg1 & Upk3a \\
    Gclc & Col3a1 & Dynll1 & Hmgb1 & Gng5 & Ralbp1 & Ucma & Col11a1 \\
    Comp & Pdzk1ip1 & Ndufb7 & Calm4 & Anxa5 & Apoc4 & Gm13889 & Krt15 \\
    Eif5 & Glycam1 & Ube2c & Psma3 & Rpl24 & Rpl38 & Eef1g & Gapdh \\
    \hline
  \end{tabular}
\end{table}

\begin{table}[p]
  \centering
  \caption{Tabula Muris SHAP gene list (part 3 of 3; remaining 288 genes).}
  \label{supp_table:tm_shap_genes_part_3}
  \begin{tabular}{llllllll}
    \hline
    Acp5 & Rap1b & Hspb8 & Col2a1 & Il6 & Tomm20 & Dsp & Klf4 \\
    Malat1 & Epsti1 & Ccl2 & Has1 & Neat1 & Prelid1 & Calr & Hspb1 \\
    P4hb & Ppib & Fdps & Mup3 & Cxcl14 & Cxcl2 & x2010107E04Rik & Hrsp12 \\
    Miox & Cat & Col8a1 & Cxcl10 & Mgst1 & C1qc & Rpl13a & Mpo \\
    Lcn2 & Prtn3 & Eif3e & Gas5 & Impdh2 & Npm1 & Tagln2 & Cox7b \\
    Hsp90ab1 & Tcp1 & Procr & Psmb1 & Trf & Emp1 & x1500015O10Rik & Hist1h2ao \\
    Myl7 & Tpm1 & Rab11a & Akr1b8 & Rpl36 & Scgb3a2 & Sbpl & Scgb3a1 \\
    Tff2 & Ndufa1 & Serpinb5 & Eif3f & Prdx6 & Areg & Cldn4 & Hnrnpa3 \\
    Eln & Tnfrsf12a & Tmem27 & Acy3 & Aldob & Cyp4b1 & Guca2b & Ranbp1 \\
    Anxa2 & Klk1 & Tppp3 & Psmb2 & Calm2 & Ttc36 & Tcn2 & Tecr \\
    Hilpda & Gstm1 & Ttr & Cyp2e1 & Scd1 & Esd & Sep15 & Egf \\
    Wfdc15b & Sprr1a & x1110032A04Rik & Bpifa1 & Mia1 & Sftpa1 & Fgg & Gltp \\
    Aldh3a1 & Dsg1a & Timp1 & Ccl19 & Fgb & Sult2b1 & Cd63 & Homer2 \\
    Rbp4 & Fmo5 & Sbsn & Snrpf & Cox5b & Scp2 & Atp5j2 & Cox7c \\
    Cma1 & Retnla & Ctnnbip1 & Erh & Serpinh1 & Fn1 & Ywhae & Set \\
    Apoc3 & Fga & Cyp3a11 & Mup20 & Spink3 & Mrpl33 & Apoa2 & Alb \\
    Serpina1d & Hn1 & Thbs1 & Fam25c & Gjb2 & Barx2 & Krt24 & Gem \\
    Ube2s & Spp1 & Gc & Krtdap & Car3 & Gsta3 & Ahsg & Krt10 \\
    Serpina1c & Serpina3k & Apoc1 & Kdm6b & Dut & Gpx4 & Tubb2a & Csf3 \\
    Epcam & Expi & Clu & S100g & Tmem14c & Atp5c1 & Col1a2 & Fgfbp1 \\
    Sprr1b & Cnfn & Selenbp1 & Dapl1 & Cst6 & Krt13 & Krt4 & Ccl20 \\
    Krt6b & x2310002L13Rik & Ly6g6c & Krt23 & Tgm3 & Atp5g1 & Col1a1 & Nupr1 \\
    Lmo1 & Tesc & Inmt & Psapl1 & Efhd2 & Apod & Calb1 & Col10a1 \\
    Krt17 & Chi3l1 & Serpinb2 & Mfge8 & Calca & Reg3g & Selp & Krt35 \\
    Igfbp5 & Serpinb3a & Elovl4 & x2310002J15Rik & Cldn7 & Ctsk & Prr9 & Msmp \\
    x2310079G19Rik & Lce3c & Serpina1a & Rn45s & Serpinb12 & Defb4 & Rbp2 & Klk14 \\
    Ptx3 & Adh7 & Gstp1 & Sox9 & Csn3 & Lrrc15 & Kap & x8430408G22Rik \\
    Atp2a2 & Myh6 & Ttn & Ada & Lor & Krt76 & Slurp1 & Crct1 \\
    Darc & Krt36 & Acan & Cytl1 & Basp1 & Krt84 & Tchh & Chi3l4 \\
    Lipf & x3110079O15Rik & G0s2 & Bpifb1 & Slc14a2 & Krt85 & Sprr2i & Otor \\
    Dcpp2 & Krt33b & Dcpp3 & Krtap3\_3 & Mfap4 & x1600029D21Rik & Car1 & Aqp2 \\
    Lyg1 & Serpina1e & Krtap13\_1 & Serpinb3c & Pf4 & Ppbp & Sprr3 & Ppa1 \\
    Krt34 & Krt81 & Bhmt & Krt31 & Dcpp1 & Krtap3\_1 & Krt42 & x5430421N21Rik \\
    Krtap13 & Krtap9\_3 & Psors1c2 & Krt33a & Krt86 & Krtap3\_2 & x2310034C09Rik & x2310061N02Rik \\
    Cyp2c69 & Hamp & Klk10 & Ptms & Cox7a2 & Mt4 & Pam & Pgls \\
    Aldh1a1 & Capns1 & Sod1 & Cebpb & Fstl1 & Gsn & Ifi205 & Igfbp6 \\
    \hline
  \end{tabular}
\end{table}

%% file: files/pan_label_abbreviation.tex
\begin{table}[pt]
\centering
\caption{Pan Cancer dataset cancer types abbreviation and full name.}
\label{supp_tab:pan_label_abbreviatio}
\begin{tabular}{ll}
\hline
\textbf{Abbreviation} & \textbf{Full Name} \\
\hline
LAML & Acute Myeloid Leukemia \\
ACC & Adrenocortical Cancer \\
BLCA & Bladder Urothelial Carcinoma \\
LGG & Brain Lower Grade Glioma \\
BRCA & Breast Invasive Carcinoma \\
CESC & Cervical \& Endocervical Cancer \\
CHOL & Cholangiocarcinoma \\
COAD & Colon Adenocarcinoma \\
DLBC & Diffuse Large B-cell Lymphoma \\
ESCA & Esophageal Carcinoma \\
HNSC & Head \& Neck Squamous Cell Carcinoma \\
KICH & Kidney Chromophobe \\
KIRC & Kidney Clear Cell Carcinoma \\
KIRP & Kidney Papillary Cell Carcinoma \\
LIHC & Liver Hepatocellular Carcinoma \\
LUAD & Lung Adenocarcinoma \\
LUSC & Lung Squamous Cell Carcinoma \\
MESO & Mesothelioma \\
OV & Ovarian Serous Cystadenocarcinoma \\
PAAD & Pancreatic Adenocarcinoma \\
PCPG & Pheochromocytoma \& Paraganglioma \\
PRAD & Prostate Adenocarcinoma \\
READ & Rectum Adenocarcinoma \\
SARC & Sarcoma \\
SKCM & Skin Cutaneous Melanoma \\
STAD & Stomach Adenocarcinoma \\
TGCT & Testicular Germ Cell Tumor \\
THYM & Thymoma \\
THCA & Thyroid Carcinoma \\
UCS & Uterine Carcinosarcoma \\
UCEC & Uterine Corpus Endometrioid Carcinoma \\
UVM & Uveal Melanoma \\
    \hline
\end{tabular}
\end{table}

%% file: files/pan_labels.tex
\begin{table}[p]
  \centering
  \caption{Pan-cancer label key (indices 1--32, in model order).}
  \label{supp_table:pan_class_name}
  \begin{tabular}{rlrlrlrl}
    \hline
    1 & LAML & 2 & ACC & 3 & BLCA & 4 & LGG \\
    5 & BRCA & 6 & CESC & 7 & CHOL & 8 & COAD \\
    9 & DLBC & 10 & ESCA & 11 & HNSC & 12 & KICH \\
    13 & KIRC & 14 & KIRP & 15 & LIHC & 16 & LUAD \\
    17 & LUSC & 18 & MESO & 19 & OV & 20 & PAAD \\
    21 & PCPG & 22 & PRAD & 23 & READ & 24 & SARC \\
    25 & SKCM & 26 & STAD & 27 & TGCT & 28 & THYM \\
    29 & THCA & 30 & UCS & 31 & UCEC & 32 & UVM \\
    \hline
  \end{tabular}
\end{table}

%% file: files/pan_gene_list.tex

\begin{table}[p]
  \centering
  \caption{Gene list for the pan-cancer SHAP heatmap (part 1 of 2).}
  \label{supp_table:pan_shap_genes_part_1}
  \begin{tabular}{llllllll}
    \hline
    CHST4 & TRPA1 & PRAC & NKX2-3 & NOX1 & TYR & DLX6 & DLX6AS \\
    UGT1A7 & KRTDAP & TMPRSS11A & PGC & ISL1 & REG4 & RSPO3 & SCGB2A1 \\
    LYPD2 & UCA1 & CALML3 & REG3A & SERPINB4 & TSIX & ACTA1 & NR0B1 \\
    ADAMDEC1 & BLK & CD19 & TRIM31 & LOC100124692 & DKK1 & XK & NTS \\
    SFTPA1 & OSR1 & KIAA2022 & WISP1 & HOXC4 & CALB2 & GALNT9 & HOXA11 \\
    PTPN5 & C1orf130 & CXCL11 & CSMD2 & TBX1 & LGR5 & NKX2-5 & C1orf168 \\
    LIX1 & STX19 & C1orf161 & EMX2OS & LOC440905 & NNAT & TH & CHD5 \\
    DUSP15 & IL1F9 & SLURP1 & COX6B2 & OTX1 & SCGB3A2 & STK31 & B3GNT6 \\
    MUC5B & SEMA3A & C20orf114 & MMP12 & PLUNC & CXCL17 & C19orf59 & SCEL \\
    LOC100131551 & IGFL2 & CLEC18C & PRODH2 & CCBE1 & CLCNKA & ATP6V0A4 & LRRTM1 \\
    ST6GALNAC5 & TMEM61 & CXADR & FA2H & ESRRG & HOXC8 & HOXC9 & HOXD9 \\
    AREG & FAM184B & LOC146336 & GRAMD2 & RASEF & ROR2 & LAD1 & SHANK2 \\
    EN1 & HOXD13 & MIA & MGC45800 & DPF1 & TMEM125 & ANO9 & HAP1 \\
    MAPK15 & XDH & RSPO4 & TP53AIP1 & KRT4 & SVOPL & MUC13 & LCN12 \\
    RNF183 & MKX & AQP2 & DMRT3 & LOC440173 & ANGPTL1 & IGSF10 & MAB21L1 \\
    DLX1 & CACNG4 & VTCN1 & HMGA2 & DLX2 & SIX1 & EYA4 & HAPLN4 \\
    KIRREL2 & CASQ1 & EPYC & SSTR1 & FBN3 & ZIC5 & B3GALT2 & GABRG3 \\
    COMP & NTNG1 & COL11A1 & HOXA11AS & AQP9 & CCL21 & SCTR & C1orf172 \\
    GRID1 & TMEM35 & FOXD3 & MLANA & FUT6 & NBLA00301 & LPPR5 & ZNF536 \\
    ROPN1B & C2orf54 & FOXP2 & PDZK1IP1 & PLA2G10 & LCN2 & C6orf222 & CYP2B6 \\
    MYO1A & RNF186 & CCL11 & OVOL1 & AADAC & CASP14 & FABP1 & FAM3D \\
    LEMD1 & CCL13 & HDC & AQP5 & PRSS35 & GAP43 & TNNI2 & NEFM \\
    SLC22A6 & FAM189A1 & PKHD1 & ELFN2 & PCSK1N & FABP7 & ST8SIA6 & PDZD3 \\
    FCAMR & HAVCR1 & SHH & ADCYAP1 & ZDHHC8P1 & CNTN3 & HAS1 & MYO3A \\
    PGLYRP3 & XIST & EPHX3 & KRT80 & CSF3 & HEPACAM2 & LRRC8E & FOXN1 \\
    PRSS21 & CDHR2 & MEP1A & AQP6 & CD38 & C4orf7 & GAL3ST3 & AGR3 \\
    PADI3 & ACTL8 & GJB7 & HOXC11 & HOXA7 & RHOD & DARC & WISP3 \\
    LRP4 & PRKG2 & ALK & HCN1 & PDIA2 & ZG16B & SCGB2A2 & SPRR2B \\
    WNT10A & C19orf21 & CNGA1 & TNFRSF17 & COL8A1 & FCRLA & C3orf55 & FERMT1 \\
    PIP5K1B & CLCA4 & KCNK1 & RIMBP2 & HAL & SLCO1B3 & AGR2 & ITGB6 \\
    DACH1 & MCHR1 & C19orf46 & PRKAA2 & CES4 & TMPRSS3 & FOXL2 & IL1A \\
    PLA2G2F & CECR2 & NTF4 & WNT7B & GPR115 & SPRR2C & CTSG & EPHA10 \\
    HS3ST6 & TMSB4Y & OPRK1 & FSD1 & PPP1R14D & LRRC4C & UGT1A10 & TNS4 \\
    C3orf72 & HOXC13 & MEOX1 & TMEM171 & MT3 & SLC38A4 & DIRAS2 & FBLN1 \\
    GREM2 & WNK2 & VEPH1 & CLEC18B & MASP1 & CDK5R2 & MPPED1 & GDPD2 \\
    SLC7A3 & TCF21 & TMEM132D & PCDH8 & SLITRK2 & CR2 & SLC35F3 & MOGAT2 \\
    PDX1 & PHGR1 & PRR11 & MNX1 & GJB3 & MUC2 & PLA2G4F & TJP3 \\
    ALDH3B2 & FXYD1 & COL4A6 & EPN3 & ABCA12 & ANXA8 & MYBPC1 & SORCS1 \\
    KIAA1755 & GALNT13 & UNC93A & SLC28A3 & CD164L2 & FAM178B & ANKRD56 & CLUL1 \\
    PRSS16 & CBLC & PRR15 & GGT6 & KRT19 & ABO & PPP1R1B & NIPAL1 \\
    CHST9 & SMPDL3B & AKR1C1 & CYP4F11 & CHRNA4 & C8B & ORM2 & SERPINA11 \\
    GSG1L & SP5 & SGK2 & CEACAM7 & NAT8L & CLRN3 & CDX2 & PRSS3 \\
    VWA5B2 & C1orf106 & STYK1 & CAPN8 & NETO1 & ARMC3 & PCDHA12 & CHST6 \\
    KRT7 & TRPM3 & IL17D & KCNA1 & SLC5A1 & NALCN & USH1G & LRRN1 \\
    MXRA5 & TPSAB1 & C10orf107 & SLC35F1 & GSDMC & ODAM & SBSN & DUOXA1 \\
    KLK12 & EN2 & NKAIN1 & DLGAP1 & PCDH10 & C13orf33 & PABPC1L2B & CCDC135 \\
    LPAR3 & GABRB3 & KLK7 & SLC7A9 & TYRP1 & IRX4 & ATP2B3 & HSD17B2 \\
    MMP1 & CD70 & PITX1 & CGREF1 & SYT7 & CMTM5 & ITIH5L & IGSF9B \\
    C2 & ENPP6 & SCN2B & LRFN5 & TDRD9 & MAT1A & ADARB2 & PRIMA1 \\
    GPR172B & GJB5 & LOC643763 & FAM196B & LOC283174 & TNNT1 & GBP6 & GPR109A \\
    KCNH6 & TMEM90B & NLGN4Y & NPY6R & PDPN & MAMDC2 & POU2F3 & UGT1A6 \\
    CST1 & LOC100190940 & FOXA1 & PVRL4 & SERTAD4 & ARHGDIG & MALL & TBX18 \\
    OVOL2 & KRT78 & AKR1B15 & RPTN & ASPG & IGFL1 & ADAM33 & SYT14 \\
    TPO & CYP26A1 & HOXA13 & ARL14 & CCL15 & DEFB1 & ANO5 & ENAM \\
    KRT6C & ZIC2 & ATP2C2 & AGTR1 & EGFL6 & SH3RF2 & SLC44A5 & C1orf210 \\
    GSTA1 & STRA6 & CPA4 & GPX2 & AK5 & C1orf95 & FAM167A & C5orf46 \\
    FOXI1 & CNTN1 & DRP2 & BEST3 & SPAG17 & HOXA10 & SLITRK3 & CYP2C9 \\
    PIP & HTR1D & ALPP & LOC100130933 & BTNL3 & DLX3 & FAM5B & KRT31 \\
    IL20RA & CNGA3 & DSCAM & PRR18 & A2ML1 & WFDC5 & MAGEA9B & WNT3A \\
    ELF5 & RNASE7 & NKAIN2 & SPRR2E & S100A12 & NOG & SDR16C5 & SPRR2D \\
    \hline
  \end{tabular}
\end{table}

\begin{table}[p]
  \centering
  \caption{Gene list for the pan-cancer SHAP heatmap (part 2 of 2).}
  \label{supp_table:pan_shap_genes_part_2}
  \begin{tabular}{llllllll}
    \hline
    IGFBP1 & CYP3A4 & TCEAL2 & CDH16 & SLC12A1 & ITIH1 & KNG1 & ACMSD \\
    ERMN & SLC22A11 & KCNJ16 & PKLR & DMGDH & POU3F3 & SLC17A4 & FAM151A \\
    SLC2A2 & ESYT3 & DSG3 & TMEM132C & SYT1 & MSX2 & CD1A & GFRA3 \\
    PSORS1C2 & TNNC2 & LOC286002 & MUM1L1 & TRIM15 & TMEM139 & LGI1 & MFAP5 \\
    CPA6 & ST18 & TMEM100 & CCDC160 & LOC723809 & TNNI1 & SCRG1 & VSIG1 \\
    CAMK2A & CNTFR & KRT1 & CXCL6 & BHMT & USP9Y & DDX3Y & PRKY \\
    CAPN13 & IGF2AS & UTY & C10orf82 & ACTC1 & NUDT10 & CRABP1 & RGS22 \\
    CIDEC & DLK1 & NMU & SCARA5 & AURKB & KRT16 & MYO3B & SLC45A2 \\
    PRUNE2 & TDO2 & FGFBP1 & MDGA2 & ZFY & TAGLN3 & BEND4 & MSI1 \\
    CA9 & PKDCC & DLK2 & RHCG & ARSF & CA10 & CRTAC1 & CPEB1 \\
    SYTL1 & ABCC6P1 & SPRR2G & BCL2L14 & TMEFF2 & MKRN3 & ADCY8 & FCN1 \\
    SPEF1 & TSPAN12 & HOXB8 & TPSB2 & C1orf186 & CDC20B & TBC1D3G & HTR3A \\
    FLRT3 & HPD & GPR37L1 & PCDHA13 & ENTPD8 & PLN & PROM2 & C15orf48 \\
    TTBK1 & SLC4A10 & PRLR & WNT2 & BIRC7 & CYP17A1 & INHA & CD161 \\
    CXCL13 & ACPP & PDZRN4 & TUBBP5 & CTSL2 & KRT15 & STC1 & DCDC2 \\
    CLDN9 & PLA2G2D & BAAT & HRG & CACNA1E & GNG4 & CPLX3 & ISX \\
    TCN1 & LOC100131726 & EHF & OR51E2 & EDIL3 & GYLTL1B & CACNA2D1 & FGF14 \\
    SOX9 & FAM133A & SLC26A3 & LEP & SLC6A11 & STK33 & HKDC1 & NKX2-2 \\
    IL6 & MAGEA11 & FGA & RIMS1 & CPA3 & SDR42E1 & BTBD17 & KRT75 \\
    DQX1 & LUZP2 & GJB6 & MAGEA4 & C1orf64 & GLYATL2 & SLC38A11 & ANKRD30A \\
    KLHDC7A & LRAT & APOH & CYP2C18 & ABCG8 & AHSG & SLC6A1 & C4BPB \\
    CPB2 & CRP & CFHR2 & CFHR5 & APOA4 & PLA2G12B & THPO & UGT2B15 \\
    TM4SF4 & CD5L & APOC3 & HAO1 & UPB1 & CREB3L3 & ALB & SERPINA7 \\
    G6PC & PHF21B & PLCXD3 & SYT9 & PTCHD1 & NRSN1 & RPRM & CHST8 \\
    C11orf87 & C13orf30 & DAZ1 & SERPINA4 & ANXA13 & GPR128 & CADM2 & PARM1 \\
    PCSK1 & KSR2 & SIGLEC8 & GABBR2 & LRRTM4 & COL28A1 & SRRM4 & ADRA1B \\
    TMEM59L & CYP11A1 & ENPP3 & ST8SIA2 & TRDN & CYP4F12 & FAT3 & ILDR1 \\
    ADAM22 & TDGF1 & PCDHA10 & SNORD116-4 & TMEM82 & USH1C & P2RX6 & MGAT5B \\
    PLA2G4D & MAGEA12 & MAGEA3 & OMG & CPNE6 & GAD1 & PCDHAC2 & ARG1 \\
    MYEOV & CHGA & DBH & SCG2 & SLC7A14 & ACSM5 & SLC6A19 & CYP4Z2P \\
    TGM5 & NPAS3 & CNDP1 & LOC647309 & CA12 & FEV & PCSK2 & CALY \\
    TMEM145 & TMEM151A & GPR39 & GDAP1L1 & ANKRD30B & C7orf52 & KIAA0125 & RIMS4 \\
    SLC5A8 & PKNOX2 & GSTT1 & HOXB6 & GP2 & SERPINA6 & C6orf218 & FAM181A \\
    PEBP4 & UGT3A1 & A2BP1 & TMEM195 & PCDHGC4 & PCDHGC5 & GPR17 & CAPSL \\
    SLC6A3 & FBXO2 & KCNB1 & SULT4A1 & CDH22 & MUCL1 & IP6K3 & SEZ6 \\
    KLK5 & BCMO1 & CDH20 & ELAVL3 & SNCG & LOC255167 & TIMD4 & DAO \\
    HGD & LRRTM3 & ACTL6B & SPHKAP & VWA3A & AMBP & SYT3 & HAO2 \\
    ACSM2A & SLC17A3 & DNAI1 & FLG & GABRA3 & C3orf15 & AKR1B10 & CACNG7 \\
    B4GALNT4 & MGAT4C & ADH4 & SLC8A3 & IGDCC3 & ACTN2 & UGT1A3 & GRIK2 \\
    SLC10A1 & CLCNKB & MT1A & GFAP & CHIT1 & TMEM150B & SERPINA10 & CYP4F2 \\
    GRM5 & EPHB6 & SOD3 & C6orf174 & KCNQ2 & ACSM2B & SLC30A10 & KLB \\
    NKX2-1 & NCRNA00185 & STMN4 & FAM57B & CAMKV & SLC28A1 & GAL3ST1 & TNFSF12-TNFSF13 \\
    UMOD & DRD2 & OLIG2 & QRFPR & GPC3 & ADH1B & PPP4R4 & AFF2 \\
    WIT1 & DPEP1 & HPR & DCT & GRIA4 & SLC13A1 & GATA4 & CNTN6 \\
    SH3GL3 & FAM135B & ACCN4 & MMD2 & HEPN1 & ZFP57 & PCDHA1 & PIWIL1 \\
    CCKBR & SLN & GTSF1 & PVALB & GRM4 & HSD11B1 & SLC5A12 & C14orf73 \\
    NPR3 & DIO1 & ABCC6P2 & CTAG1B & FIGF & NEB & GC & RNF128 \\
    SMYD1 & C6orf142 & ST8SIA3 & UGT2B10 & MTTP & AFP & GLYAT & NAT8 \\
    THRSP & CYP8B1 & MOGAT3 & CP & VGF & HEPACAM & PAQR9 & ENPP1 \\
    PRSS33 & HPCAL4 & LRRC19 & L1TD1 & MAGEA1 & TFPI2 & MYT1L & NXF2 \\
    TCL1A & TBX15 & CHRNB2 & MAGEA2 & SULT2A1 & FADS6 & LOC100271831 & CA3 \\
    AMDHD1 & TKTL1 & BICC1 & SHD & SLC26A7 & FAM131C & LHFPL4 & SLC30A2 \\
    ZG16 & LIPG & RBM24 & BTBD16 & MYO18B & RHBG & TRPM8 & ZNHIT2 \\
    HBA1 & SCN4A & MAGEC2 & DNAH11 & LCN10 & ADRA2C & CRYM & SCRT1 \\
    LOC84856 & MGAM & OIT3 & C19orf77 & EPS8L3 & VNN3 & MT1H & CYP4A11 \\
    INHBE & F7 & C10orf108 & NR1I2 & OGN & KLK3 & ASPDH & FGG \\
    S100A14 & PDE1A & RIC3 & CHL1 & CHRDL1 & CNTD2 & CPE & RGS7BP \\
    DDX25 & MMP13 & ATP13A4 & THSD7B &  &  &  &  \\
    FAM123A & KIAA1045 & SLC5A10 & GRP & CDSN & ELMOD1 & LOC154822 & WDR49 \\
    PRSS1 & AJAP1 & ID4 & CCL20 & CD1E & PCDH11X & TTLL6 & F12 \\
    NRG2 & DMBX1 & SCN2A & APOA1 & MAB21L2 & NAT2 & DCX & SLC7A10 \\
    UNC5D & FLJ16779 & C12orf53 & NKAIN4 & C19orf33 & KLK8 & RASSF10 & BBOX1 \\
    CNNM1 & KIAA0319 & UPK3A & TNNT3 & DSG1 & SLC6A15 & DMRTC1B & CDH10 \\
    KCNIP1 & LSAMP & PKP3 & ASPA & LRRTM2 & FTCD & C8orf47 & NR1H4 \\
    ELOVL2 & GDA & LRRIQ1 & ODZ2 & VTN & PSCA & ZNF114 & CCL14-CCL15 \\
    CFHR3 & HSD17B13 & UGT1A1 & CFHR1 &  &  &  &  \\
    \hline
  \end{tabular}
\end{table}

%% file: files/pnet_gene_dendogram.tex

\begin{table}[p]
  \centering
  \caption{Prostate gene dendrogram key (bottom-to-top order).}
  \label{supp_table:pnet_gene_dendrogram_key}
  \begin{tabular}{llllllll}
    \hline
    1: SEMG2 & 2: RPS27 & 3: RPS29 & 4: A1BG & 5: APOC1 & 6: PLG & 7: SERPING1 & 8: RPS13 \\
    9: C4B & 10: TSPAN8 & 11: ATP5F1A & 12: RPL23 & 13: FTL & 14: APCS & 15: APOA2 & 16: ALB \\
    17: FABP1 & 18: SEMG1 & 19: RPS18 & 20: RGS2 & 21: NDUFA1 & 22: SEC61G & 23: DAPL1 & 24: PABPC1 \\
    25: RPL7 & 26: HNRNPA2B1 & 27: LUM & 28: ATP5MF & 29: XBP1 & 30: LMAN2 & 31: PDCD4 & 32: IFITM1 \\
    33: PRDX1 & 34: UBA52 & 35: NAP1L1 & 36: RPS26 & 37: NDUFC2 & 38: PDIA4 & 39: EIF3E & 40: RPS5 \\
    41: SSR2 & 42: IFITM3 & 43: MYL12A & 44: NFKBIA & 45: NME2 & 46: HSPE1 & 47: ANXA2 & 48: MUC1 \\
    49: HEXB & 50: PDIA3 & 51: NDUFB10 & 52: TFF3 & 53: H4C4 & 54: MT1E & 55: ATP6V0C & 56: PDLIM5 \\
    57: MZT2A & 58: SELENOW & 59: MDK & 60: OLFM4 & 61: SERHL2 & 62: CCN1 & 63: DHCR24 & 64: PUF60 \\
    65: F3 & 66: HLA\_DMB & 67: CFL1 & 68: HSP90AB1 & 69: ST6GAL1 & 70: RARRES2 & 71: SELENOH & 72: EIF4B \\
    73: RPL37A & 74: RPL19 & 75: TLE5 & 76: COMT & 77: APOA1 & 78: STEAP2 & 79: TIMP1 & 80: CP \\
    81: PRRC2A & 82: FGG & 83: C1S & 84: CALM2 & 85: RPL35 & 86: RPS27A & 87: CPE & 88: CES1 \\
    89: TF & 90: RPL39 & 91: HSD17B6 & 92: RPS10 & 93: FADS2 & 94: S100A10 & 95: DYNLL1 & 96: TMA7 \\
    97: RPS3 & 98: COX6C & 99: RPL27 & 100: ACTG1 & 101: GADD45G & 102: SLC25A39 & 103: CHCHD2 & 104: NDUFB2 \\
    105: LDHA & 106: AMBP & 107: RHOC & 108: BCAP31 & 109: RBP4 & 110: REXO2 & 111: SRSF5 & 112: EPHX1 \\
    113: KRT19 & 114: RPL18A & 115: PEX10 & 116: RPS16 & 117: HP & 118: SRSF2 & 119: FABP5 & 120: QARS1 \\
    121: HNRNPA1 & 122: LTBP4 & 123: S100A8 & 124: FMOD & 125: RPS17 & 126: SLPI & 127: MT1F & 128: RPS15 \\
    129: ATP6V0E2 & 130: CD74 & 131: LUC7L2 & 132: ATP5PD & 133: CUTA & 134: ACTN4 & 135: HLA\_B & 136: GNAS \\
    137: MIEN1 & 138: RPL18 & 139: PIEZO1 & 140: NCAPD3 & 141: SELENOP & 142: GFUS & 143: SERP1 & 144: CEL \\
    145: A2M & 146: APOD & 147: TSTD1 & 148: DLK1 & 149: NDUFA13 & 150: NPC2 & 151: TPD52 & 152: APLP2 \\
    153: PLPP1 & 154: DDT & 155: DUS1L & 156: NDUFA2 & 157: APOC3 & 158: EIF4A2 & 159: SUMO2 & 160: APOE \\
    161: HINT1 & 162: ITIH4 & 163: TTR & 164: CFB & 165: RPL14 & 166: EEF1A1 & 167: ATP5MG & 168: COL4A1 \\
    169: COX7B & 170: HMGN2 & 171: RPS23 & 172: WFDC2 & 173: KRT18 & 174: NOP53 & 175: KLK3 & 176: CANX \\
    177: TUBB & 178: NEFH & 179: RAN & 180: YBX1 & 181: BANF1 & 182: SKP1 & 183: NDUFB9 & 184: COX6A1 \\
    185: COX8A & 186: EIF4A1 & 187: NRP1 & 188: C4A & 189: EIF3C & 190: PPDPF & 191: NUPR1 & 192: SARAF \\
    193: CSRP1 & 194: HMGB2 & 195: TMEM256 & 196: EIF3H & 197: P4HB & 198: DES & 199: MYL9 & 200: PPIB \\
    201: MYH11 & 202: PRKDC & 203: CST2 & 204: ANPEP & 205: STEAP4 & 206: ACTA2 & 207: BCAM & 208: CREB3L1 \\
    209: LTF & 210: C1orf43 & 211: TMSB10 & 212: ATP5MC1 & 213: CST3 & 214: PTGDS & 215: FLNA & 216: S100A4 \\
    217: SERPINA1 & 218: RPL7A & 219: SRRM2 & 220: AMD1 & 221: NPDC1 & 222: TPM1 & 223: EIF3L & 224: GABARAP \\
    225: KRT8 & 226: CD81 & 227: KRT5 & 228: ARPC3 & 229: RPLP0 & 230: TAF1 & 231: ANG & 232: HPX \\
    233: ITIH1 & 234: AHSG & 235: RPL29 & 236: VTN & 237: FGA & 238: IGLL5 & 239: FGL1 & 240: POR \\
    241: RPS12 & 242: MYL12B & 243: RPL32 & 244: UBB & 245: TOMM7 & 246: FXYD3 & 247: MICOS10 & 248: CPSF1 \\
    249: ORM2 & 250: SRP14 & 251: ATP6V1G1 & 252: UBC & 253: PSMD2 & 254: RPL4 & 255: RPS6 & 256: SPINT2 \\
    257: TMEM59 & 258: APEX1 & 259: CSNK1E & 260: COL4A2 & 261: PKM & 262: BGN & 263: SON & 264: BTF3 \\
    265: MESP1 & 266: COL9A2 & 267: SLC25A6 & 268: RPL23A & 269: VIM & 270: ANXA5 & 271: C1QB & 272: RPLP1 \\
    273: HSPA5 & 274: TMED3 & 275: ST13 & 276: ZG16B & 277: CAPNS1 & 278: CTSZ & 279: RPS19 & 280: PTMS \\
    281: SERPINH1 & 282: CD9 & 283: PLXND1 & 284: RPL10 & 285: RPL24 & 286: PARK7 & 287: ATP5MC2 & 288: RPL22 \\
    289: ATP6V0E1 & 290: RPL13A & 291: RPS14 & 292: GPX1 & 293: SLC44A4 & 294: EIF5A & 295: STMN1 & 296: BSG \\
    297: TPT1 & 298: PRSS8 & 299: TSPAN1 & 300: DSTN & 301: ATF4 & 302: THY1 &  &  \\
    \hline
  \end{tabular}
\end{table}